\documentclass[journal]{IEEEtran}
\ifCLASSOPTIONcompsoc
  \usepackage[nocompress]{cite}
\else
  \usepackage{cite}
\fi
\ifCLASSINFOpdf
\else
\fi
\usepackage[utf8]{inputenc} 
\usepackage[T1]{fontenc}    
\usepackage{url}            
\usepackage{amsfonts}       
\usepackage{nicefrac}       
\usepackage{microtype}      
\usepackage{graphicx}
\usepackage{mathtools}
\usepackage{multirow}
\usepackage{subfigure}
\usepackage{tabularx}
\usepackage{amsmath,amsfonts,amssymb}
\usepackage{wrapfig,lipsum,booktabs}
\usepackage{xcolor}
\usepackage[pagebackref=true,breaklinks=true,letterpaper=true,colorlinks,bookmarks=false,citecolor=cyan]{hyperref}
\usepackage{balance}
\def\eg{\textit{e.g.}}
\def\ie{\textit{i.e.}}

\usepackage{tikz}

\usepackage{enumitem}

\begin{document}
\title{Continual Attentive Fusion for Incremental Learning in Semantic Segmentation}
%
%
%
%

\author{Guanglei Yang,
        Enrico Fini,
        Dan Xu,
        Paolo Rota,
        Mingli Ding$^*$,
        Hao Tang,\\
        Xavier Alameda-Pineda,~\IEEEmembership{Senior Member,~IEEE,}
        Elisa Ricci,~\IEEEmembership{Member,~IEEE}%
\IEEEcompsocitemizethanks{
\IEEEcompsocthanksitem Guanglei Yang and Mingli Ding are with School of Instrument Science
and Engineering, Harbin Institute of Technology (HIT), Harbin, China. E-mail: \{yangguanglei,dingml\}@hit.edu.cn. $^*$Corresponding author. \protect
\IEEEcompsocthanksitem Enrico Fini, Paolo Rota, and Elisa Ricci are with the Department of Information
Engineering and Computer Science, University of Trento, Italy. E-mail: \{enrico.fini, paolo.rota, elisa.ricci\}@unitn.it.\protect
\IEEEcompsocthanksitem Dan Xu is with the Department of Computer Science and Engineering,
Hong Kong University of Science and Technology. E-mail: danxu@cse.ust.hk.\protect
\IEEEcompsocthanksitem Hao Tang is with the Department of Information Technology and Electrical Engineering, ETH Zurich. E-mail: hao.tang@vision.ee.ethz.ch  \protect
\IEEEcompsocthanksitem Xavier Alameda-Pineda is with the RobotLearn Group, INRIA. E-mail:
xavier.alameda-pineda@inria.fr.\protect
\IEEEcompsocthanksitem Elisa Ricci is with Deep Visual Learning group at Fondazione Bruno
Kessler, Trento, Italy.\protect

}
}

%
%

\markboth{Submitted to IEEE Transactions on Multimedia}%
{Shell \MakeLowercase{\textit{et al.}}: Bare Demo of IEEEtran.cls for Computer Society Journals}
%



\IEEEtitleabstractindextext{%
\begin{abstract}
Over the past years, semantic segmentation, as many other tasks in computer vision, benefited from the progress in deep neural networks, 
resulting in significantly improved performance. However, deep architectures trained with gradient-based techniques suffer from catastrophic forgetting, which is the tendency to forget previously learned knowledge while learning new tasks. Aiming at devising strategies to counteract this effect, incremental learning approaches have gained popularity over the past years. However, the first incremental learning methods for semantic segmentation appeared only recently. While effective, these approaches do not account for a crucial aspect in pixel-level dense prediction problems, \textit{i.e.} the role of attention mechanisms. To fill this gap, in this paper we introduce a novel attentive feature distillation approach to mitigate catastrophic forgetting while accounting for semantic spatial- and channel-level dependencies. Furthermore, we propose a {continual attentive fusion} structure, which takes advantage of the attention learned from the new and the old tasks while learning features for the new task. Finally, we also introduce a novel strategy to account for the background class in the distillation loss, thus preventing biased predictions. We demonstrate the effectiveness of our approach with an extensive evaluation on Pascal-VOC 2012 and ADE20K, setting a new state of the art.
\end{abstract}
\begin{IEEEkeywords}
Knowledge Distillation, Incremental Learning, Semantic Segmentation.
\end{IEEEkeywords}}

\maketitle

\IEEEdisplaynontitleabstractindextext

%
\IEEEpeerreviewmaketitle

\section{Introduction}
\label{sec:intro}

During the last decade, the emergence of deep learning has lead to several breakthroughs in many computer vision and multimedia tasks. 
Semantic segmentation, the problem of assigning a semantic label to each pixel in an image, was no exception to this trend \cite{minaee2020image}. 
Sophisticated deep neural networks as fully convolutional networks (FCNs) \cite{long2015fully} or dilated convolution models \cite{chen2018encoder}, together with the availability of large human-annotated datasets and powerful hardware led to exceptional results on challenging semantic segmentation benchmarks~\cite{long2015fully,chen2018encoder,zhao2017pyramid,zhang2018exfuse,zhang2018context,yuan2019object}.

Although deep learning models are approaching and even exceeding human-level performance on numerous tasks, including semantic segmentation, artificial neural networks encounter serious difficulties when it comes to incremental learning (IL). In other words, they struggle to preserve past knowledge when attempting to learn multiple tasks sequentially. This is due to the well known catastrophic forgetting issue~\cite{mccloskey1989catastrophic}, which is the tendency of a deep network to forget previously learned tasks. Unfortunately, this problem is intrinsic to the optimization techniques (\textit{e.g.}\ gradient descent) used to train neural networks. Nonetheless, the ability of learning a sequence of tasks is unarguably a highly desirable property of artificial intelligent systems. 

\begin{figure}[t]
\centering
\subfigure{
    \begin{minipage}{0.19\linewidth}
        \centering
        \includegraphics[width=0.995\textwidth,height=0.5in]{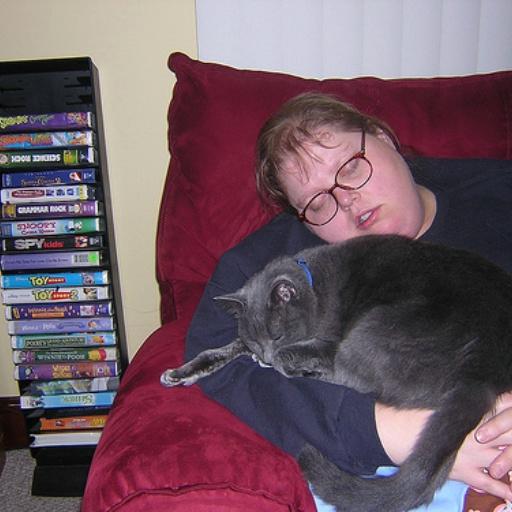}\\ \vspace{0.05cm}
        \includegraphics[width=0.995\textwidth,height=0.5in]{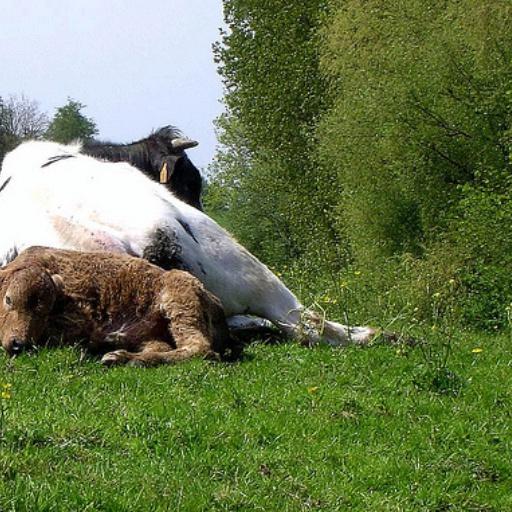}\\ \vspace{0.05cm}
        \includegraphics[width=0.995\textwidth,height=0.5in]{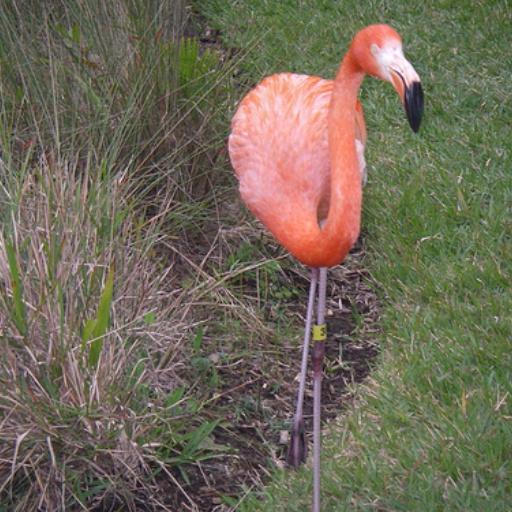}\\ \vspace{0.05cm}
        \includegraphics[width=0.995\textwidth,height=0.5in]{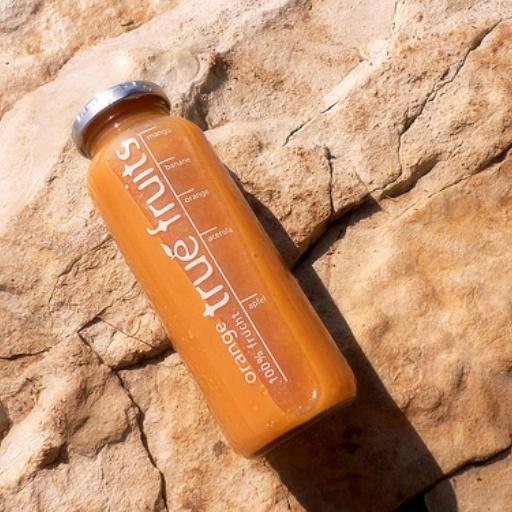}\\ \vspace{0.05cm}
        \includegraphics[width=0.995\textwidth,height=0.5in]{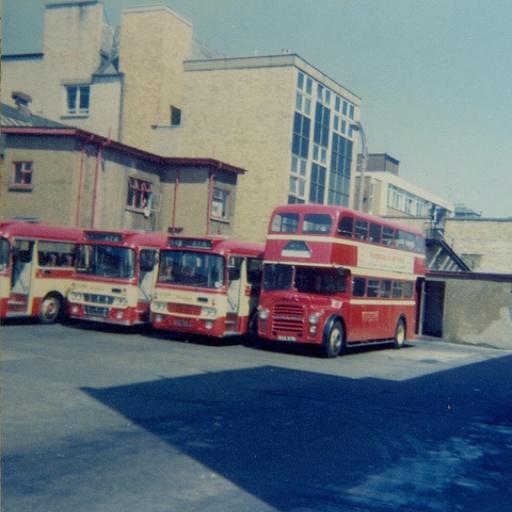}\\   
        \scriptsize{(a) Input Image}
    \end{minipage}%
}%
\subfigure{
    \begin{minipage}{0.19\linewidth}
        \centering
        \includegraphics[width=0.995\textwidth,height=0.5in]{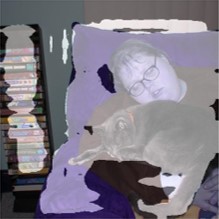}\\ \vspace{0.05cm}
        \includegraphics[width=0.995\textwidth,height=0.5in]{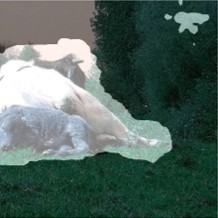}\\ \vspace{0.05cm}
        \includegraphics[width=0.995\textwidth,height=0.5in]{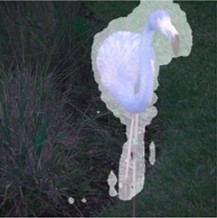}\\ \vspace{0.05cm}
        \includegraphics[width=0.995\textwidth,height=0.5in]{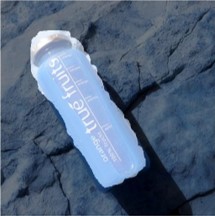}\\  \vspace{0.05cm}
        \includegraphics[width=0.995\textwidth,height=0.5in]{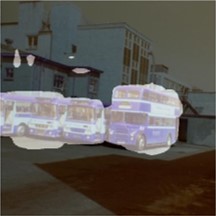}\\ 
        \scriptsize{(b) Attention}
    \end{minipage}%
}%
\subfigure{
    \begin{minipage}{0.19\linewidth}
        \centering
        \includegraphics[width=0.995\textwidth,height=0.5in]{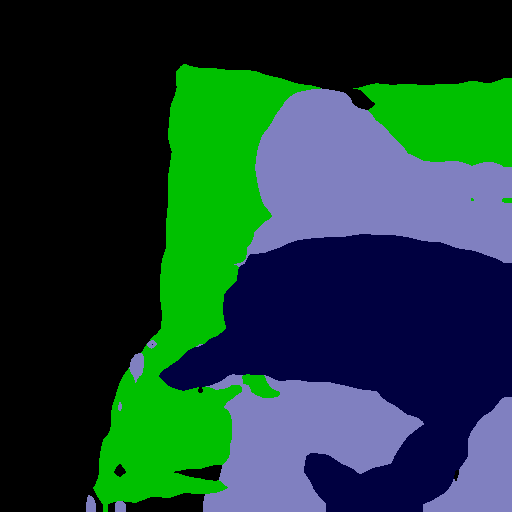}\\ \vspace{0.05cm}
        \includegraphics[width=0.995\textwidth,height=0.5in]{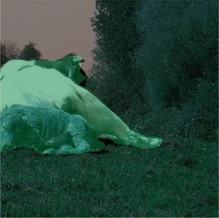}\\ \vspace{0.05cm}
        \includegraphics[width=0.995\textwidth,height=0.5in]{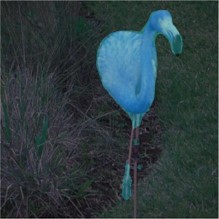}\\ \vspace{0.05cm}
        \includegraphics[width=0.995\textwidth,height=0.5in]{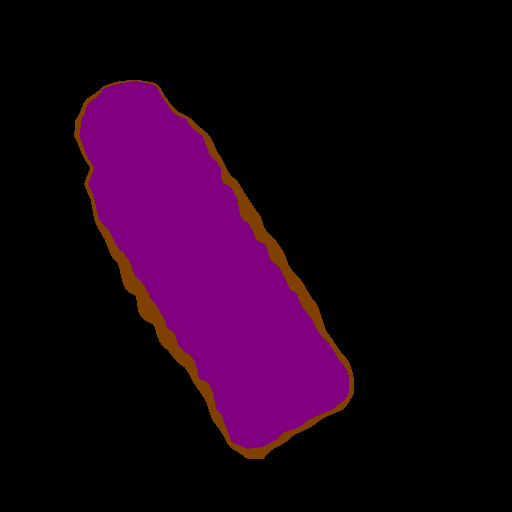}\\  \vspace{0.05cm}
        \includegraphics[width=0.995\textwidth,height=0.5in]{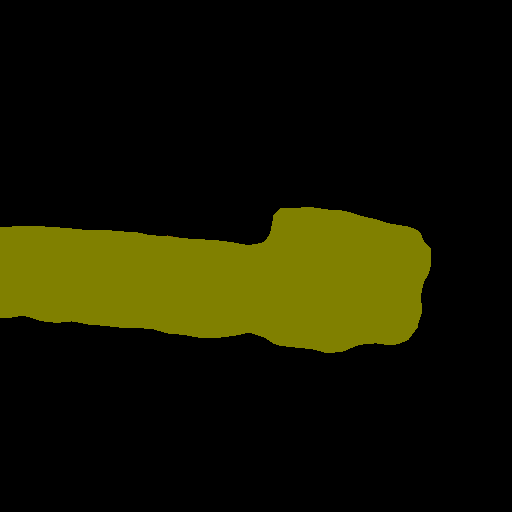}\\ 
        \scriptsize{(c) Ours}
    \end{minipage}%
}%
\subfigure{
    \begin{minipage}{0.19\linewidth}
        \centering
        \includegraphics[width=0.995\textwidth,height=0.5in]{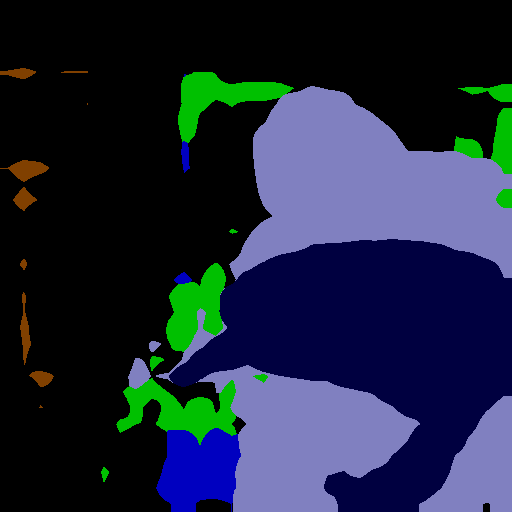}\\ \vspace{0.05cm}
        \includegraphics[width=0.995\textwidth,height=0.5in]{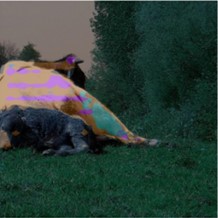}\\ \vspace{0.05cm}
        \includegraphics[width=0.995\textwidth,height=0.5in]{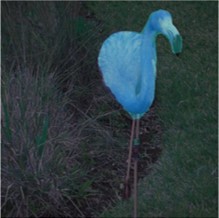}\\ \vspace{0.05cm}
        \includegraphics[width=0.995\textwidth,height=0.5in]{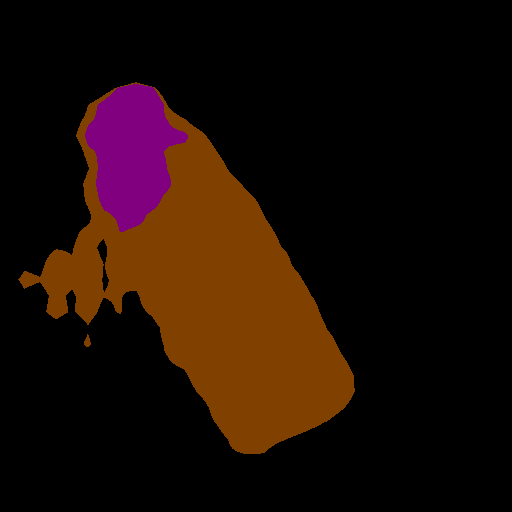}\\  \vspace{0.05cm}
        \includegraphics[width=0.995\textwidth,height=0.5in]{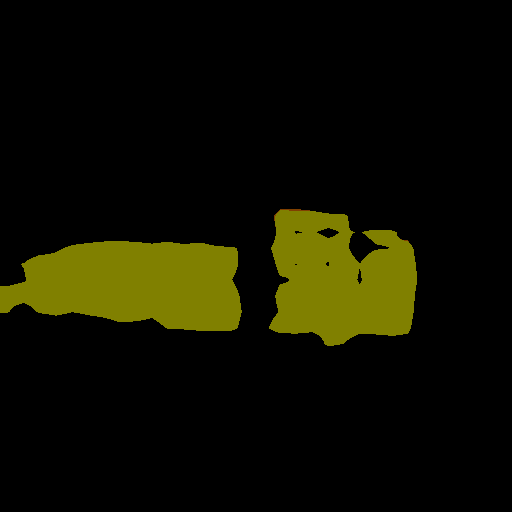}\\ 
        \scriptsize{(d) MiB \cite{cermelli2020modeling}}
    \end{minipage}%
}%
\subfigure{
    \begin{minipage}{0.19\linewidth}
        \centering
        \includegraphics[width=0.995\textwidth,height=0.5in]{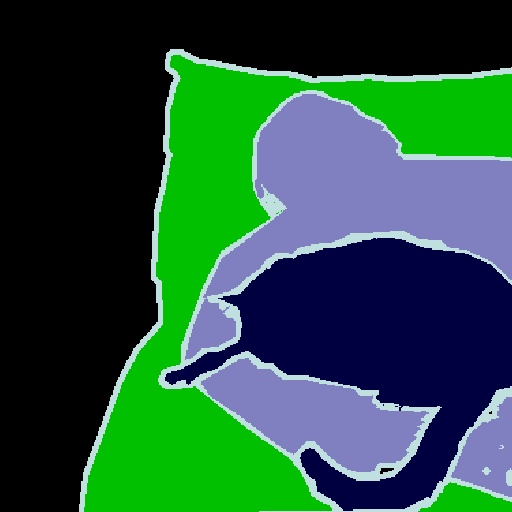}\\ \vspace{0.05cm}
        \includegraphics[width=0.995\textwidth,height=0.5in]{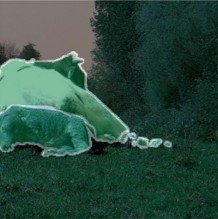}\\ \vspace{0.05cm}
        \includegraphics[width=0.995\textwidth,height=0.5in]{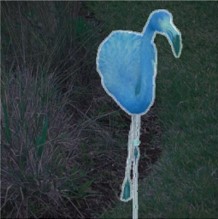}\\ \vspace{0.05cm}
        \includegraphics[width=0.995\textwidth,height=0.5in]{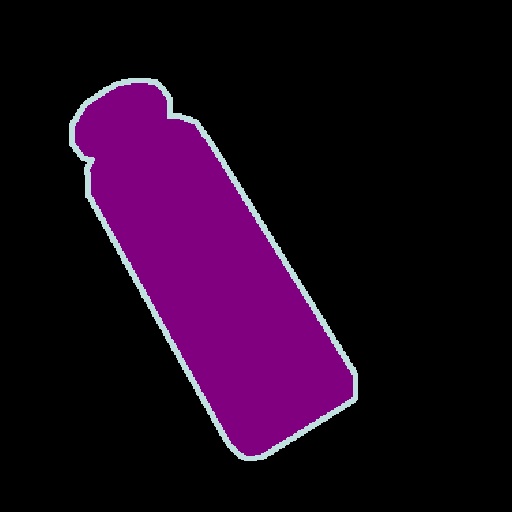}\\  \vspace{0.05cm}
        \includegraphics[width=0.995\textwidth,height=0.5in]{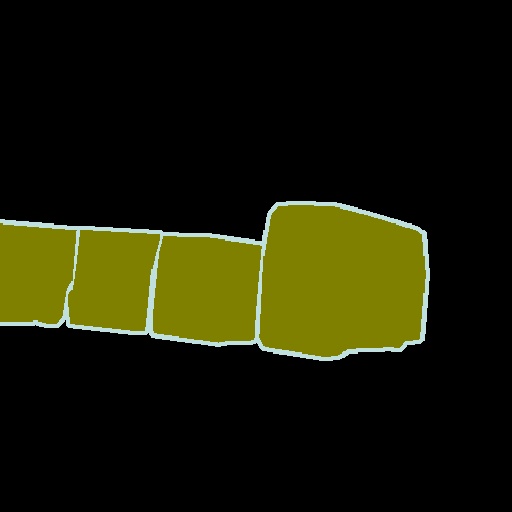}\\ 
        \scriptsize{(e) GT}
    \end{minipage}%
}%
\caption{Given an input image, by leveraging from attention maps (b) computed with the proposed continual attentive fusion (CAF) module, our method produces segmentation maps (c) more similar to ground truths (e) than those computed with previous methods such as MiB \cite{cermelli2020modeling} in (d). 
}
\label{fig:vis_qua_att}
\end{figure}

For this reason, IL has recently gained much attention, and in particular, incremental class learning (ICL), where the ability of a model to discriminate current and past classes simultaneously without knowing the task at test time is evaluated. Over the past years, several works have proposed ICL approaches, 
mostly focusing on image classification~\cite{li2017learning,kirkpatrick2017overcoming,chaudhry2018riemannian,rebuffi2017icarl,hou2019learning,douillard2020podnet,iscen2020memory,fini2020online}, object detection~\cite{shmelkov2017incremental,liu2020multitask}, image retrieval~\cite{chen2021feature,tian2020complementary} and emotion
recognition~\cite{thuseethan2021deep,zhang2020weakly,fujii2020hierarchical} . So far, much less attention has been devoted to ICL for semantic segmentation. Recently, Cermelli \textit{et al.}~\cite{cermelli2020modeling} showed promising results introducing a distillation-based framework that accounts for background distribution shifts between tasks to prevent biased predictions. Concurrently, recent studies demonstrated the effectiveness of attention mechanisms on semantic segmentation~\cite{chen2016attention,xu2017learningdeep,fu2019dual,zhan2018unsupervised}. Although attempts were made to exploit attention for learning tasks sequentially in the context of object classification~\cite{dhar2019learning} and detection~\cite{liu2020multitask}, they cannot be effectively adopted for incremental semantic segmentation. In this paper, we bridge this gap and propose the first attention-based ICL method for semantic segmentation. 

Our contributions can be summarized as follows:
\begin{itemize}
    \item We propose a novel deep architecture for ICL which embeds a new continual attentive fusion (CAF) module. 
    Given the features of the current and previous models, CAF produces structured self-attention weights to update the current features. We demonstrate that this module by itself reduces catastrophic forgetting significantly and leads to improved segmentation maps with respect to previous methods such as MiB \cite{cermelli2020modeling} (see Figure \ref{fig:vis_qua_att}).
    \item We introduce a novel attentive distillation loss that leverages both channel-wise and spatial attention to transfer more relevant knowledge into the current model. By correctly weighting the importance of each channel, the attentive distillation loss enables the model to keep the performance on old tasks without using any explicit information from the past.
    \item We devise a simple and effective method for improving the balance between old classes and background probabilities that only depends on the inferred ratio of old and new classes in a given image, thus avoiding introducing additional hyper-parameters.
    \item The proposed approach sets the new state-of-the-art on two common datasets for ICL for semantic segmentation, namely Pascal-VOC 2012 \cite{everingham2007pascal} and ADE20K \cite{zhou2017scene}. 
\end{itemize}


\begin{figure*}[t]
    \centering
    \includegraphics[width=1\textwidth]{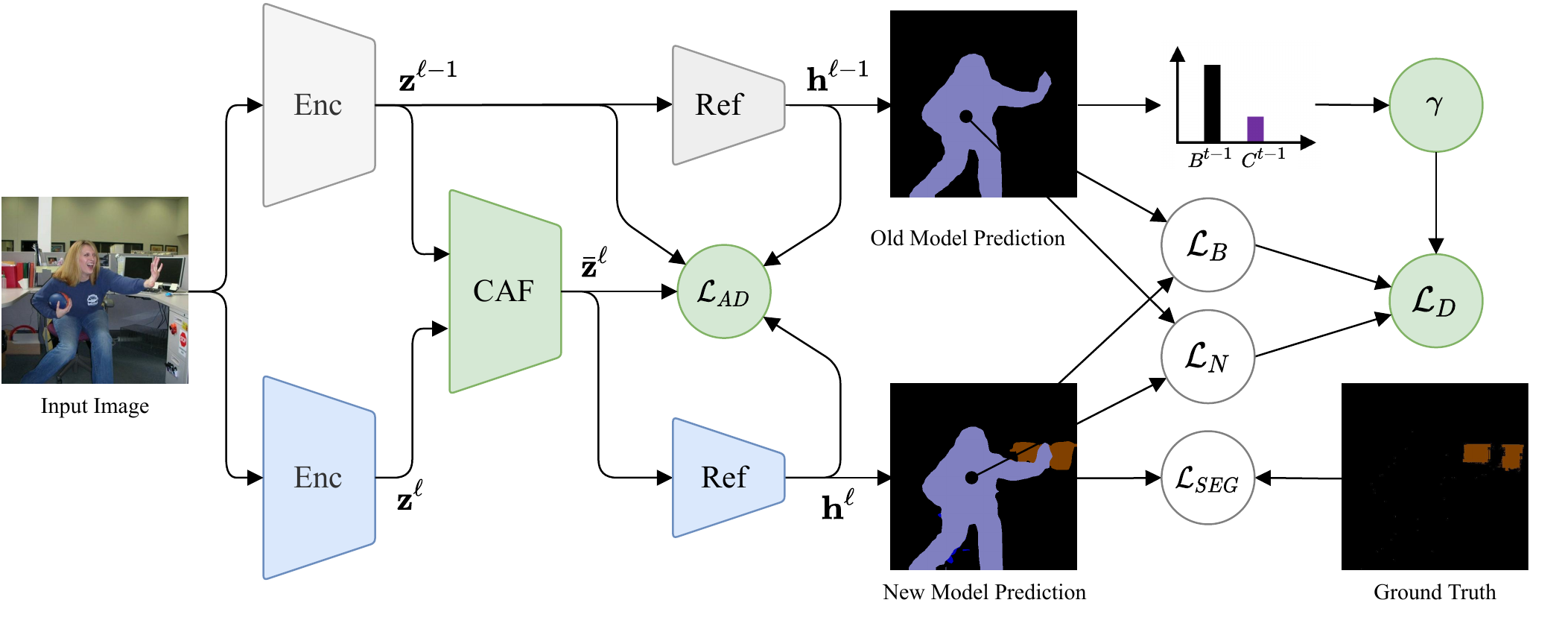}
    \caption{The overall scheme of our approach at incremental learning step $\ell$. The grey modules correspond to the network learned at the previous incremental learning step, which is frozen at step $\ell$ while the blue modules correspond to the network trained at step $\ell$.
    Our three contributions, namely the continual attentive fusion (CAF), the attentive distillation loss ($\mathcal{L}_{\operatorname{AD}}$) and the balanced knowledge distillation ($\mathcal{L}_{\operatorname{D}}$), are highlighted in green. }
    \label{fig:overall}
\end{figure*}
\section{Related Works}
\label{sec:related}

\noindent \textbf{Incremental Learning.}
Modern artificial neural networks are haunted by the well known catastrophic forgetting problem: the tendency of neural models to severely degrade performance on previous tasks when training on new ones. This issue has been largely studied in the literature in the last few decades~\cite{mccloskey1989catastrophic}, leading to a wide variety of IL approaches.
According to \cite{de2019continual}, prior art in this field can be organized in three categories: replay-based methods~\cite{rebuffi2017icarl,castro2018end,shin2017continual,hou2019learning,ostapenko2019learning,wu2018memory}, regularization-based techniques~\cite{kirkpatrick2017overcoming,chaudhry2018riemannian,zenke2017continual,li2017learning,dhar2019learning} and parameter isolation-based approaches~\cite{mallya2018packnet,mallya2018piggyback}. Replay-based methods consist in storing~\cite{rebuffi2017icarl,castro2018end,hou2019learning,wu2019large} or generating~\cite{shin2017continual,wu2018memory,ostapenko2019learning} examples of the first task which are then reused in subsequent learning stages. Regularization-based methods are either penalizing changes of a subset of parameters while learning on new tasks~\cite{zenke2017continual,chaudhry2018riemannian,kirkpatrick2017overcoming,aljundi2018memory} or employing distillation to force the network not to forget past knowledge~\cite{li2017learning,dhar2019learning,fini2020online}. Parameter-isolation based approaches are built on the idea of having task-specific set of parameters. 
Despite the interest in the problem, the large majority of the literature focuses on classification. A pioneering work of Shmelkov et al.~\cite{shmelkov2017incremental} exploits distillation~\cite{li2017learning} for class discovery in detection. In this paper, we address ICL in semantic segmentation.

\noindent \textbf{Incremental Learning in Semantic Segmentation.} Deep learning brought great progress in semantic segmentation~\cite{long2015fully,chen2018encoder,lin2017refinenet,zhang2018exfuse,zhao2017pyramid}.
Despite the abundant literature on this task, very few works are tackling ICL for this task~\cite{michieli2019incremental,tasar2019incremental,ozdemir2018learn,ozdemir2019extending,cermelli2020modeling}. Moreover, these studies address the problem from different perspectives and utilize contrasting experimental settings. For instance, in~\cite{michieli2019incremental}, the authors were the first to study this task proposing an approach which operates both on the output and on the intermediate representations of the segmentation model. 
\cite{ozdemir2018learn} presented a method to sample prototypical examples of the old classes to be used as a rehearsal in the new task. However, both \cite{michieli2019incremental,ozdemir2018learn} assume that some information from the previous task will be available during the training of the second task. Other approaches \cite{tasar2019incremental,ozdemir2019extending} described ICL methods which are specialized for certain subfields (\textit{i.e.} remote sensing and computer assisted radiology and surgery), lacking generality. More recently, \cite{cermelli2020modeling} attempted to fix the semantic distribution shift in the background class, showing significant performance boost. However, their approach operate at loss level, while network architectural changes to improve segmentation maps are not considered. 
In contrast, in this paper we overcome several limitations of the previous literature, and propose a new architectural solution for rehearsal-free ICL in semantic segmentation. 

\noindent \textbf{Attention Mechanisms.}
Several works considered attention models within deep architectures to improve performance~\cite{tang2020xinggan,xiao2015application,tang2020dual,xu2015show,duan2021audio,tang2019multi,chorowski2015attention,tang2021attentiongan,ding2020lanet,duan2021cascade,yang2021transformers,li2021multi,xu2018structured}. 
Focusing only on pixel-wise dense prediction, Chen \textit{et al.}~\cite{chen2016attention} first described an attention model to combine multi-scale features learned by a FCN for semantic segmentation. Zhang \textit{et al.}~\cite{zhang2018context} designed EncNet, a network equipped with a channel attention mechanism to model global context. Similarly, Zhao \textit{et al.}~\cite{zhao2018psanet} proposed to account for pixel-wise dependencies introducing relative position information across the spatial dimension within the convolutional layers. Other works~\cite{fu2019dual} introduced attention to model contextual and semantic dependencies, respectively. Zhong \textit{et al.}~\cite{zhong2020squeeze} considered spatial and channel inter-dependencies in their squeeze-and-attention network. Xu \textit{et al.}~\cite{xu2017learningdeep,xu2017multi} described attention gates, introduced to control the message passing among variables, thus integrating attention into a probabilistic model formulation.
Our work differs significantly from previous art, since (i) we use spatial and channel-wise attention to help the network discovering new classes and (ii) attention is also counteract the semantic distribution shift of the background class between the two tasks.

\newcommand{\image}{\mathbf{x}}
\newcommand{\seen}{\mathcal{S}}
\newcommand{\unseen}{\mathcal{U}}
\newcommand{\features}{\mathbf{z}}
\newcommand{\featuresbar}{\bar{\mathbf{z}}}
\newcommand{\nonlocalfeatures}{\mathbf{v}}
\newcommand{\hidden}{\mathbf{h}}


\section{Proposed Method}
The problem of image segmentation is that of inferring the correct label $s_p$ for each pixel $p$ in the input image $\image$, among the available class labels $\seen$. For the sake of simplicity all images are assumed to be of the same size $|\image|=P$. Tools for semantic segmentation can vary significantly in nature, but we will consider the general case of function $\phi_\omega:\mathcal{X}\rightarrow\mathbb{R}^{|\seen|\times P}$ conceived to predict a probability distribution over the class labels for each pixel of the input image, where $\mathcal{X}$ is the input image space and $\omega$ are the parameters of this function. More precisely, $\phi_\omega(\image)[w,h,s]$ is supposed to represent the probability of the pixel at position $(w,h)$ in image $\mathbf{x}$ belonging to class $s\in\seen$. Concretely, the function approximator $\phi_\omega$ can be further divided into three sub-networks (see Figure~\ref{fig:overall}): (i) an encoder realized by a residual neural network 
that extracts a feature map $\features$; (ii) a refinement model that projects the features into a refined version $\hidden$; and (iii) a classifier that produces a probability distribution $p$ over the class labels. The set of parameters can be learned with the help of a training set $\mathcal{D}\subset \mathcal{X}\times \seen^P$. 

Traditionally, semantic segmentation considers a fixed set of classes $\seen$. More recently, the community started looking at learning these classes in a \textit{continuous} setup, \textit{i.e.} considering a series of \textit{incremental learning steps}, indexed by $\ell$. At each learning step, an extra set of categories is added to the semantic segmentation task. In the following, we denote by $\seen_{\ell-1}$ the set of classes learned --seen-- before learning step $\ell$, and by $\unseen_\ell$ the set of new --unseen-- classes added at learning step $\ell$. The following holds: $\seen_\ell=\unseen_\ell\cup\seen_{\ell-1}$ and $\unseen_\ell\cap\seen_{\ell-1}=\emptyset$. This means that at every learning step, the size of the output classifier increases with respect to the previous learning step. Consequently, the training set at step $\ell$ will be denoted by $\mathcal{D}_\ell\subset \mathcal{X}\times \unseen_\ell^P$ and the inference function by $\phi^\ell_\omega:\mathcal{X}\rightarrow\mathbb{R}^{|\seen_\ell|\times P}$. The index $\ell$ also applies to the various features extracted by our architecture in Figure~\ref{fig:overall}, namely the convolutional features $\features^\ell$ and the refined features $\hidden^\ell$.

The methodological contributions of the proposed method are three. First, a continual attentive fusion module exploiting the convolutional features of both tasks, $\features^{\ell-1}$ and $\features^\ell$, to compute a structured attention tensor used to transform $\features^\ell$ taking $\features^{\ell-1}$ into account (see Section~\ref{sec:x-att}). At test time, where $\features^{\ell-1}$ are unavailable, several strategies are evaluated in the experiments. Second, a self-attention feature distillation loss that leverages both channel-wise and spatial attention to transfer more relevant knowledge into the current step. The loss acts on both the convolutional $\features$ and the refined $\hidden$ features (see Section~\ref{sec:afdloss}). Third, a balanced knowledge distillation loss that account for the overpresence of the background class. Indeed, when comparing old and new segmentation maps, the new classes must be merged with the background, thus overestimating the amount of background pixels. We propose a simple yet very effective method to rebalance the learning in Section~\ref{sec:bkdist}.
In the following, we describe in details our three contributions.

\subsection{Continual Attentive Fusion}
\label{sec:x-att}

\begin{figure*}[h]
    \centering
    \includegraphics[width=1\textwidth]{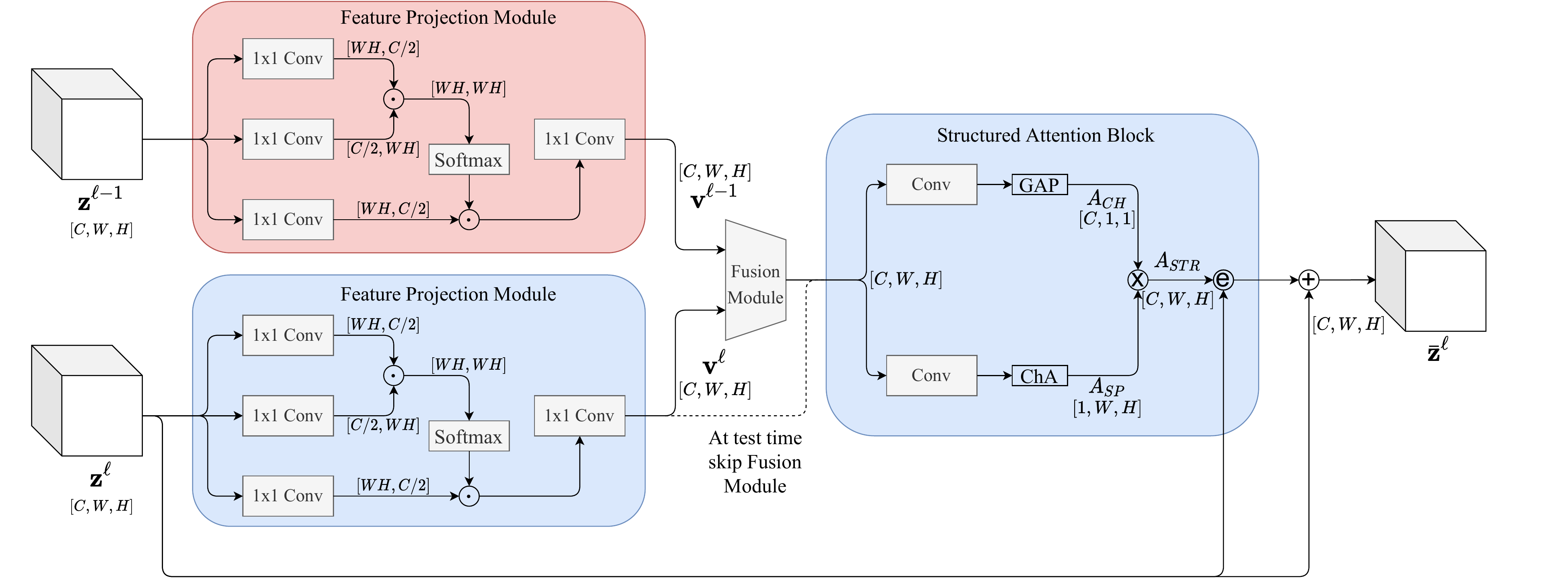}
    \caption{A detailed view of the CAF module. Non-local features $\nonlocalfeatures^{\ell-1}$ and $\nonlocalfeatures^{\ell}$ are computed independently from the features of the two learning steps $\features^{\ell-1}$ and $\features^{\ell}$ via a series of $1\times 1$ convolutions and matrix multiplications $\odot$. After a fusion module, a channel attention vector and a spatial attention map are computed then aggregated via a tensor product $\otimes$ into a structured attention tensor. The element-wise product (circled $\text{e}$) is used to apply the attention to the original features. A residual connection provides the updated feature map $\featuresbar^\ell$. The whole structure represents the architecture at training time while red background parts and the fusion module are discarded at test time. }
    \label{fig:overall_cm}
\end{figure*}

The continual attentive fusion (CAF) module is specifically conceived to compute self-attention from the features of the current IL step $\ell$ as well as from the previous one $\ell-1$. In practice, it is composed of two main blocks, with the purpose of (i) computing features independently for $\features^\ell$ and $\features^{\ell-1}$ and (ii) fusing these features into a structured attention tensor. The diagram of CAF is shown in Figure~\ref{fig:overall_cm}.

In order to compute the features to be fused, denoted by $\nonlocalfeatures^\ell$ and $\nonlocalfeatures^{\ell-1}$, 
we draw inspiration from non-local neural networks~\cite{wang2018non}.
Each input feature map $\features$ is fed to three different $1\times 1$ convolutional layers. The output of the first two convolutions is used to obtain a non-local self-attention tensor, which is then used to weight the output of the third convolution, thus obtaining $\nonlocalfeatures$. The non-local features corresponding to $\features^{\ell-1}$ and $\features^\ell$ are denoted by $\nonlocalfeatures^{\ell-1}$ and $\nonlocalfeatures^\ell$ respectively, and correspond to the upper and bottom non-local blocks of Figure~\ref{fig:overall_cm}.
The {projected} features corresponding to $\features^{\ell-1}$ and $\features^\ell$ are denoted by $\nonlocalfeatures^{\ell-1}$ and $\nonlocalfeatures^\ell$ respectively, and correspond to the upper and bottom feature refinement blocks in Figure~\ref{fig:overall_cm}.

As stated above, the {projected} features are aggregated to compute a structured attention tensor. To do so, we employ a fusion module (see Figure~\ref{fig:overall_cm}) that first concatenates the {projected} features along the channel dimension and then feeds them to a convolutional layer to reduce the number of channels by two, thus back to the original amount. The output of the fusion module is a tensor $\nonlocalfeatures^{(\ell-1,\ell)}$ containing the information coming from both models. Note that this module is used at training time, but can be skipped at test time if $\features^{\ell-1}$ is not computed (\textit{e.g.}\ to reduce resource usage).

The fusion tensor, $\nonlocalfeatures^{(\ell-1,\ell)}$ is fed to two convolutional layers so as to extract a channel-wise attention vector and a spatial attention map, via global average pooling (GAP) and channel-wise average (ChA) operations respectively:
\begin{equation}
  \boldsymbol{A}_{\textsc{sp}}^\ell[w,h] = \frac{1}{C}\sum_{c=1}^{C} (\omega_{\textsc{sp}}*\nonlocalfeatures^{(\ell-1,\ell)})[c,w,h],
\end{equation}
\begin{equation}
  \boldsymbol{A}_{\textsc{ch}}^\ell[c] = \frac{1}{WH}\sum_{h,w=1}^{H,W} (\omega_{\textsc{ch}}*\nonlocalfeatures^{(\ell-1,\ell)})[c,w,h],
\end{equation}
where $\omega_{\textsc{sp}}$ and $\omega_{\textsc{ch}}$ are the weights of the convolutions. The attention vector and spatial map are used to construct a structured attention tensor via a tensor product:
\begin{equation}
    \boldsymbol{A}_{\textsc{str}} = \boldsymbol{A}_{\textsc{ch}}\otimes \boldsymbol{A}_{\textsc{sp}},
\end{equation}
which is finally used, together with a residual connection, to obtain the updated feature map $\featuresbar^{\ell}$:
\begin{equation}
    \featuresbar^{\ell} = (\boldsymbol{I}+\boldsymbol{A}_{\textsc{str}})\,\nonlocalfeatures^{(\ell-1,\ell)}= (\boldsymbol{I}+\boldsymbol{A}_{\textsc{ch}}\otimes \boldsymbol{A}_{\textsc{sp}})\,\nonlocalfeatures^{(\ell-1,\ell)}.
\end{equation}
In this way, the new features $\featuresbar^{\ell}$ are computed via a structured self-attention tensor, which at its turn is computed from an aggregation of {projected} features from both the current and the previous learning steps $\features^\ell$ and $\features^{\ell-1}$. Alternatives to this fusion step that avoid using $\features^{\ell-1}$ at test time will be discussed in the experimental section.

\subsection{Attentive Feature Distillation} \label{sec:afd}
\label{sec:afdloss}
Many existing methods apply feature distillation to preserve previous knowledge~\cite{li2017learning,dhar2019learning,liu2020multitask,tasar2019incremental}.
Most of them treat the channels in the feature map equally, \textit{i.e.} channels are weighted uniformly in the distillation loss. However, the features tend to drift to a new configuration to discriminate the classes of the task at hand, regardless of the type of precautions employed to avoid forgetting. Hopefully, some portion of the features will change substantially to adapt to the new task, while most of them will remain reasonably close to their previous configuration. This undermines the assumption that all the channels should be treated equally. We conjecture this as a main limitation of previous works in ICL for semantic segmentation. To overcome this issue, we employ the squeeze-and-excitation (SE) module~\cite{hu2018squeeze} to generate channel-wise attention as follows:
\begin{equation}
\label{eq:att ch}
    \operatorname{AD}_{\textsc{ch}}\left(\mathbf{m}\right)=\psi\left(\omega_{2}^{\mathbf{m}}*\sigma\left(\omega_{1}^{\mathbf{m}}*\operatorname{AvgPool}(\mathbf{m})\right)\right),
\end{equation}
where $\psi(\cdot)$ and $\sigma(\cdot)$ represent the Sigmoid  and ReLU activation functions, and $\mathbf{m}$ is a generic feature map we use here as a placeholder. Note the superscript in the weight matrices $\omega_{1}^{\mathbf{m}}$ and $\omega_{2}^{\mathbf{m}}$, meaning that those weights are specific to the feature map input to $\operatorname{AD}_{\textsc{ch}}\left(\cdot\right)$.  For a generic feature map $\mathbf{m}$ of dimensions $[C,W,H]$, $\operatorname{AD}_{\textsc{ch}}\left(\cdot\right)$ will be of size $[C,1,1]$.

Similarly, since the background is very complex and can excite features of other classes, the context information is crucial for semantic segmentation. We would like to leverage the inferred probability distributions of the classes in the background to improve the distillation process. However, as for the channels, it might be appropriate to let the network decide which parts of the background are more important to distill. Hence, we utilize a self spatial attention:
\begin{equation}
\label{eq:att sp}
    \operatorname{AD}_{\textsc{sp}}\left(\textbf{m}\right)=\frac{\sum_{j=1}^{C}\textbf{m}_j^2}{\left\|\sum_{j=1}^{C}\textbf{m}_j^2\right\|_{\operatorname{F}}}.
\end{equation}
The size of $\operatorname{AD}_{\textsc{sp}}$ will be $[1,W,H]$. Spatial and channel-wise attention are combined through a tensor product:
\begin{equation}
    \operatorname{AD}\left(\mathbf{m}\right) = \left(\operatorname{AD}_{\textsc{ch}}\left(\mathbf{m}\right)\otimes\operatorname{AD}_{\textsc{sp}}\left(\mathbf{m}\right)+1\right)\,\mathbf{m},
\end{equation}
where the second product is element-wise between two tensors of the same size. Overall, $\operatorname{AD}\left(\mathbf{m}\right)$ is weighting the original tensor with a structured and learned self-attention tensor. In our framework, we use this formulation to distill knowledge from the previous task into the new one, therefore defining a structured self-attention distillation loss:
\begin{equation}
\label{eq:loss ad}
    \mathcal{L}_{\textsc{AD}}=\lVert\operatorname{AD}(\featuresbar^\ell)-\operatorname{AD}(\features^{\ell-1})\rVert_{\operatorname{F}}^2 + \Vert\operatorname{AD}(\hidden^\ell)-\operatorname{AD}(\hidden^{\ell-1})\rVert_{\operatorname{F}}^2.
\end{equation}
Very importantly, $\mathcal{L}_{\textsc{AD}}$ is applied to both features $\features$ and $\hidden$ of both the old and new incremental learning steps $\ell-1$ and $\ell$. 

\subsection{Balanced Knowledge Distillation}
\label{sec:bkdist}
As mentioned, in the context of incremental learning in semantic segmentation, distillation~\cite{hinton2015distilling} plays a core role in transferring knowledge from the old model into the new one, mitigating catastrophic forgetting. A typical definition for the distillation loss is:
\begin{equation}
\label{eq:kd old}
    \mathcal{L}_{\textsc{UD}}=\beta \sum_{w,h=0}^{W,H}\sum_{s=0}^{|\seen_{\ell-1}|}\phi_\omega^{\ell-1}(\image)[w,h,s]\,\log \phi_\omega^\ell(\image)[w,h,s],
\end{equation}
where $\phi_\omega^{\ell}(\image)[w,h,s]$ and $\phi_\omega^{\ell-1}(\image)[w,h,s]$ are the probabilities of a pixel at position $(h,w)$ to belong to class $s$ as inferred by the new and old model respectively, while $\beta=-\frac{1}{HW}$ is a normalization factor.

Assuming that the background class was part of the previous learning step, $\textsc{B}_{\ell-1}\in\seen_{\ell-1}$, we can decompose the previous loss into two contributions: from the background $\mathcal{L}_{\textsc{B}}$ and from the other classes $\mathcal{L}_{\textsc{N}}$. During the learning of the previous step $\ell-1$, the pixels belonging to the classes unknown at step $\ell-1$ but known at step $\ell$, that is $\unseen_{\ell}$, were assigned to the background class. This leads to an imbalance in $\mathcal{L}_{\textsc{B}}$, due to the fact that $\phi_\omega^{\ell-1}$ is not aware of the new classes. In order to address this issue, \cite{cermelli2020modeling} rewrites $\phi_\omega^\ell$ as:
\begin{equation} \label{eq:kd mib} \small
    \hat{\phi}_\omega^\ell(\image)[h,w,s]=\left \{\begin{array}{lc}
         \phi_\omega^\ell(\image)[h,w,s] & s \neq \textsc{B}_{\ell-1}, \\
        \sum_{s'\in\unseen_\ell\cup\{\textsc{B}_\ell\}} \phi_\omega^\ell(\image)[h,w,s'] & s= \textsc{B}_{\ell-1},
    \end{array} \right.
\end{equation}
thus aggregating the new background class $\textsc{B}_\ell$ together with the unseen classes $\unseen_\ell$ to emulate the background class at the previous learning step $\textsc{B}_{\ell-1}$. The opposite also holds, since we suppose that annotations for classes of previous time steps are not available, and therefore belong to the background class at step $\ell$. This has no impact in the distillation loss, but in the supervised segmentation loss in~(\ref{eq:overall loss}). 

\begin{figure}[t]
\centering
\subfigure[{Disjoint}]{
    \begin{minipage}{0.8\linewidth}
        \centering
        \includegraphics[width=1\textwidth]{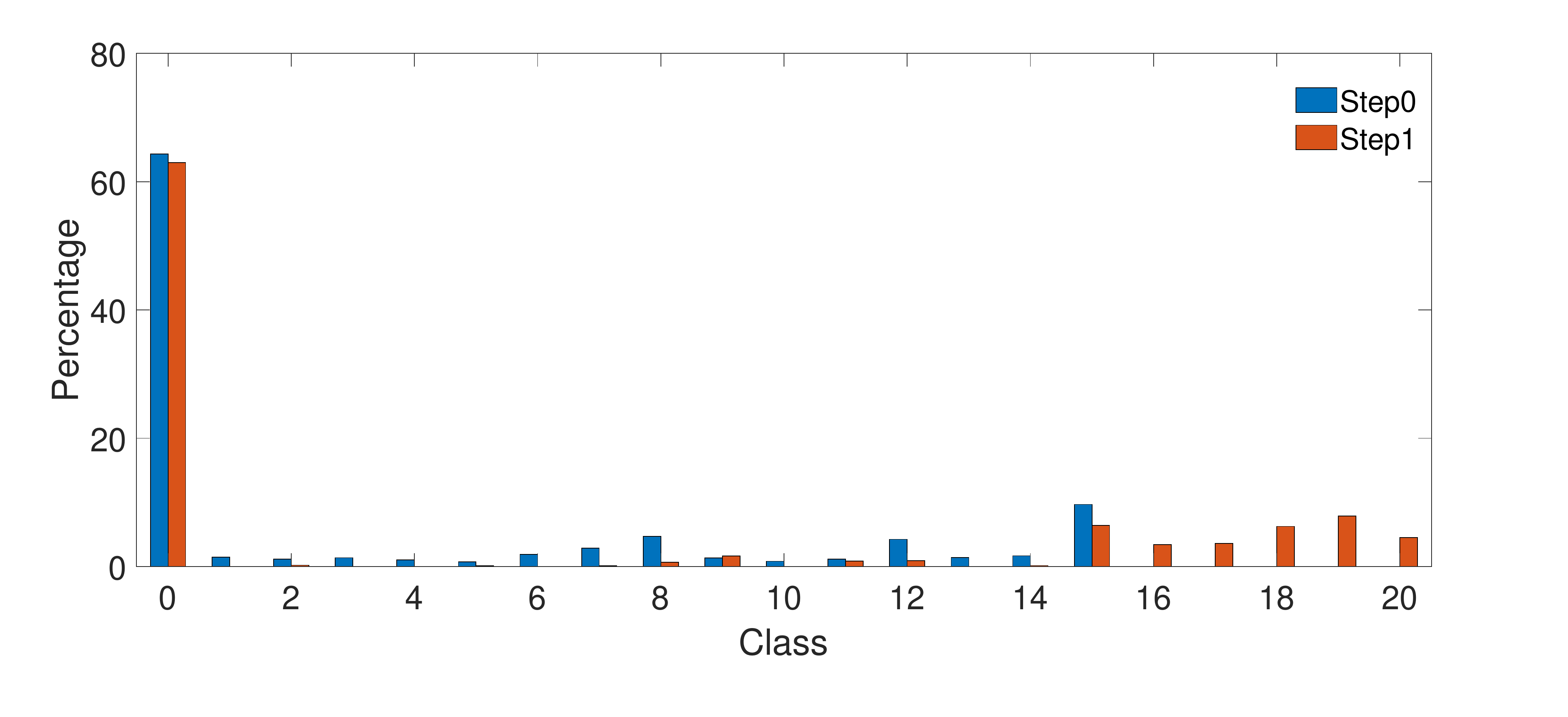}\\
    \end{minipage}%
}\\
\subfigure[{Overlapped}]{
    \begin{minipage}{0.8\linewidth}
        \centering
        \includegraphics[width=1\textwidth]{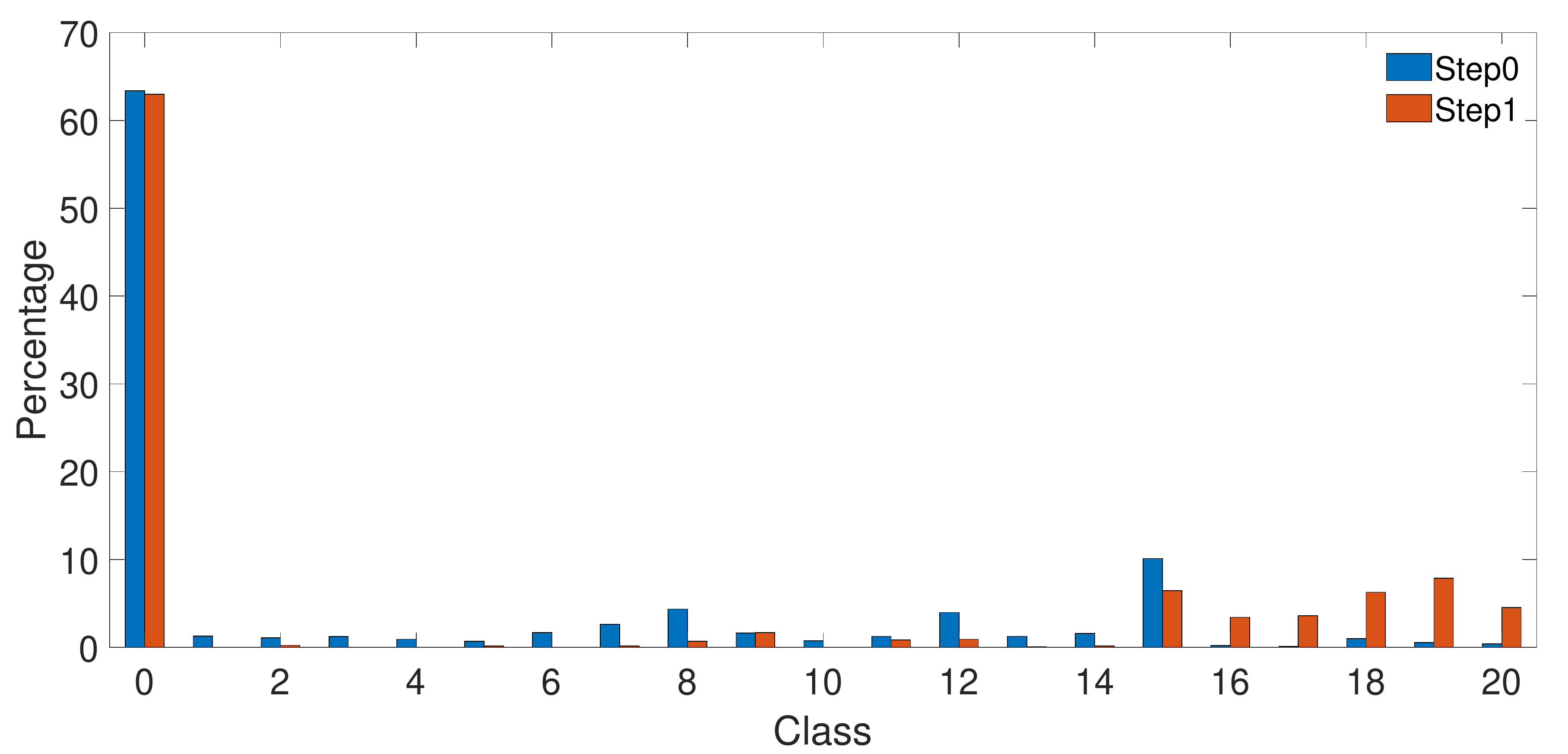}\\
    \end{minipage}%
}%
\caption{The statistical distribution of different classes under different steps in VOC 2012 15-5 disjoint setting and overlapped setting. }
\label{fig:distribution}
\end{figure}

According to Figure~\ref{fig:distribution}, even if the new $\hat{\phi}_\omega^\ell$ accounts for the imbalance between the old and new probability distributions for distillation, the background is usually the most represented class by far. Thus, $\mathcal{L}_{\textsc{B}}$ has a much stronger contribution to the overall loss as compared to $\mathcal{L}_{\textsc{N}}$. As a consequence, the network basically ignores the information of old classes preventing effective knowledge distillation. To overcome this issue, we propose to introduce a balancing parameter $\gamma$ to re-weight the influence between $\mathcal{L}_{\textsc{B}}$ and $\mathcal{L}_{\textsc{N}}$ in the distillation loss. Formally, the expression of $\gamma$ writes as:
\begin{equation}
\label{eq:gamma}
\!\!\gamma=\frac{\sum_{s\in \seen_{t-1}/\{\textsc{B}_{\ell-1}\}} \operatorname{Softmax}\left(\operatorname{AvgPool}\left(\phi_\omega^{\ell-1}(\image\right)[s])\right)}{\operatorname{Softmax}\left(\operatorname{AvgPool}\left(\phi_\omega^{\ell-1}\left(\image\right)[\textsc{B}_{\ell-1}]\right)\right)},
\end{equation}
and allows us to step further than what was proposed in~\cite{cermelli2020modeling}, and weight the distillation as follows:
\begin{equation}
    \label{eq:bkd}
    \mathcal{L}_D=\gamma\mathcal{L}_B+\mathcal{L}_{N},
\end{equation}
where $\mathcal{L}_B$ and $\mathcal{L}_{N}$ and the background and non-background contributions to the unweighted distillation loss in~(\ref{eq:kd old}).

\subsection{Overall Loss}
The two losses described in the previous sections are used, together with the standard supervised loss for semantic segmentation, to train the overall architecture. In more detail, the final loss is expressed as:
\begin{equation}\label{eq:overall loss}
    \mathcal{L}=\mathcal{L}_{\textsc{SEG}}+\lambda_{\textsc{AD}}\mathcal{L}_{\textsc{AD}}+\lambda_{\textsc{D}}\mathcal{L}_{\textsc{D}},
\end{equation}
where $\lambda_{\textsc{AD}}$ and  $\lambda_{\textsc{D}}$ are the weights of the attention distillation and knowledge  distillation losses defined in Sections~\ref{sec:afdloss} and~\ref{sec:bkdist}, respectively, and $\mathcal{L}_{\textsc{SEG}}$ is the supervised segmentation loss (pixel-wise cross-entropy) previously defined as follows:

\begin{equation}
    \mathcal{L}_{\textsc{SEG}}=-\frac{1}{HW}\sum_{w,h=0}^{W,H}\sum_{s=0}^{|\seen_{\ell}|}\log \tilde{\phi}_\omega^\ell(\image)[w,h,s],
\end{equation}
where:
\begin{equation}
\small
    \tilde{\phi}_\omega^\ell(\image)[h,w,s]=\left \{\begin{array}{lc}
         \phi_\omega^\ell(\image)[h,w,s] & s \neq \textsc{B}_{\ell}, \\
        \sum_{s'\in\seen_{\ell-1}\cup\{\textsc{B}_\ell\}} \phi_\omega^\ell(\image)[h,w,s'] & s= \textsc{B}_{\ell}.
    \end{array} \right.
\end{equation}
Notice that the output probabilities $\tilde{\phi}$ defined here for the segmentation loss are different from the output probabilities $\hat{\phi}$ defined in the main paper for the background distillation loss. Indeed, while $\tilde{\phi}$ aggregates the previous classes to the current background class (so that all previous classes become background for the segmentation loss), the output probabilities defined in the paper $\hat{\phi}$ aggregate the new classes to the current background, because the background at the previous incremental step includes the new classes.
\begin{table*}[!t]
\caption{Mean IoU on the Pascal-VOC 2012 dataset for different incremental class learning scenarios.$*$ means results come from re-implementation.}
\vspace{-3mm}

\label{tab:overall voc}
\resizebox{\textwidth}{!}{%
\begin{tabular}{ccccccccccccccccccc}
\toprule
\multirow{3.5}{*}{Method} & \multicolumn{6}{c|}{19-1} & \multicolumn{6}{c|}{15-5} & \multicolumn{6}{c}{15-1} \\\cmidrule{2-19}
& \multicolumn{3}{c|}{Disjoint} & \multicolumn{3}{c|}{Overlapped} & \multicolumn{3}{c|}{Disjoint} & \multicolumn{3}{c|}{Overlapped} & \multicolumn{3}{c|}{Disjoint} & \multicolumn{3}{c}{Overlapped} \\\cmidrule{2-19}
& 1-19 & \multicolumn{1}{c|}{20} & \multicolumn{1}{c|}{all} & 1-19 & \multicolumn{1}{c|}{20} & \multicolumn{1}{c|}{all} & 1-15 & \multicolumn{1}{c|}{16-20} & \multicolumn{1}{c|}{all} & 1-15 & \multicolumn{1}{c|}{16-20} & \multicolumn{1}{c|}{all} & 1-15 & \multicolumn{1}{c|}{16-20} & \multicolumn{1}{c|}{all} & 1-15 & \multicolumn{1}{c|}{16-20} & all \\
\midrule
\multicolumn{1}{l|}{FT} & 5.8 & \multicolumn{1}{c|}{12.3} & \multicolumn{1}{c|}{6.2} & 6.8 & \multicolumn{1}{c|}{12.9} & \multicolumn{1}{c|}{7.1} & 1.1 & \multicolumn{1}{c|}{33.6} & \multicolumn{1}{c|}{9.2} & 2.1 & \multicolumn{1}{c|}{33.1} & \multicolumn{1}{c|}{9.8} & 0.2 & \multicolumn{1}{c|}{1.8} & \multicolumn{1}{c|}{0.6} & 0.2 & \multicolumn{1}{c|}{1.8} & 0.6 \\
\multicolumn{1}{l|}{PI \cite{zenke2017continual}} & 5.4 & \multicolumn{1}{c|}{14.1} & \multicolumn{1}{c|}{5.9} & 7.5 & \multicolumn{1}{c|}{14.0} & \multicolumn{1}{c|}{7.8} & 1.3 & \multicolumn{1}{c|}{34.1} & \multicolumn{1}{c|}{9.5} & 1.6 & \multicolumn{1}{c|}{33.3} & \multicolumn{1}{c|}{9.5} & 0.0 & \multicolumn{1}{c|}{1.8} & \multicolumn{1}{c|}{0.4} & 0.0 & \multicolumn{1}{c|}{1.8} & 0.4 \\
\multicolumn{1}{l|}{EWC \cite{kirkpatrick2017overcoming}} & 23.2 & \multicolumn{1}{c|}{16.0} & \multicolumn{1}{c|}{22.9} & 26.9 & \multicolumn{1}{c|}{14.0} & \multicolumn{1}{c|}{26.3} & 26.7 & \multicolumn{1}{c|}{37.7} & \multicolumn{1}{c|}{29.4} & 24.3 & \multicolumn{1}{c|}{35.5} & \multicolumn{1}{c|}{27.1} & 0.3 & \multicolumn{1}{c|}{4.3} & \multicolumn{1}{c|}{1.3} & 0.3 & \multicolumn{1}{c|}{4.3} & 1.3 \\
\multicolumn{1}{l|}{RW \cite{chaudhry2018riemannian}} & 19.4 & \multicolumn{1}{c|}{15.7} & \multicolumn{1}{c|}{19.2} & 23.3 & \multicolumn{1}{c|}{14.2} & \multicolumn{1}{c|}{22.9} & 17.9 & \multicolumn{1}{c|}{36.9} & \multicolumn{1}{c|}{22.7} & 16.6 & \multicolumn{1}{c|}{34.9} & \multicolumn{1}{c|}{21.2} & 0.2 & \multicolumn{1}{c|}{5.4} & \multicolumn{1}{c|}{1.5} & 0.0 & \multicolumn{1}{c|}{5.2} & 1.3 \\
\multicolumn{1}{l|}{LwF \cite{li2017learning}} & 53.0 & \multicolumn{1}{c|}{9.1} & \multicolumn{1}{c|}{50.8} & 51.2 & \multicolumn{1}{c|}{8.5} & \multicolumn{1}{c|}{49.1} & 58.4 & \multicolumn{1}{c|}{37.4} & \multicolumn{1}{c|}{53.1} & 58.9 & \multicolumn{1}{c|}{36.6} & \multicolumn{1}{c|}{53.3} & 0.8 & \multicolumn{1}{c|}{3.6} & \multicolumn{1}{c|}{1.5} & 1.0 & \multicolumn{1}{c|}{3.9} & 1.8 \\
\multicolumn{1}{l|}{LwF-MC \cite{rebuffi2017icarl}} & 63.0 & \multicolumn{1}{c|}{13.2} & \multicolumn{1}{c|}{60.5} & 64.4 & \multicolumn{1}{c|}{13.3} & \multicolumn{1}{c|}{61.9} & 67.2 & \multicolumn{1}{c|}{41.2} & \multicolumn{1}{c|}{60.7} & 58.1 & \multicolumn{1}{c|}{35.0} & \multicolumn{1}{c|}{52.3} & 4.5 & \multicolumn{1}{c|}{7.0} & \multicolumn{1}{c|}{5.2} & 6.4 & \multicolumn{1}{c|}{8.4} & 6.9 \\
\multicolumn{1}{l|}{ILT \cite{michieli2019incremental}} & 69.1 & \multicolumn{1}{c|}{16.4} & \multicolumn{1}{c|}{66.4} & 67.1 & \multicolumn{1}{c|}{12.3} & \multicolumn{1}{c|}{64.4} & 63.2 & \multicolumn{1}{c|}{39.5} & \multicolumn{1}{c|}{57.3} & 66.3 & \multicolumn{1}{c|}{40.6} & \multicolumn{1}{c|}{59.9} & 3.7 & \multicolumn{1}{c|}{5.7} & \multicolumn{1}{c|}{4.2} & 4.9 & \multicolumn{1}{c|}{7.8} & 5.7 \\
\multicolumn{1}{l|}{MiB \cite{cermelli2020modeling}} & 69.6 & \multicolumn{1}{c|}{25.6} & \multicolumn{1}{c|}{67.4} & 70.2 & \multicolumn{1}{c|}{22.1} & \multicolumn{1}{c|}{67.8} & 71.8 & \multicolumn{1}{c|}{{43.3}} & \multicolumn{1}{c|}{64.7} & 75.5 & \multicolumn{1}{c|}{49.4} & \multicolumn{1}{c|}{69.0} & 46.2 & \multicolumn{1}{c|}{12.9} & \multicolumn{1}{c|}{37.9} & 35.1 & \multicolumn{1}{c|}{13.5} & 29.7 \\
\multicolumn{1}{l|}{SDR~\cite{michieli2021continual}} & 70.8 & \multicolumn{1}{c|}{{31.4} } & \multicolumn{1}{c|}{68.9} & 71.3 & \multicolumn{1}{c|}{23.4} & \multicolumn{1}{c|}{69.0} & 74.6 & \multicolumn{1}{c|}{\textbf{44.1}} & \multicolumn{1}{c|}{\textbf{67.3}} & 76.3 & \multicolumn{1}{c|}{\textbf{50.2}} & \multicolumn{1}{c|}{70.1} & \textbf{59.4} & \multicolumn{1}{c|}{14.3} & \multicolumn{1}{c|}{\textbf{48.7}} & 47.3 & \multicolumn{1}{c|}{{14.7}} & 39.5 \\
\multicolumn{1}{l|}{PLOP$^*$~\cite{douillard2020plop}} & 75.1 & \multicolumn{1}{c|}{\textbf{38.2}} & \multicolumn{1}{c|}{73.2} & {75.0} & \multicolumn{1}{c|}{{39.1}} & \multicolumn{1}{c|}{{73.2}} & 66.5 & \multicolumn{1}{c|}{{39.6}} & \multicolumn{1}{c|}{59.8} & {74.7} & \multicolumn{1}{c|}{{49.8}} & \multicolumn{1}{c|}{{68.5}} & 49.0 & \multicolumn{1}{c|}{{13.8}} & \multicolumn{1}{c|}{40.2} & \textbf{65.2} & \multicolumn{1}{c|}{{\textbf{22.4}}} & \textbf{54.5} \\
\multicolumn{1}{l|}{Ours} & \textbf{75.5} & \multicolumn{1}{c|}{{30.8}} & \multicolumn{1}{c|}{\textbf{73.3}} & \textbf{75.5} & \multicolumn{1}{c|}{\textbf{34.8}} & \multicolumn{1}{c|}{\textbf{73.4}} & \textbf{72.9} & \multicolumn{1}{c|}{42.1} & \multicolumn{1}{c|}{{65.2}} & \textbf{77.2} & \multicolumn{1}{c|}{{49.9}} & \multicolumn{1}{c|}{\textbf{70.4}} & {57.2} & \multicolumn{1}{c|}{\textbf{15.5}} & \multicolumn{1}{c|}{{46.7}} & {55.7} & \multicolumn{1}{c|}{{14.1}} & {45.3} \\
\midrule
\multicolumn{1}{l|}{Joint} & 77.4 & \multicolumn{1}{c|}{78.0} & \multicolumn{1}{c|}{77.4} & 77.4 & \multicolumn{1}{c|}{78.0} & \multicolumn{1}{c|}{77.4} & 79.1 & \multicolumn{1}{c|}{72.6} & \multicolumn{1}{c|}{77.4} & 79.1 & \multicolumn{1}{c|}{72.6} & \multicolumn{1}{c|}{77.4} & 79.1 & \multicolumn{1}{c|}{72.6} & \multicolumn{1}{c|}{77.4} & 79.1 & \multicolumn{1}{c|}{72.6} & 77.4\\
\bottomrule
\end{tabular}%
}
\end{table*}

\begin{table*}[]
\caption{Mean IoU on the ADE20K dataset for different incremental class learning scenarios.$*$ means results come from re-implementation.}

\label{tab:overall ade}
\resizebox{\textwidth}{!}{%
\begin{tabular}{ccccccccccccccc}
\toprule
\multirow{2.5}{*}{Method} & \multicolumn{3}{c|}{100-50} & \multicolumn{7}{c|}{100-10} & \multicolumn{4}{c}{50-50} \\\cmidrule{2-15}
& 1-100 & \multicolumn{1}{c|}{101-150} & \multicolumn{1}{c|}{all} & 1-100 & 100-110 & 110-120 & 120-130 & 130-140 & \multicolumn{1}{c|}{140-150} & \multicolumn{1}{c|}{all} & 1-50 & 51-100 & \multicolumn{1}{c|}{101-150} & all \\
\midrule
\multicolumn{1}{l|}{FT} & 0.0 & \multicolumn{1}{c|}{24.9} & \multicolumn{1}{c|}{8.3} & 0.0 & 0.0 & 0.0 & 0.0 & 0.0 & \multicolumn{1}{c|}{16.6} & \multicolumn{1}{c|}{1.1} & 0.0 & 0.0 & \multicolumn{1}{c|}{22.0} & 7.3 \\
\multicolumn{1}{l|}{LwF\cite{li2017learning}} & 21.1 & \multicolumn{1}{c|}{25.6} & \multicolumn{1}{c|}{22.6} & 0.1 & 0.0 & 0.4 & 2.6 & 4.6 & \multicolumn{1}{c|}{16.9} & \multicolumn{1}{c|}{1.7} & 5.7 & 12.9 & \multicolumn{1}{c|}{22.8} & 13.9 \\
\multicolumn{1}{l|}{LwF-MC\cite{rebuffi2017icarl}} & 34.2 & \multicolumn{1}{c|}{10.5} & \multicolumn{1}{c|}{26.3} & 18.7 & 2.5 & 8.7 & 4.1 & 6.5 & \multicolumn{1}{c|}{5.1} & \multicolumn{1}{c|}{14.3} & 27.8 & 7.0 & \multicolumn{1}{c|}{10.4} & 15.1 \\
\multicolumn{1}{l|}{ILT\cite{michieli2019incremental}} & 22.9 & \multicolumn{1}{c|}{18.9} & \multicolumn{1}{c|}{21.6} & 0.3 & 0.0 & 1.0 & 2.1 & 4.6 & \multicolumn{1}{c|}{10.7} & \multicolumn{1}{c|}{1.4} & 8.4 & 9.7 & \multicolumn{1}{c|}{14.3} & 10.8 \\
\multicolumn{1}{l|}{Inc. Seg \cite{yan2021framework}} & 36.6 & \multicolumn{1}{c|}{0.4} & \multicolumn{1}{c|}{24.6} & {32.4} & 0.0 & 0.2 & 0.0 & 0.0 & \multicolumn{1}{c|}{0.0} & \multicolumn{1}{c|}{21.7} & 40.2 & 1.3 & \multicolumn{1}{c|}{0.3} & 14.1 \\
\multicolumn{1}{l|}{MiB\cite{cermelli2020modeling}} & \textbf{37.9} & \multicolumn{1}{c|}{27.9} & \multicolumn{1}{c|}{34.6} & 31.8 & 10.4 & 14.8 & 12.8 & \textbf{13.6} & \multicolumn{1}{c|}{\textbf{18.7}} & \multicolumn{1}{c|}{25.9} & 35.5 & 22.2 & \multicolumn{1}{c|}{\textbf{23.6}} & 27.0 \\
\multicolumn{1}{l|}{SDR \cite{michieli2021continual}} & 37.5 & \multicolumn{1}{c|}{25.5} & \multicolumn{1}{c|}{33.5} & 28.9 & - & - & - & - & \multicolumn{1}{c|}{-} & \multicolumn{1}{c|}{23.2} & {{42.9}} & - & \multicolumn{1}{c|}{\textbf{-}} & 
{{31.3}} \\
\multicolumn{1}{l|}{{PLOP$^*$}~\cite{douillard2020plop}} & 29.8 & \multicolumn{1}{c|}{4.2} & \multicolumn{1}{c|}{22.2 } & {32.1} &  1.9 & 10.0 & 0.8 & 1.2 & \multicolumn{1}{c|}{0.1} & \multicolumn{1}{c|}{22.3} & 19.2 & 0.4 & \multicolumn{1}{c|}{0.4} & 6.6 \\
\multicolumn{1}{l|}{Ours} & 37.3 & \multicolumn{1}{c|}{\textbf{31.9}} & \multicolumn{1}{c|}{\textbf{35.5} } & \textbf{39.0} & \textbf{14.6} & \textbf{22.0} & \textbf{25.4} & 12.1 & \multicolumn{1}{c|}{13.1} & \multicolumn{1}{c|}{\textbf{31.8}} & \textbf{47.5} & \textbf{30.6} & \multicolumn{1}{c|}{23.0} & \textbf{33.7} \\
\midrule
\multicolumn{1}{l|}{Joint} & 44.3 & \multicolumn{1}{c|}{28.2} & \multicolumn{1}{c|}{38.9} & 44.3 & 26.1 & 42.8 & 26.7 & 28.1 & \multicolumn{1}{c|}{17.3} & \multicolumn{1}{c|}{38.9} & 51.1 & 38.3 & \multicolumn{1}{c|}{28.2} & 38.9\\
\bottomrule
\end{tabular}%
}
\end{table*}

\section{Experiments}

In this section, we demonstrate the effectiveness of our approach through extensive experiments on the two publicly available benchmarks for ICL in semantic segmentation. 

\subsection{Experimental Setups}

\noindent \textbf{Datasets.} We consider two datasets in our experiments.
The PASCAL-VOC 2012 dataset~\cite{everingham2007pascal} contains 10,582 and 1,449 in the training and validation set, respectively. Pixels can be associated to 21 different classes (20 plus the background). 
Following~\cite{michieli2019incremental,shmelkov2017incremental,cermelli2020modeling,tasar2019incremental}, we define two experimental settings: disjoint and overlapped. 
Following~\cite{michieli2019incremental}, the disjoint setup assumes that the new set is disjoint from previously used samples, \ie\ $(\cup_{j=0,...,k-1} \mathcal{D}_j\cap \mathcal{D}_k=\emptyset)$. Following~\cite{shmelkov2017incremental}, in the
overlapped setting each training step contains all the images that have at least one pixel of a novel class, no matter what other classes are also included. It is important to know that in this case, training images may contain pixels of unseen classes (thus labeled as background). This is a more realistic setup since it does not make any restriction on the objects present in the images.
Following previous work~\cite{shmelkov2017incremental,michieli2019incremental,cermelli2020modeling}, we perform three
different experiments concerning the addition of one class (19-1), five classes all at once (15-5), and five classes added one-by-one in alphabetical order (15-1), and report mean IoU.

The ADE20K~\cite{zhou2017scene} is a large-scale dataset with 150 classes. Differently from Pascal-VOC 2012, this dataset contains non-object classes (\eg\ sky, building, wall). We create the incremental datasets $\mathcal{D}_\ell$ by splitting the whole dataset into disjoint image sets, without any constraint except ensuring a minimum number of images (\ie\ 50) containing new classes. Obviously, each $\mathcal{D}_\ell$ provides annotations only for current classes, while old and future classes are annotated as background. We report the mean IoU obtained averaging the results as in~\cite{zhou2017scene} and we perform three
different experiments: single-step addition of 50 classes (100-50), multi-step addition of 50 classes (100-10) and three steps of 50 classes (50-50).

\begin{table*}[]
\caption{Ablation study of networks' components on the Pascal-VOC 2012 disjoint 15-5 setup. Per-class IoU of the evaluated methods when the last five classes are added are reported. CAF denotes our continual attentive fusion module, AD attentive feature distillation, BKD  the balanced knowledge distillation loss and KD knowledge distillation as in~(\ref{eq:kd old}) and~(\ref{eq:kd mib}).}
\label{tab:ablation}
\resizebox{\textwidth}{!}{%
\begin{tabular}{lccccccccccccccc|c|ccccc|c}
\toprule
Method & \rotatebox{90}{aero} & \rotatebox{90}{bike} & \rotatebox{90}{bird} & \rotatebox{90}{boat} & \rotatebox{90}{bottle} & \rotatebox{90}{bus} & \rotatebox{90}{car} & \rotatebox{90}{cat} & \rotatebox{90}{chair} & \rotatebox{90}{cow} & \rotatebox{90}{din. table} & \rotatebox{90}{dog} & \rotatebox{90}{horse} & \rotatebox{90}{mbike} & \rotatebox{90}{person} & \rotatebox{90}{mIoU old} & \rotatebox{90}{plant} & \rotatebox{90}{sheep} & \rotatebox{90}{sofa} & \rotatebox{90}{train}  & \rotatebox{90}{tv} & \rotatebox{90}{mIoU} \\
\midrule
Baseline & 70.6 & 33.5 & 73.5 & 59.2 & 66.6 & 49.1 & 72.6 & 74.7 & 28.6 & 34.8 & 43.3 & 72.6 & 70.7 & 69.9 & 70.2 & 59.3 & 27.4 & 31.0 & 25.9 & 41.5 & 44.5 & 53.0 \\
+ KD &  77.9 & 36.9 & 81.1 & 65.3 & 73.4 & 54.2 & 80.1 & 82.4 & 31.6 & 38.4 & 47.7 & 80.1 & 78.0 & 77.1 & 77.4 & 65.4 & 30.3 & 34.2 & 28.6 & 45.8 & 49.1 & 58.5 \\
+ BKD & 79.1 & 37.5 & 82.3 & 66.3 & 74.5 & 55.0 & 81.4 & 83.7 & 32.1 & 39.0 & 48.5 & 81.4 & 79.2 & 78.3 & 78.6 & 66.5 & 30.7 & 34.7 & 29.1 & 46.5 & 49.8 & 59.4\\
+ AD & 79.9 & 37.9 & 83.2 & 67.0 & 75.3 & 55.6 & 82.2 & 84.5 & 32.4 & 39.4 & 48.9 & 82.2 & 80.0 & 79.1 & 79.4 & 67.1 & 31.0 & 35.0 & 29.3 & 46.9 & 50.3 & 60.0 \\
+ AD, BKD & 83.3 & 39.5 & 86.7 & 69.8 & 78.5 & 58.0 & 85.7 & 88.2 & 33.8 & 41.1 & 51.1 & 85.7 & 83.4 & 82.5 & 82.8 & 70.0 & 32.4 & 36.5 & 30.6 & 49.0 & 52.5 & 62.6 \\
\midrule
+ CAF & 85.4 & 40.5 & 88.9 & 71.6 & 80.5 & 59.4 & 87.8 & 90.3 & 34.6 & 42.1 & 52.3 & 87.8 & 85.5 & 84.5 & 84.9 & 71.7 & 33.2 & 37.4 & 31.4 & 50.2 & 53.8 & 64.1 \\
+ CAF, BKD & 86.1 & 40.8 & 89.6 & 72.1 & 81.1 & 59.9 & 88.6 & 91.1 & 34.9 & 42.4 & 52.7 & 88.5 & 86.2 & 85.2 & 85.6 & 72.3 & 33.5 & 37.8 & 31.6 & 50.6 & 54.2 & 64.6 \\
+ CAF, AD & 86.3 & 40.9 & 89.8 & 72.3 & 81.3 & 60.0 & 88.8 & 91.3 & 35.0 & 42.5 & 52.9 & 88.7 & 86.4 & 85.4 & 85.8 & 72.5 & 33.5 & 37.8 & 31.7 & 50.7 & 54.4 & 64.8 \\
+ CAF, AD, BKD & \textbf{86.8} & \textbf{41.1} & \textbf{90.4} & \textbf{72.8} & \textbf{81.8} & \textbf{60.4} & \textbf{89.3} & \textbf{91.9} & \textbf{35.2} & \textbf{42.8} & \textbf{53.2} & \textbf{89.3} & \textbf{86.9} & \textbf{85.9} & \textbf{86.3} & \textbf{72.9} & \textbf{33.7} & \textbf{38.1} & \textbf{31.9} & \textbf{51.0} & \textbf{54.7} & \textbf{65.2} \\
\bottomrule
\end{tabular}%
}
\end{table*}

\noindent \textbf{Baselines.} We compare ours method against previous ICL methods originally designed for image classification, following~\cite{cermelli2020modeling}, namely: Path Integral (PI)~\cite{zenke2017continual}, Elastic Weight Consolidation (EWC)~\cite{kirkpatrick2017overcoming}, and Riemannian Walks (RW)~\cite{chaudhry2018riemannian}. We also compare our method with Learning without Forgetting (LwF)~\cite{li2017learning} and its multi-class version (LwF-MC)~\cite{rebuffi2017icarl}. Finally, we consider previous ICL approaches for segmentation, \textit{i.e.} MiB~\cite{cermelli2020modeling}, ILT~\cite{michieli2019incremental}, SDR~\cite{michieli2021continual}, and PLOP~\cite{douillard2020plop}. In addition to the state-of-the-art methods, we report results for fine tuning (FT) and joint training (Joint). These results serve as a lower and upper bound. In FT, we train on the new task via simple fine tuning, while in Joint a unique training of both tasks is performed. 

\begin{table}[t]
 \centering
 \caption{mIoU for different incremental learning scenarios.}
\label{tab:more}
 \resizebox{0.75\linewidth}{!}{%
\begin{tabular}{cccc|ccc}
\toprule
\multirow{2}{*}{Method} & \multicolumn{3}{c|}{VOC 15-5} & \multicolumn{3}{c}{ADE 100-50} \\ \cmidrule{2-7} & 1-15   & 16-20   & all   & 1-100   & 101-150  & all   \\
\midrule
Only Step 1 & \textbf{79.6}   & -       & -     & \textbf{42.7}    & -        & -     \\
Fine Tuning & 1.1    & 33.6    & 9.2   & 0.0     & 24.9     & 8.3   \\
Ours & 72.9   & \textbf{42.1}    & \textbf{65.2}  & 37.3    & \textbf{31.9}     & \textbf{35.5}  \\

\bottomrule
\end{tabular}}
\end{table}

\noindent \textbf{Implementation Details.} We choose a ResNet-101~\cite{he2016deep} as our backbone, the Deeplab-v3 architecture~\cite{chen2017rethinking} as refinement model and the {non-local block \cite{wang2018non} as feature projection module}. We initialize our backbone with ImageNet pretraining~\cite{rota2018place} and train the full network as in~\cite{chen2017rethinking} for learning rate policy, momentum, and weight decay. We use an initial learning rate of $10^{-2}$ for the first learning step while $10^{-3}$ and $10^{-2}$ for the following ones in Pascal-VOC 2012 and ADE20K dataset respectively. We train the model with a batch-size of 24 for 30 epochs for Pascal-VOC 2012 and 60 epochs for ADE20K in every learning step. For our loss, $\lambda_D$ and $\lambda_{AD}$ are set to 10 and 1000 respectively. We apply the same data augmentation of~\cite{chen2017rethinking} and crop the images to 512 × 512 during both training and test. For setting the hyper-parameters of each method, we use the protocol of IL defined in~\cite{de2019continual,cermelli2020modeling}, using 20\% of the training set as validation. The final results are reported on the standard validation sets. 

\subsection{Experimental Results}
\label{sec:experiments}
We reports some quantitative and qualitative results associated with our method, as well as the results of an ablation study to demonstrate the merit of our technical contributions.

\par\noindent\textbf{Comparison with State-of-the-Art Methods.}
Table~\ref{tab:overall voc} and Table~\ref{tab:overall ade} compare our approach with state of the art ICL methods.
Looking at results in Table \ref{tab:overall voc}, it is clear how in the case of the Pascal-VOC 2012 dataset our method outperforms all the competitors in almost all the overlapped and the disjoint settings, often by a large margin. 
Comparative results on ADE20K are shown in Table~\ref{tab:overall ade}. 
Our model exhibits competitive performance in all tasks and often better performance than the current art by several points. Only the last step of the 100-10 task is associated to lower performance, but it is compensated by far if we consider the overall task score.

\noindent \textbf{Demeonstration of Catastrophic Forgetting.}
The catastrophic forgetting phenomenon is clearly shown in Table~\ref{tab:more}. Fine-tuning suffers from catastrophic forgetting (-78.5\% on VOC and -42.7\% on ADE20K) on the first task after training the second task. On the other hand, the performance of our method only decreases by 6.7\% on VOC and 5.4\% on ADE20K, which demonstrates the effectiveness of our method for addressing the issue of catastrophic forgetting.

\begin{table}[t]
\centering
\caption{Ablation study of our continual attentive fusion on the Pascal-VOC 2012 disjoint 15-5 setup and ADE disjoint 100-50 setup. ``Baseline + KD'' corresponds to the second row in Table~\ref{tab:ablation}, {``Projection'' denotes the feature projection module}, and ``SAB'' indicate the structured attention block. ``{Without continual attentive fusion}'' means that the attention weights only depend on the current module, while in ``{With continual attentive fusion}'' new and old model features are fused with the fusion module to generate the attention weights.}
\label{tab:ab_cma}
\resizebox{0.95\linewidth}{!}{%
\begin{tabular}{lccc|ccc}
\toprule
\multirow{2}{*}{Method} & \multicolumn{3}{c|}{VOC 15-5} & \multicolumn{3}{c}{ADE 100-50} \\
\cmidrule{2-7} & 1-15 & 16-20 & mIoU & 1-100 & 101-150 & mIoU \\
\midrule
\multicolumn{7}{c}{{Without continual attentive fusion}} \\
\midrule
Baseline + KD  & 65.4 & 37.6 & 58.5 & \textbf{37.9} & 27.9 & 34.6 \\
+ Projection & 65.4 & 38.0 & 58.6 & 36.5 & 31.3 & 34.8 \\
+ Projection, SAB & 66.5 & 38.2 & 59.4 & 36.9 & 31.5 & 35.1\\
\midrule
\multicolumn{7}{c}{{With continual attentive fusion}} \\
\midrule
+ Fusion & 70.7 & 38.7 & 62.7 & 36.8 & 31.5 & 35.0\\
+ Fusion, Projection & 71.1 & 40.6 & 63.5 & 37.1 & 31.7 & 35.3 \\
+ Fusion, Projection, SAB (CAF) & \textbf{71.7} & \textbf{41.2} & \textbf{64.1} & 37.3 & \textbf{31.9} & \textbf{35.5}\\
\bottomrule
\end{tabular}%
}
\end{table}

\noindent \textbf{Ablation Study.}
We perform an ablation study on the VOC 2012 dataset to demonstrate the impact of each component of our model. Table~\ref{tab:ablation} shows the variants of our method, obtained by gradually adding one component at a time. We decided to use a very simple baseline, which takes no precautions against catastrophic forgetting, apart from using the revisited cross entropy loss $\mathcal{L}_{\textsc{SEG}}$ as defined in \cite{cermelli2020modeling}. 
In the top part of the table, we test different combinations of loss functions without considering the CAF module. Undoubtedly, according to the results, balancing knowledge distillation (BKD) (see Section~\ref{sec:bkdist} and~(\ref{eq:bkd})) increases the performance over the standard distillation loss formulation (KD). On the other hand, the results show that attentive feature distillation (AD) mitigates catastrophic forgetting significantly better than just distilling the output probability distribution, improving mIoU by more than 4\%. The model achieves the best performance when AD and BKD are combined. 
In the bottom part of Table~\ref{tab:ablation} we report the same experiments but at this time activating our CAF module. 
Interestingly, when no IL technique is used to alleviate catastrophic forgetting, adding the CAF module dramatically increases the accuracy (+11\%). As before, both BKD and AD further improve performance, and the whole model outperforms all the other variants.

\begin{figure}[t]
    \centering
    \includegraphics[width=0.8\linewidth]{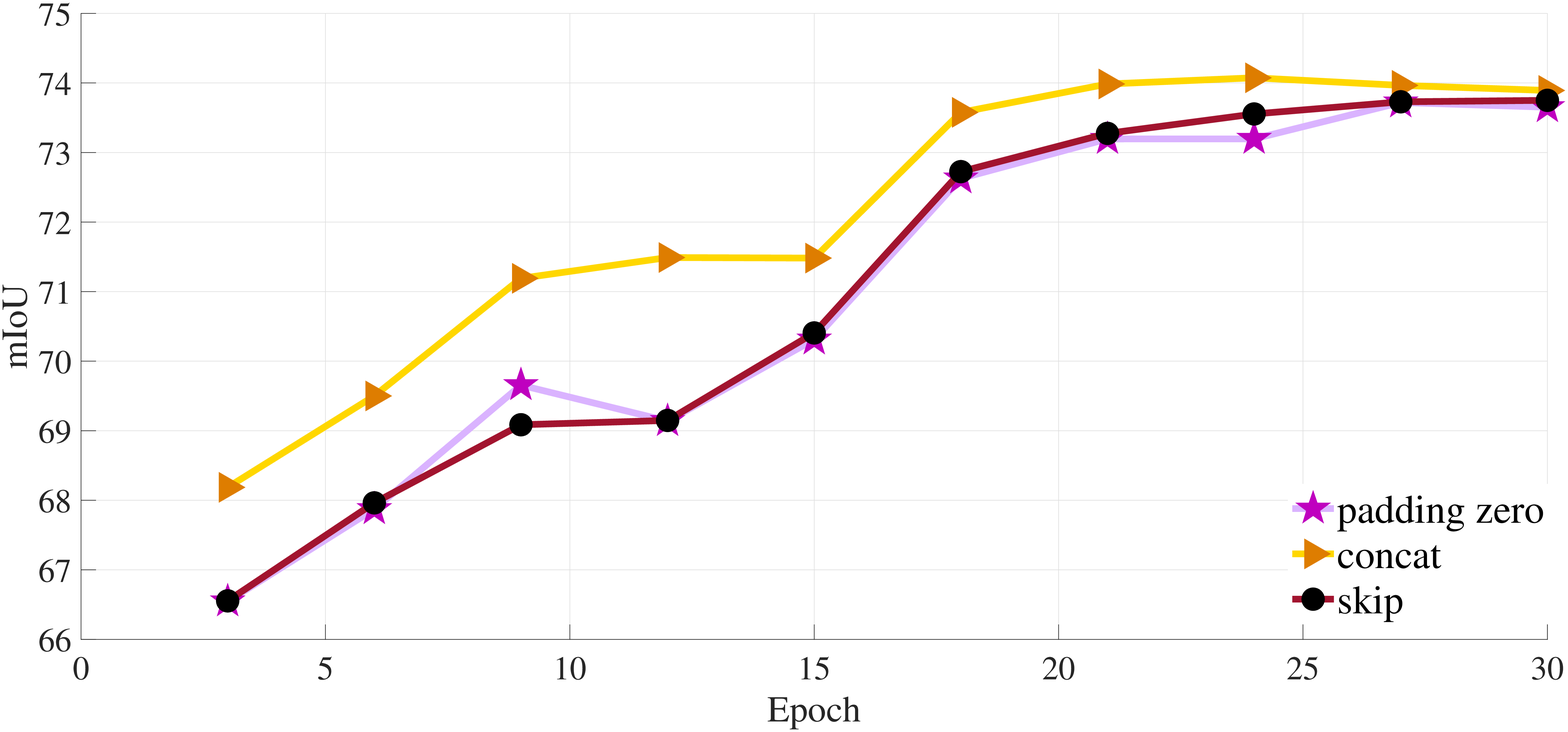} \\   
    \includegraphics[width=0.8\linewidth]{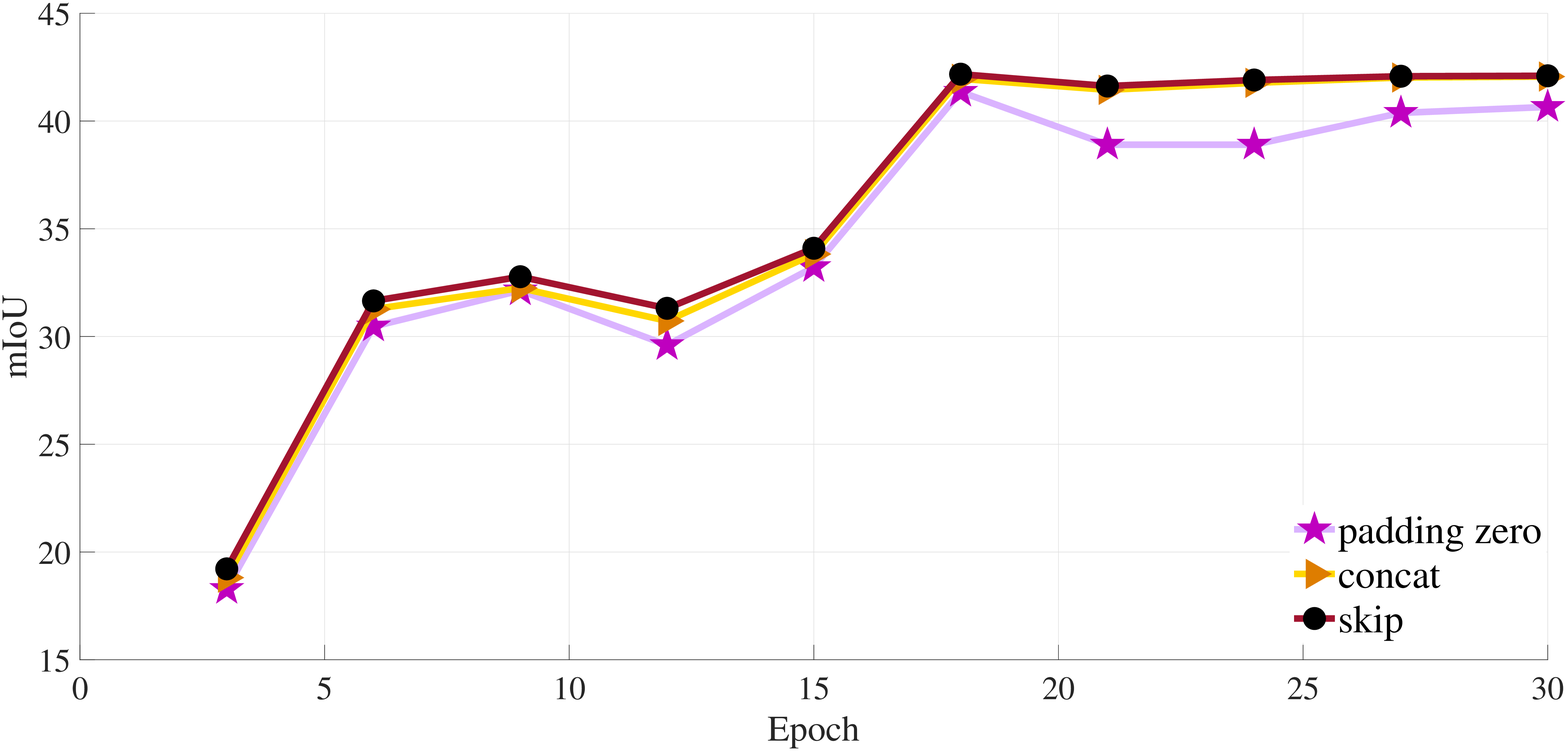} \\   
    \caption{Ablation study: comparison of different strategies for using the fusion module 
    (setting 15-5, top: old categories, bottom: new categories). The curves refer to the concatenation of $\nonlocalfeatures^{\ell}$ and zero padding (purple), to the concatenation of $\nonlocalfeatures^{\ell}$ and $\nonlocalfeatures^{\ell-1}$ (yellow) and to skipping the fusion module and using $\nonlocalfeatures^{\ell}$ (red). }
    \label{fig:ablation cm test}
\end{figure}

\begin{table}[t]
\centering
\caption{Ablation study on attention methods on Pascal-VOC 2012. FD means normal feature distillation (without attention).}
\resizebox{0.85\linewidth}{!}{%
\begin{tabular}{lcccccc}
\toprule
\multicolumn{1}{l}{\multirow{2}{*}{Method}} & \multicolumn{3}{c}{15-5 disjoint} & \multicolumn{3}{c}{19-1 disjoint} \\ \cmidrule(lr){2-4} \cmidrule(lr){5-7} 
\multicolumn{1}{l}{} & 1-15 & 16-20 & all & 1-19 & 20 & all \\
\midrule
Baseline & 59.3 & 34.1 & 53.0 & 69.7 & 24.7 & 67.4 \\
+ FD & 63.2 & 39.5 & 57.3 & 60.3 & 16.3 & 58.1 \\
+ $AD_{SP}$ & 71.5 & 42.7 & 64.3 & 74.2 & 29.2 & 71.9 \\ 
+ $AD_{CH}$ & 10.5 & 11.2 & 10.6 & 1.7 & 16.2 & 2.4 \\
+ $AD_{SP}$ \& $AD_{CH}$ & \textbf{72.5} & \textbf{41.6} & \textbf{64.8} & \textbf{75.1} & \textbf{30.0} & \textbf{72.8}\\
\bottomrule
\end{tabular}%
}
\label{tab:ab_attention}
\end{table}

In addition, we also show the results of an ablation study on the structure of the CAF module in Table~\ref{tab:ab_cma}. First, in the top block, we show that our architectural modifications (feature {projection} module, structured attention block) have minimal impact if not used in composition with continual attentive fusion, showing less than 1\% improvement. Instead, a significant improvement (about 4\%) is achieved using a continual attentive fusion scheme (fusion module is active at training time). Moreover, feature {projection} and SAT seem to be more helpful when used in combination with CAF. This evidence suggests that the improvement does not come from the additional parameters introduced in the feature {projection} and structured attention block. If that were the case, the performance would be boosted even without using the old model. Rather, it is clear that the information transfer between the two models in the CAF module is the real catalyst that enables catastrophic forgetting to be mitigated.

\begin{figure}[!t]
    \centering
    \includegraphics[width=0.8\linewidth]{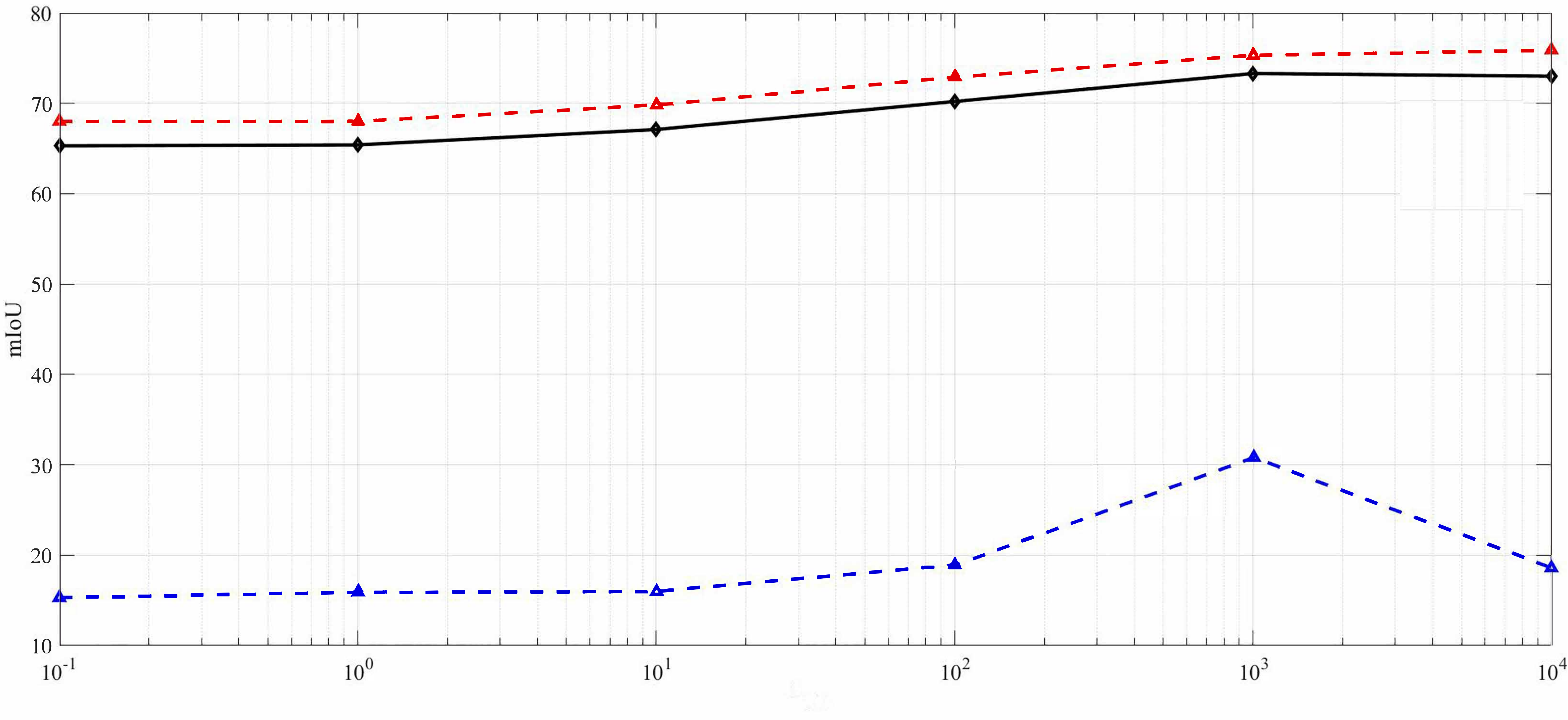} \\   
    \includegraphics[width=0.8\linewidth]{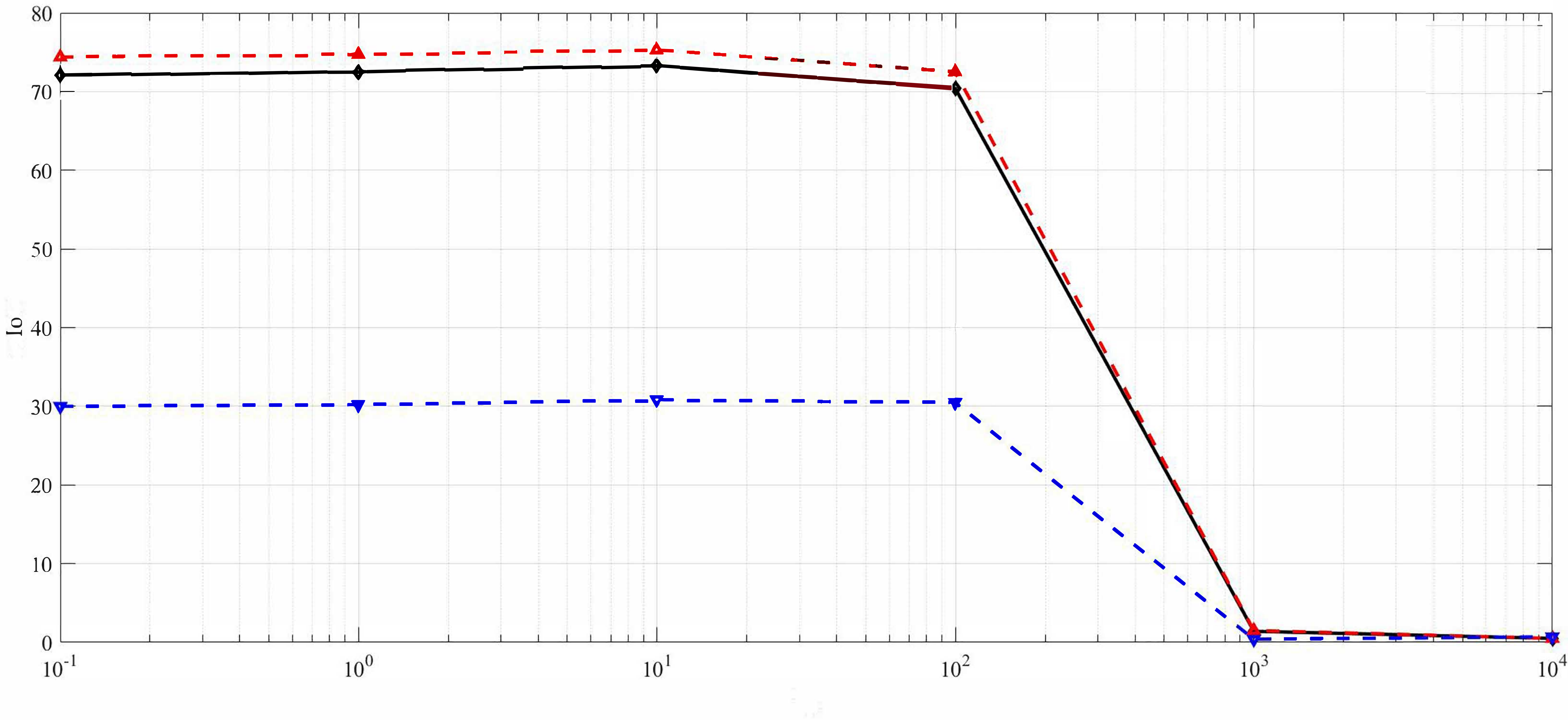} \\   
    \caption{The mIoU of the method for different values of the loss' weights on VOC2012 dataset (19-1 disjoint setup). (a) $\lambda_{\textsc{ad}}$ is fixed to 1000. (b) $\lambda_{\textsc{d}}$ is fixed to 10 . Black line donates overall mIoU; red dash line means old class mIoU; blue dash line means new class mIoU.}
    \label{fig:vis_ab_para}
\end{figure}

\begin{figure*}[!htb]
\centering
\subfigure{
    \begin{minipage}{0.191\linewidth}
        \centering
        \includegraphics[width=0.993\textwidth,height=0.7in]{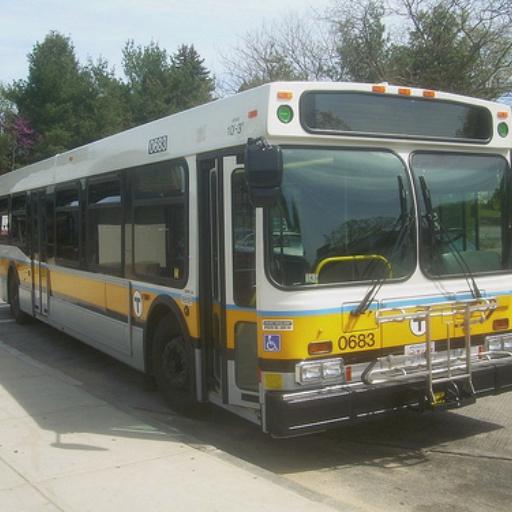}\\ \vspace{0.05cm}
        \includegraphics[width=0.993\textwidth,height=0.7in]{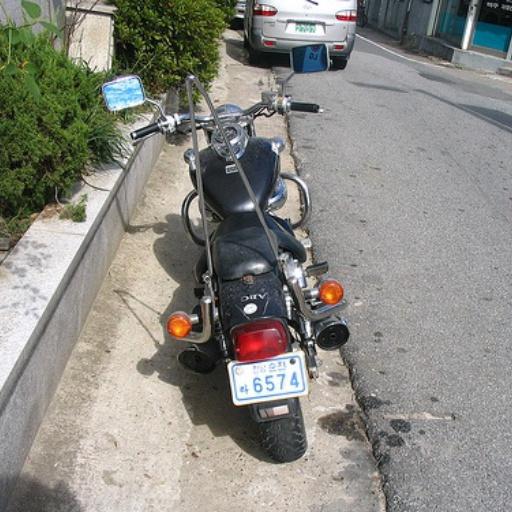}\\    \vspace{0.05cm}     
        \includegraphics[width=0.993\textwidth,height=0.7in]{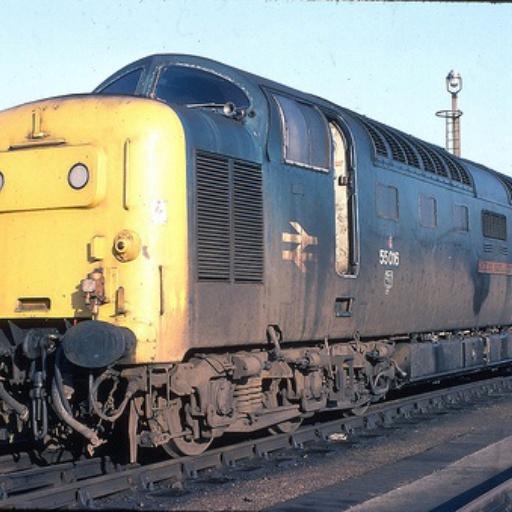}\\ \vspace{0.05cm}
        \includegraphics[width=0.993\textwidth,height=0.7in]{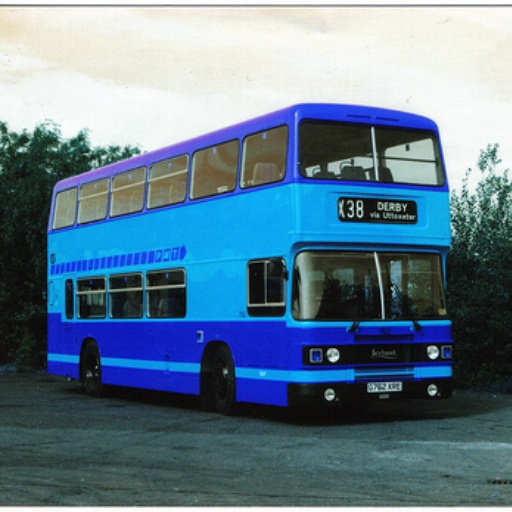}\\   \vspace{0.05cm}
        \includegraphics[width=0.993\textwidth,height=0.7in]{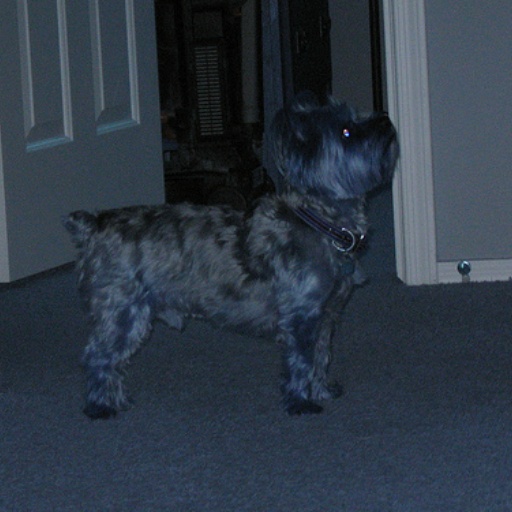}\\ \vspace{0.05cm}
        \includegraphics[width=0.993\textwidth,height=0.7in]{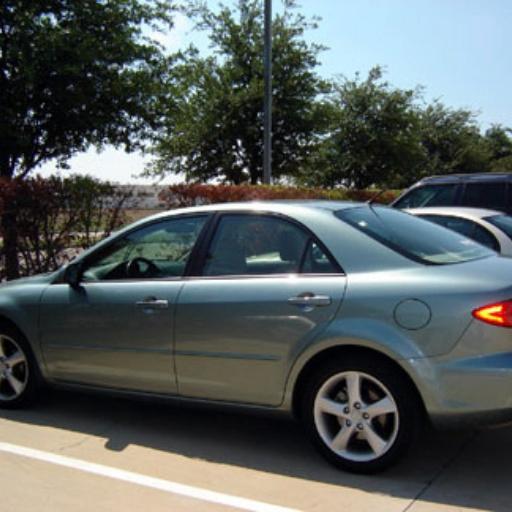}\\ \vspace{0.05cm}
        \includegraphics[width=0.993\textwidth,height=0.7in]{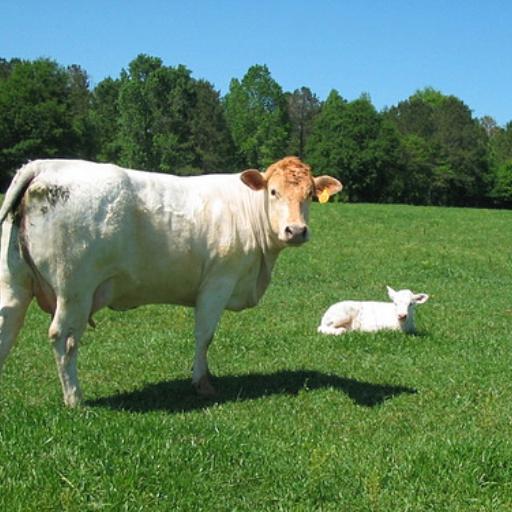}\\ \vspace{0.05cm}
        \includegraphics[width=0.993\textwidth,height=0.7in]{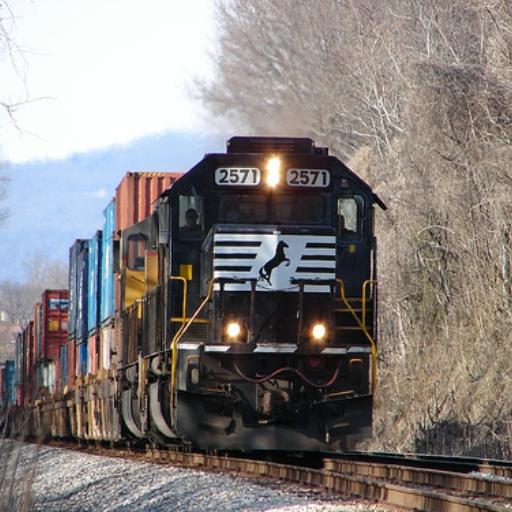}\\ \vspace{0.05cm}
        \includegraphics[width=0.993\textwidth,height=0.7in]{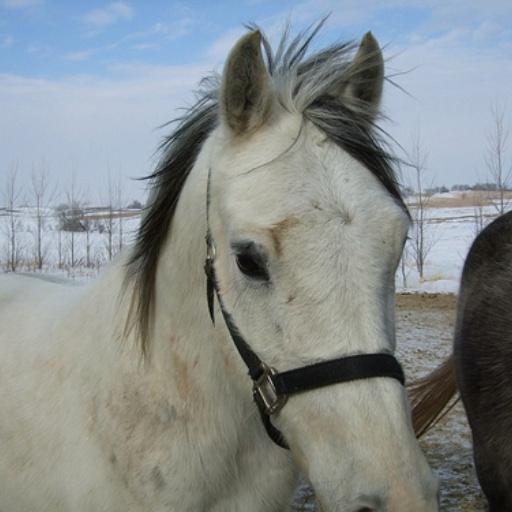}\\  \vspace{0.05cm}
        \includegraphics[width=0.993\textwidth,height=0.7in]{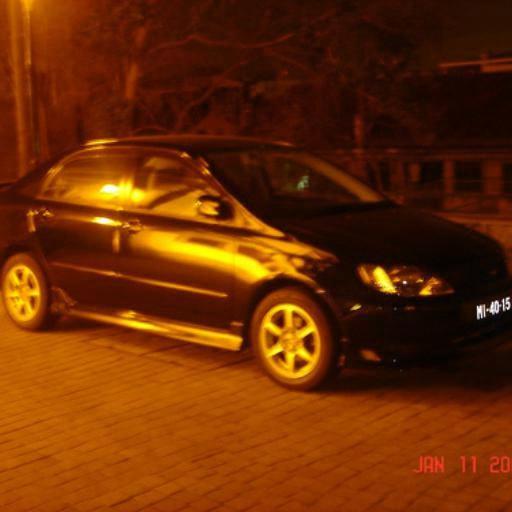}\\ \vspace{0.05cm}
        \includegraphics[width=0.993\textwidth,height=0.7in]{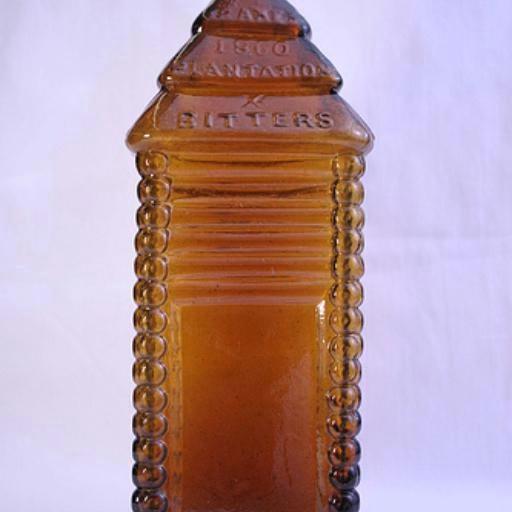}\\  
        (a) Input Image
    \end{minipage}%
}%
\subfigure{
    \begin{minipage}{0.191\linewidth}
        \centering
        \includegraphics[width=0.993\textwidth,height=0.7in]{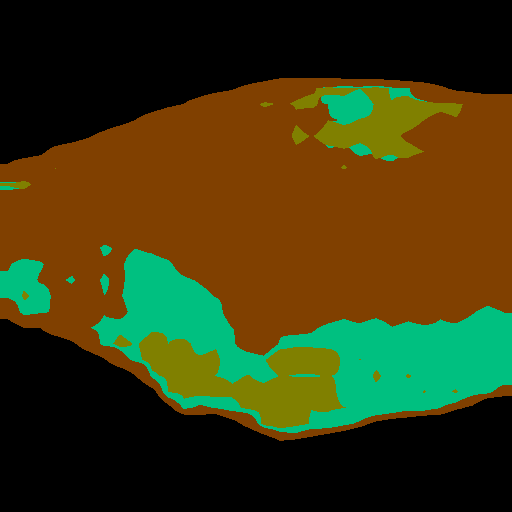}\\ \vspace{0.05cm}
        \includegraphics[width=0.993\textwidth,height=0.7in]{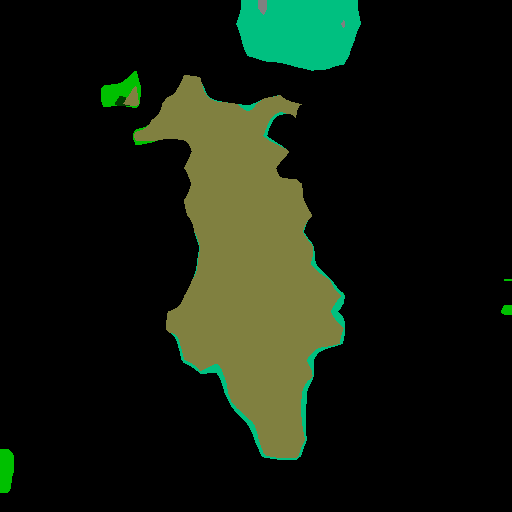}\\      \vspace{0.05cm}   
        \includegraphics[width=0.993\textwidth,height=0.7in]{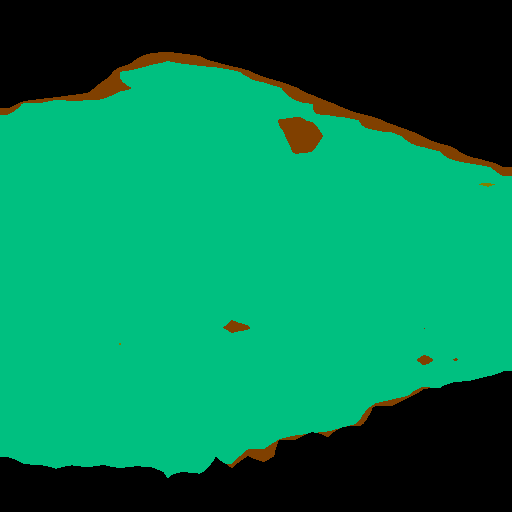}\\ \vspace{0.05cm}
        \includegraphics[width=0.993\textwidth,height=0.7in]{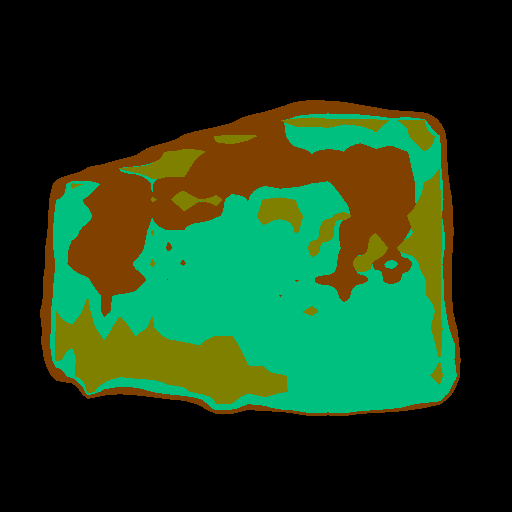}\\   \vspace{0.05cm}
        \includegraphics[width=0.993\textwidth,height=0.7in]{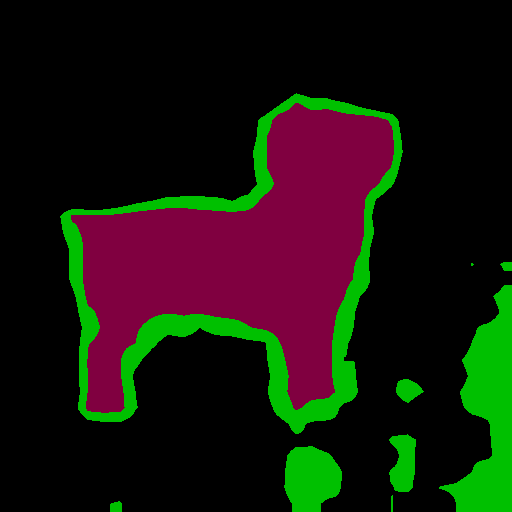}\\ \vspace{0.05cm}
        \includegraphics[width=0.993\textwidth,height=0.7in]{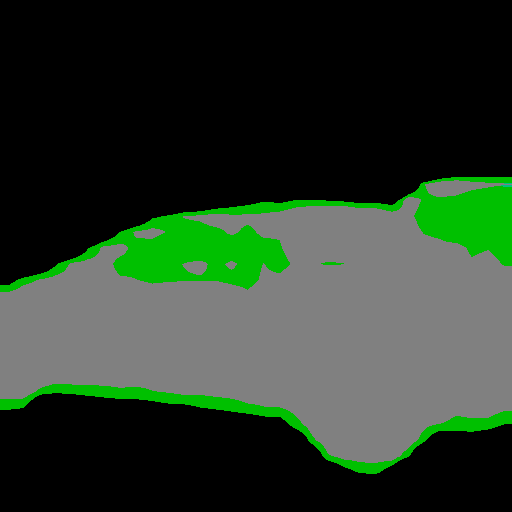}\\ \vspace{0.05cm}
        \includegraphics[width=0.993\textwidth,height=0.7in]{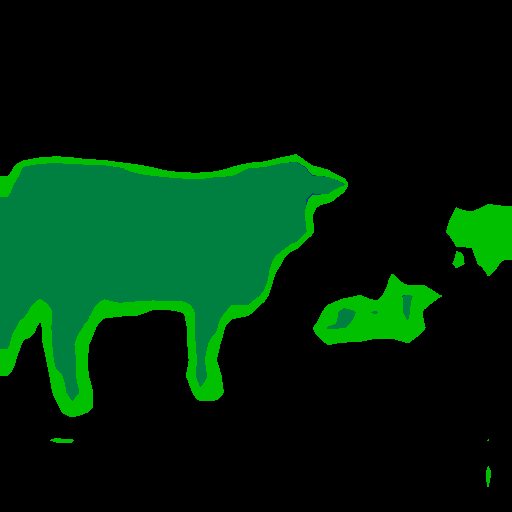}\\ \vspace{0.05cm}
        \includegraphics[width=0.993\textwidth,height=0.7in]{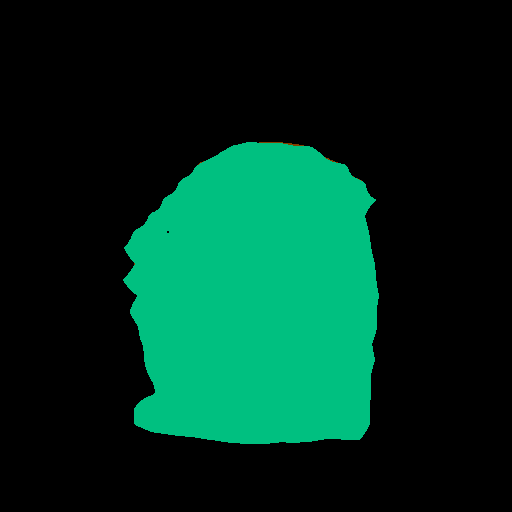}\\ \vspace{0.05cm}
        \includegraphics[width=0.993\textwidth,height=0.7in]{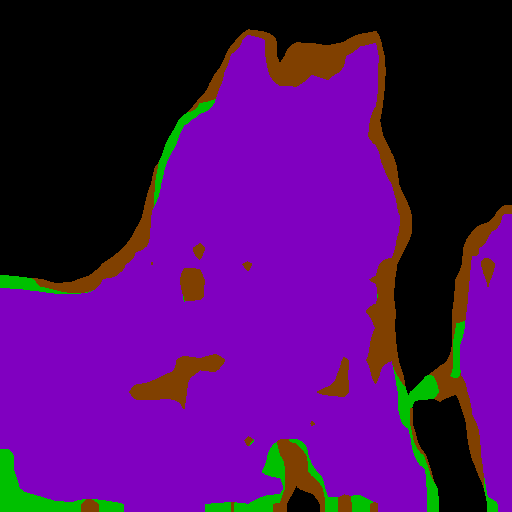}\\  \vspace{0.05cm}
        \includegraphics[width=0.993\textwidth,height=0.7in]{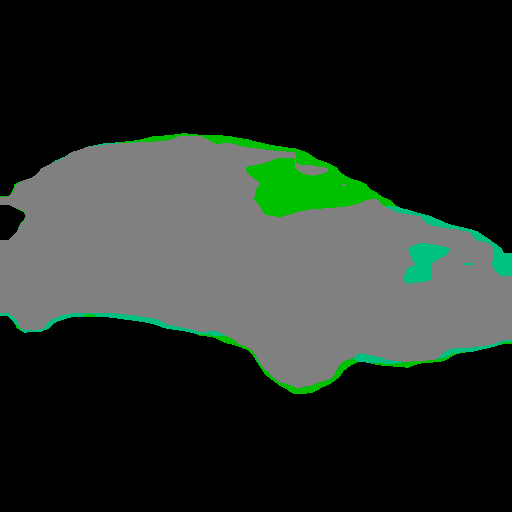}\\ \vspace{0.05cm}
        \includegraphics[width=0.993\textwidth,height=0.7in]{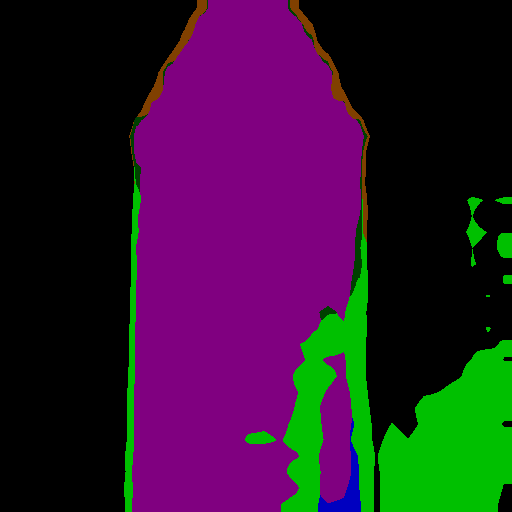}\\   
        (b) ILT~\cite{michieli2019incremental} \\
    \end{minipage}%
}%
\subfigure{
    \begin{minipage}{0.191\linewidth}
        \centering
        \includegraphics[width=0.993\textwidth,height=0.7in]{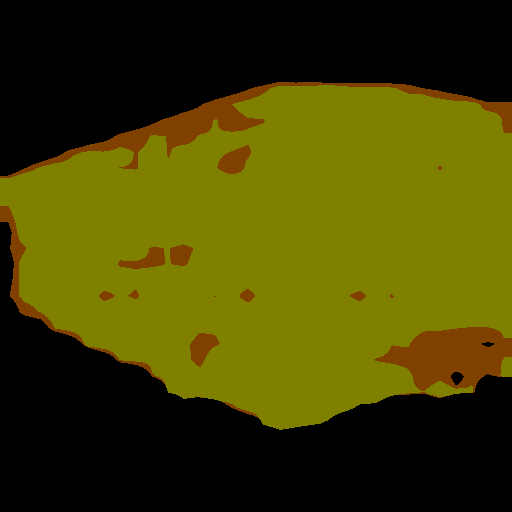}\\ \vspace{0.05cm}
        \includegraphics[width=0.993\textwidth,height=0.7in]{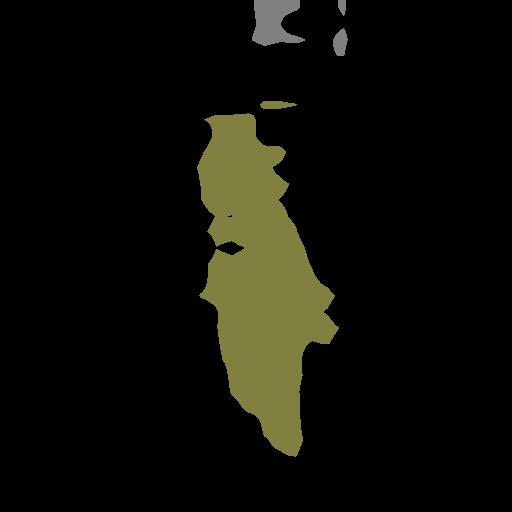}\\        \vspace{0.05cm} 
        \includegraphics[width=0.993\textwidth,height=0.7in]{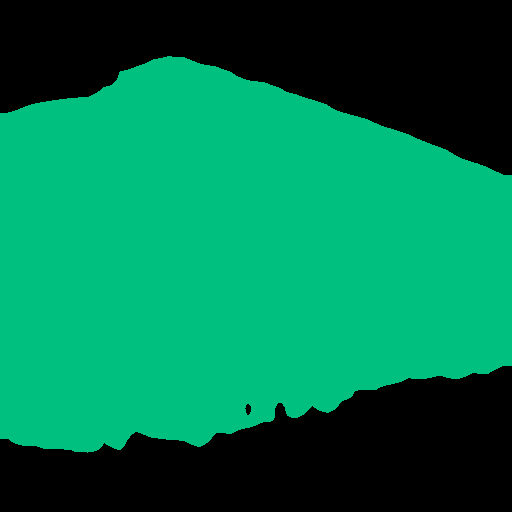}\\ \vspace{0.05cm}
        \includegraphics[width=0.993\textwidth,height=0.7in]{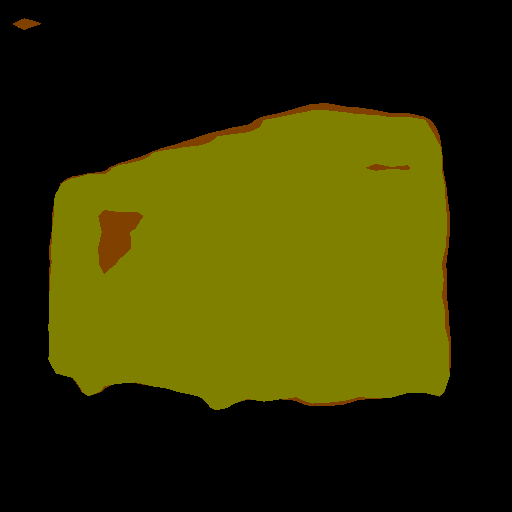}\\   \vspace{0.05cm}
        \includegraphics[width=0.993\textwidth,height=0.7in]{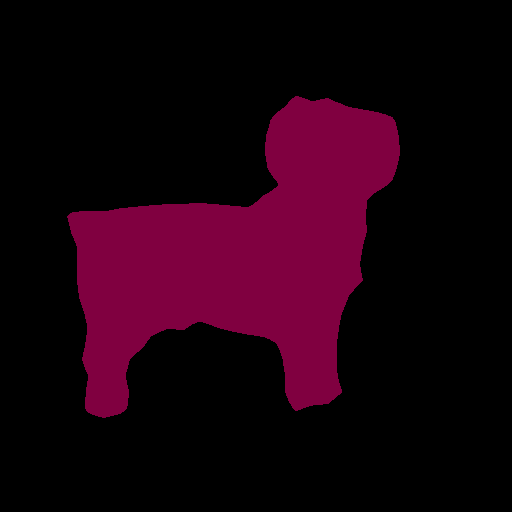}\\ \vspace{0.05cm}
        \includegraphics[width=0.993\textwidth,height=0.7in]{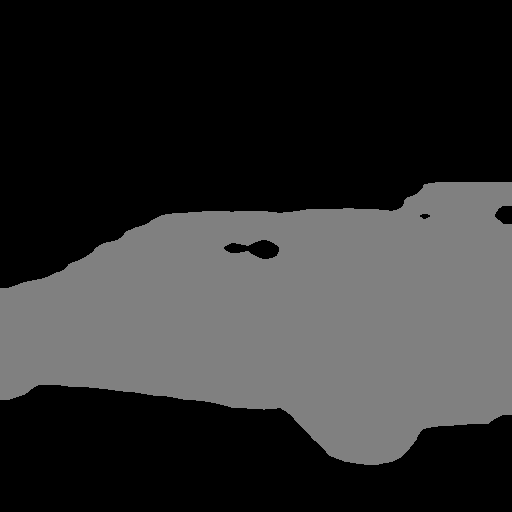}\\ \vspace{0.05cm}
        \includegraphics[width=0.993\textwidth,height=0.7in]{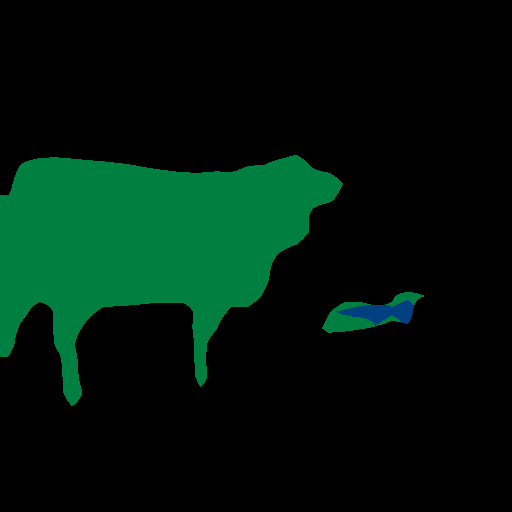}\\ \vspace{0.05cm}
        \includegraphics[width=0.993\textwidth,height=0.7in]{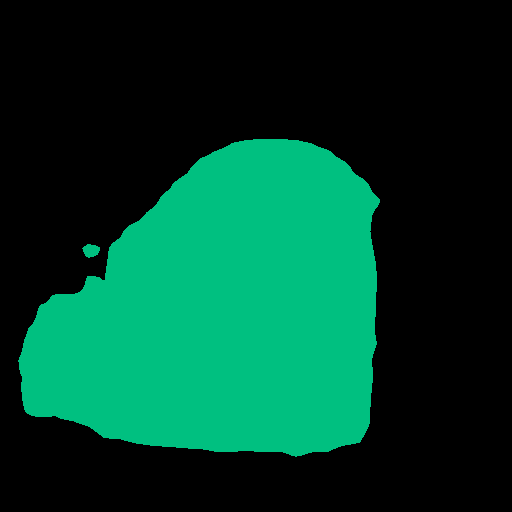}\\ \vspace{0.05cm}
        \includegraphics[width=0.993\textwidth,height=0.7in]{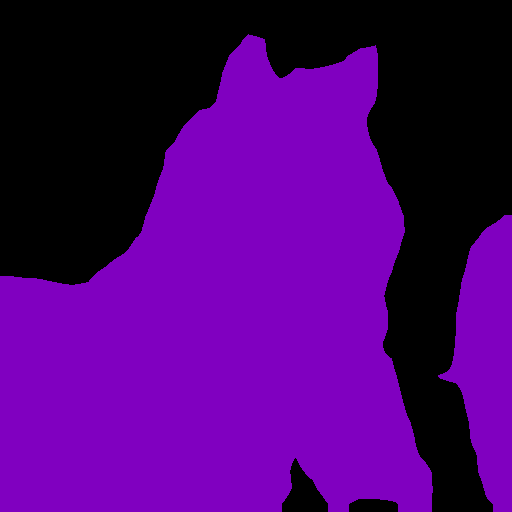}\\  \vspace{0.05cm}
        \includegraphics[width=0.993\textwidth,height=0.7in]{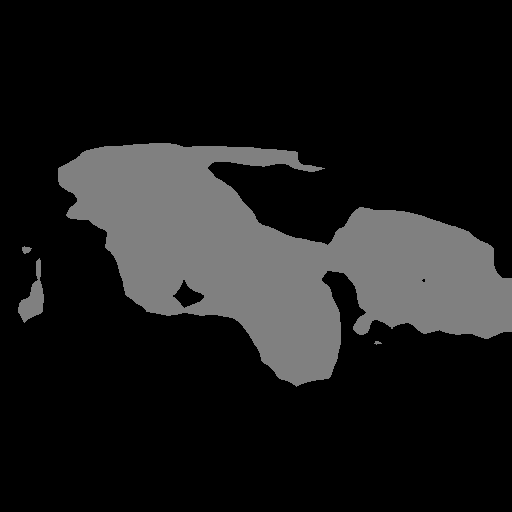}\\ \vspace{0.05cm}
        \includegraphics[width=0.993\textwidth,height=0.7in]{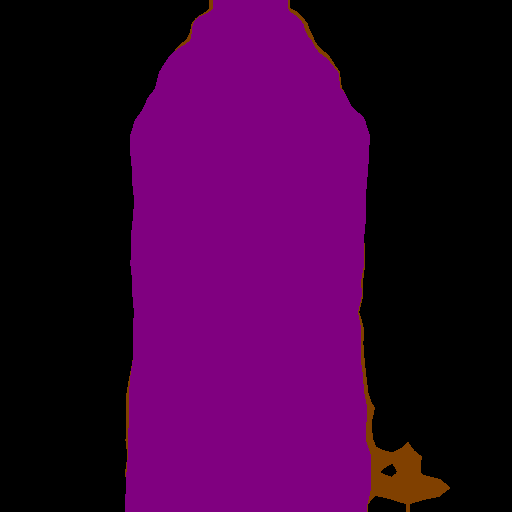}\\   
        (c) MiB~\cite{cermelli2020modeling} \\
    \end{minipage}%
}%
\subfigure{
    \begin{minipage}{0.191\linewidth}
        \centering
        \includegraphics[width=0.993\textwidth,height=0.7in]{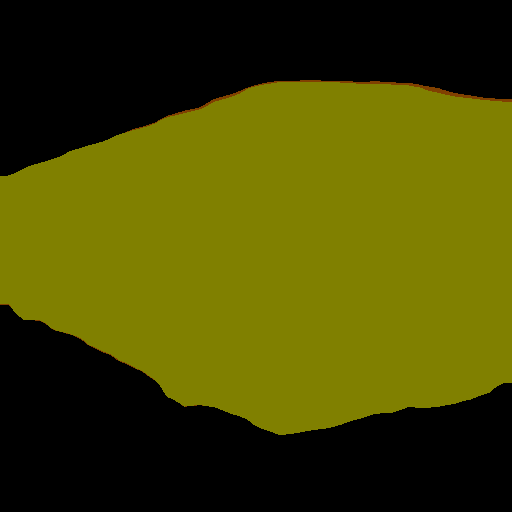}\\ \vspace{0.05cm}
        \includegraphics[width=0.993\textwidth,height=0.7in]{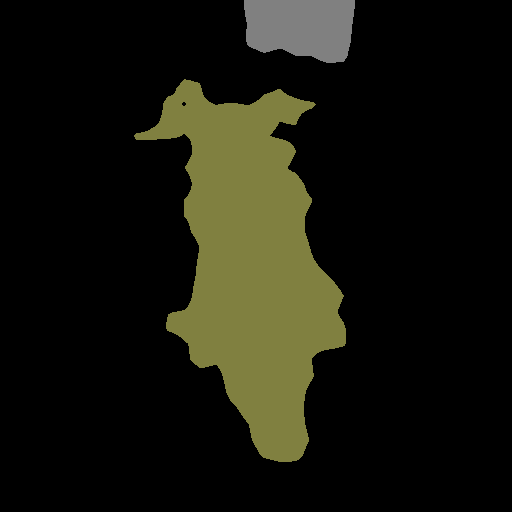}\\      \vspace{0.05cm}  
        \includegraphics[width=0.993\textwidth,height=0.7in]{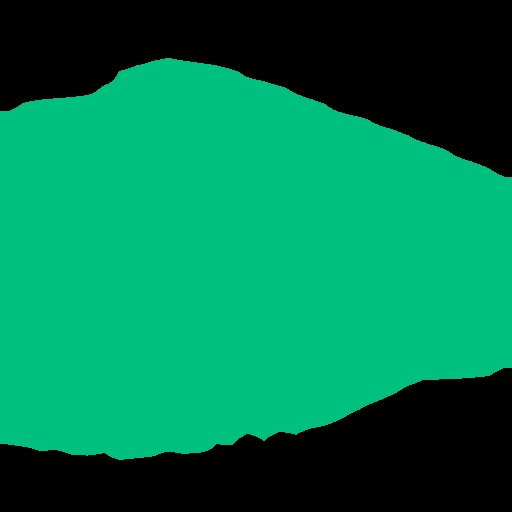}\\ \vspace{0.05cm}
        \includegraphics[width=0.993\textwidth,height=0.7in]{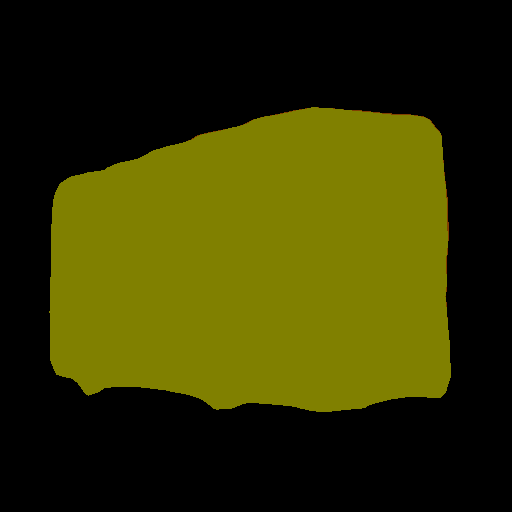}\\   \vspace{0.05cm}
        \includegraphics[width=0.993\textwidth,height=0.7in]{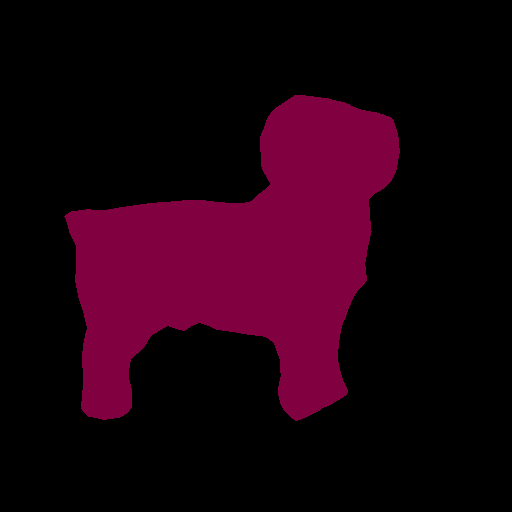}\\ \vspace{0.05cm}
        \includegraphics[width=0.993\textwidth,height=0.7in]{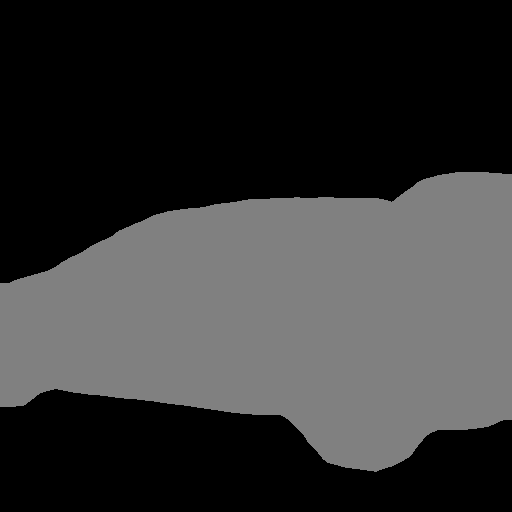}\\ \vspace{0.05cm}
        \includegraphics[width=0.993\textwidth,height=0.7in]{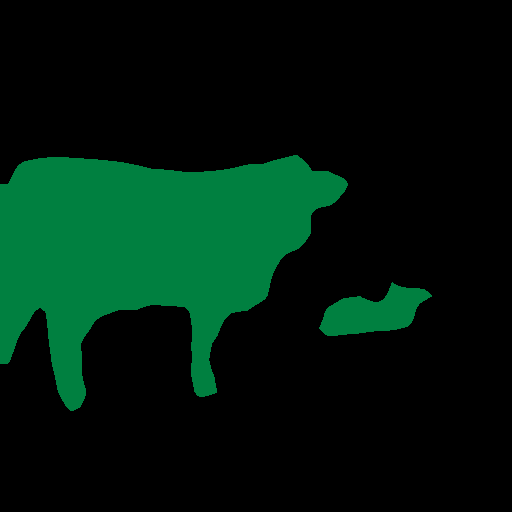}\\ \vspace{0.05cm}
        \includegraphics[width=0.993\textwidth,height=0.7in]{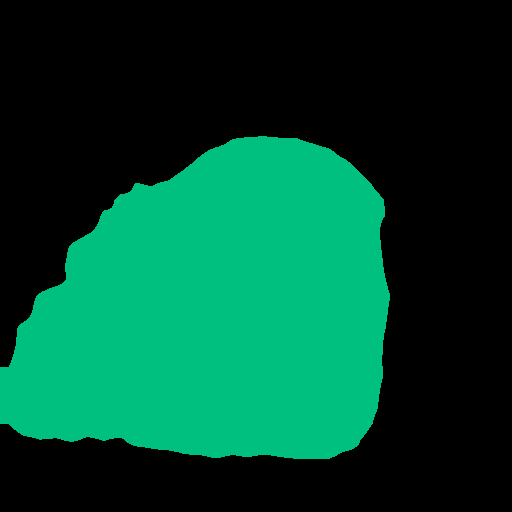}\\ \vspace{0.05cm}
        \includegraphics[width=0.993\textwidth,height=0.7in]{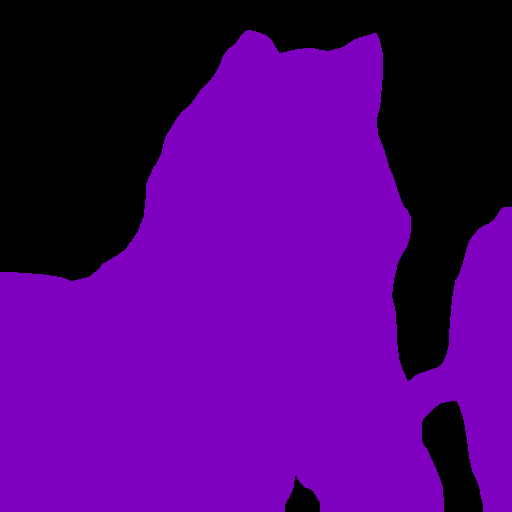}\\  \vspace{0.05cm}
        \includegraphics[width=0.993\textwidth,height=0.7in]{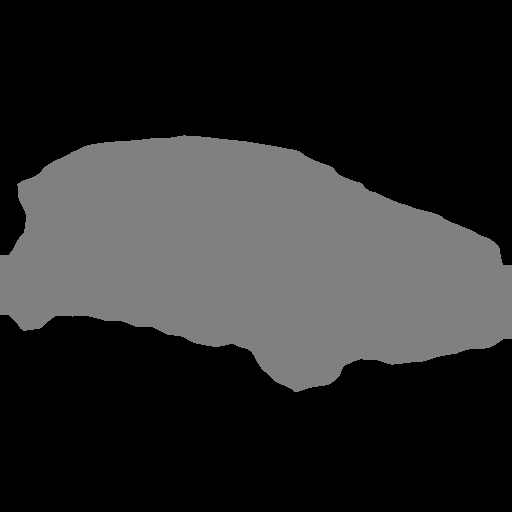}\\ \vspace{0.05cm}
        \includegraphics[width=0.993\textwidth,height=0.7in]{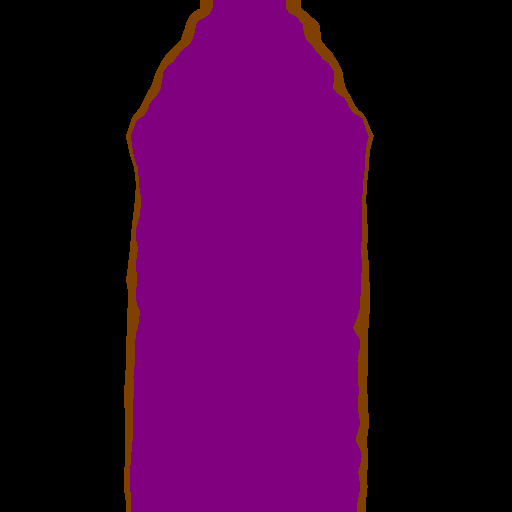}\\   
        (d) Ours \\
    \end{minipage}%
}%
\subfigure{
    \begin{minipage}{0.191\linewidth}
        \centering
        \includegraphics[width=0.993\textwidth,height=0.7in]{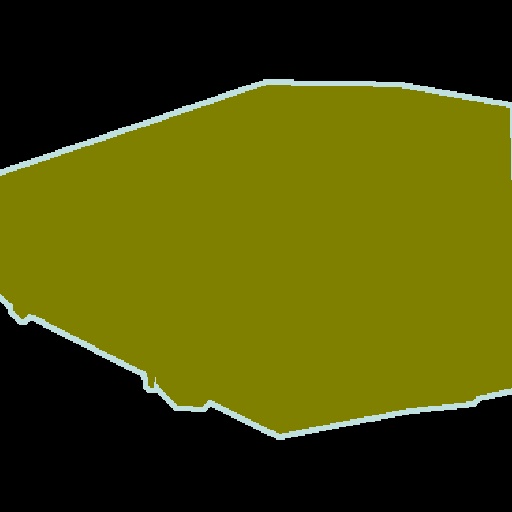}\\ \vspace{0.05cm}
        \includegraphics[width=0.993\textwidth,height=0.7in]{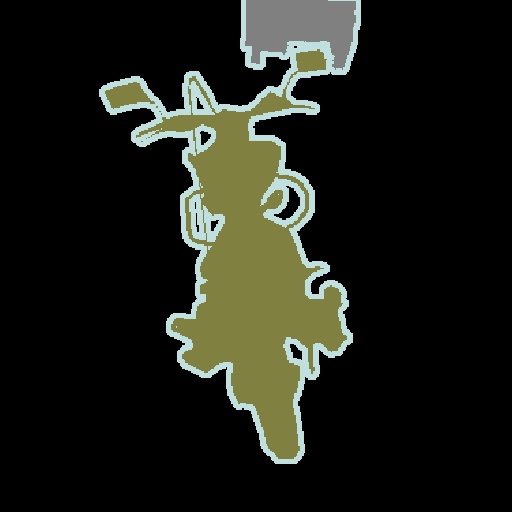}\\      \vspace{0.05cm}   
        \includegraphics[width=0.993\textwidth,height=0.7in]{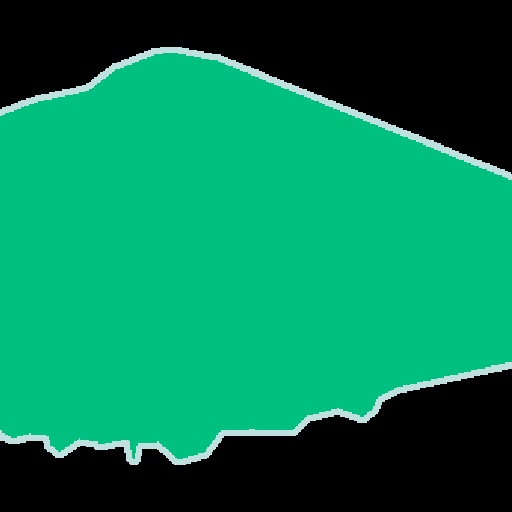}\\ \vspace{0.05cm}
        \includegraphics[width=0.993\textwidth,height=0.7in]{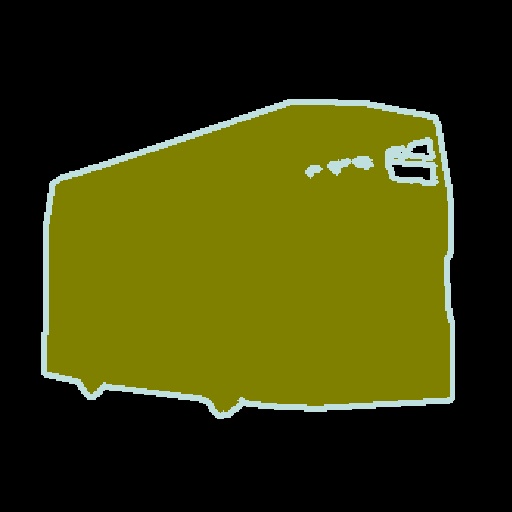}\\   \vspace{0.05cm}
        \includegraphics[width=0.993\textwidth,height=0.7in]{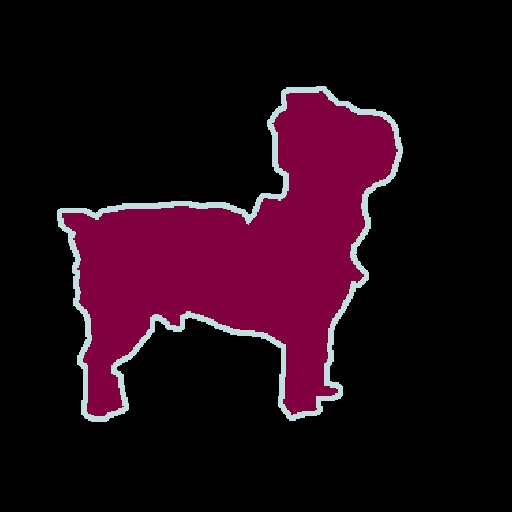}\\ \vspace{0.05cm}
        \includegraphics[width=0.993\textwidth,height=0.7in]{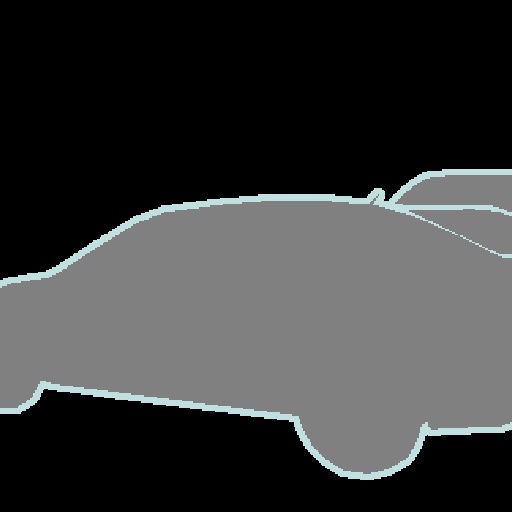}\\ \vspace{0.05cm}
        \includegraphics[width=0.993\textwidth,height=0.7in]{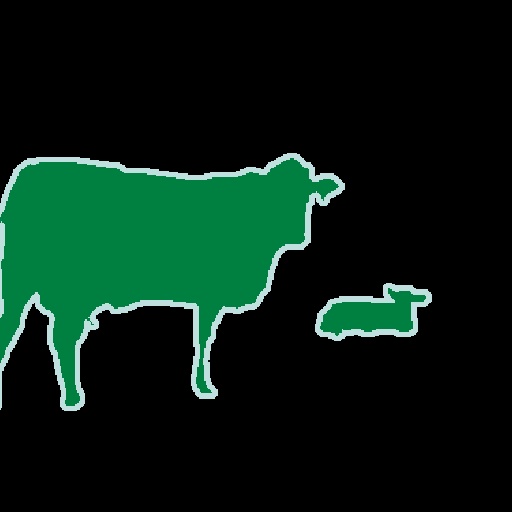}\\ \vspace{0.05cm}
        \includegraphics[width=0.993\textwidth,height=0.7in]{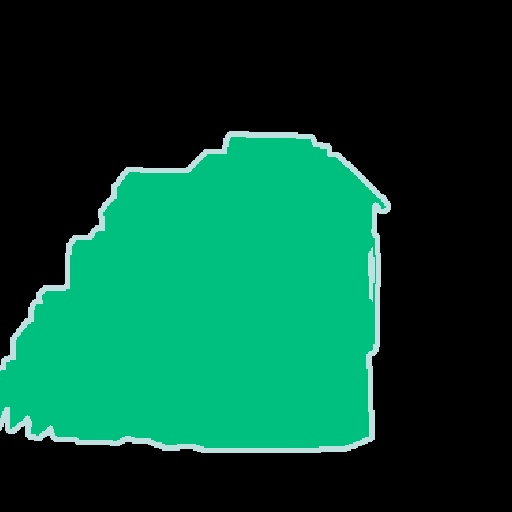}\\ \vspace{0.05cm}
        \includegraphics[width=0.993\textwidth,height=0.7in]{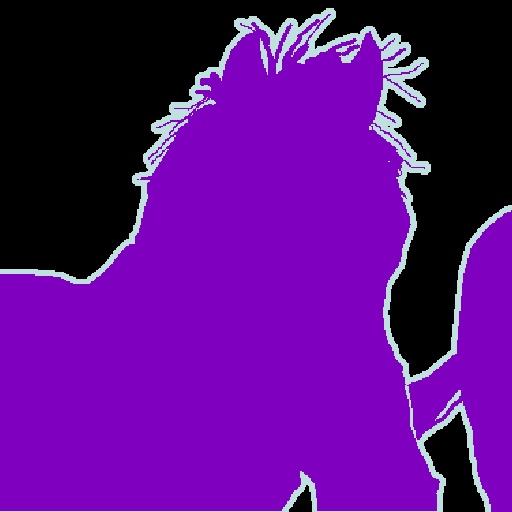}\\  \vspace{0.05cm}
        \includegraphics[width=0.993\textwidth,height=0.7in]{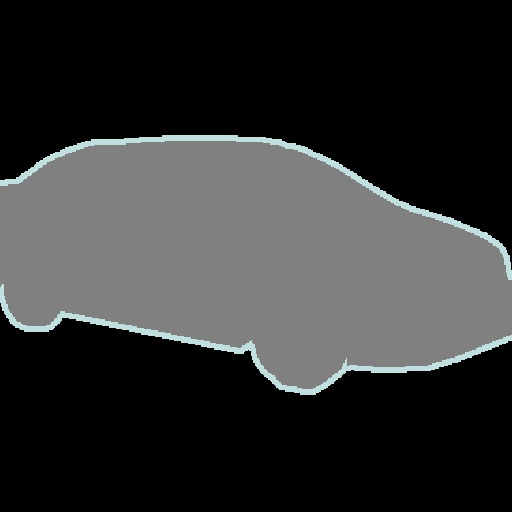}\\ \vspace{0.05cm}
        \includegraphics[width=0.993\textwidth,height=0.7in]{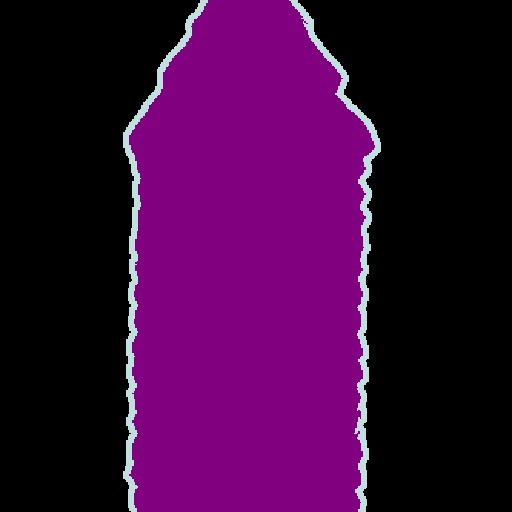}\\   
        (e) Ground Truth \\
    \end{minipage}%
}%
\centering

\caption{Qualitative results on the VOC 2012 dataset 19-1 (the first five rows) and 15-5 (the last six rows). They show the superiority of our approach on both new (\eg\ train) and old (\eg\ car, cow, bus) classes.
}

\label{fig:vis_qua_voc}
\end{figure*}

\begin{figure*}[!htb]
\centering
\subfigure{
    \begin{minipage}{0.191\linewidth}
        \centering
        \includegraphics[width=0.993\textwidth,height=0.7in]{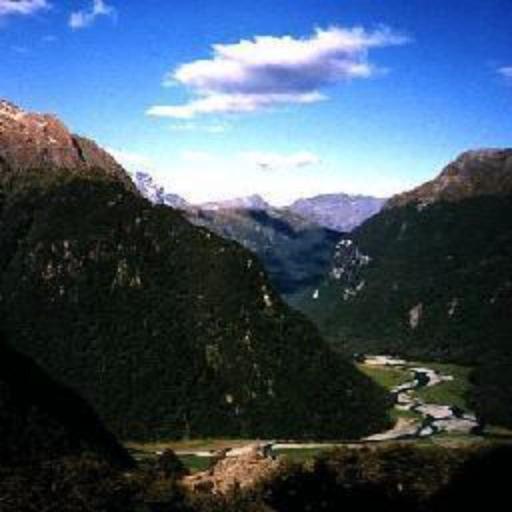}\\ \vspace{0.05cm}
        \includegraphics[width=0.993\textwidth,height=0.7in]{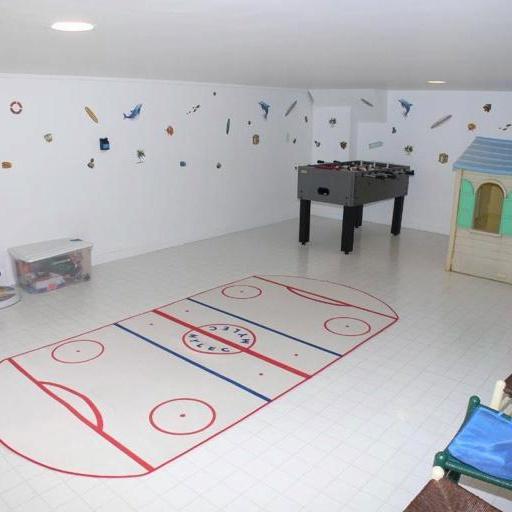}\\     \vspace{0.05cm}    
        \includegraphics[width=0.993\textwidth,height=0.7in]{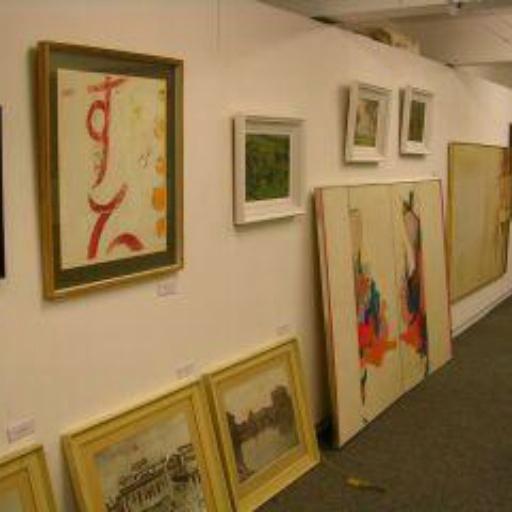}\\ \vspace{0.05cm}
        \includegraphics[width=0.993\textwidth,height=0.7in]{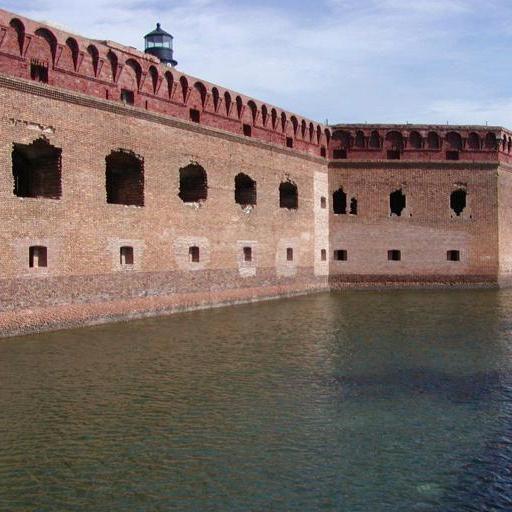}\\ \vspace{0.05cm}
        \includegraphics[width=0.993\textwidth,height=0.7in]{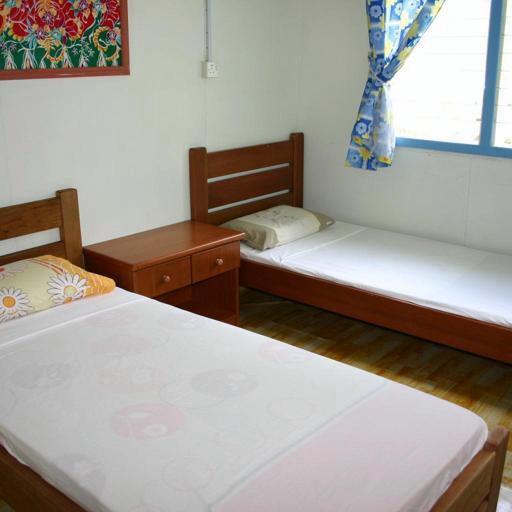}\\ \vspace{0.05cm}
        \includegraphics[width=0.993\textwidth,height=0.7in]{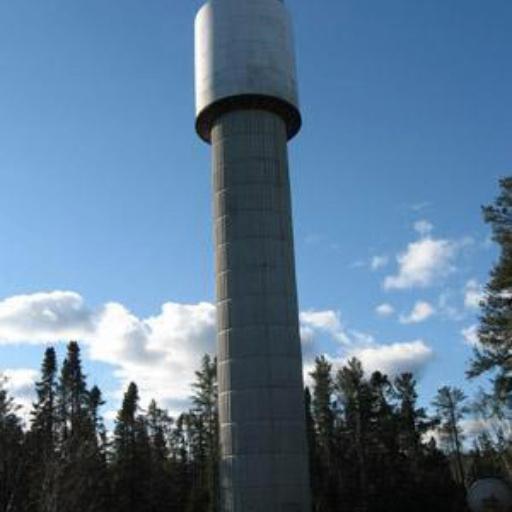}\\  
        (a) Input Image
    \end{minipage}%
}%
\subfigure{
    \begin{minipage}{0.191\linewidth}
        \centering
        \includegraphics[width=0.993\textwidth,height=0.7in]{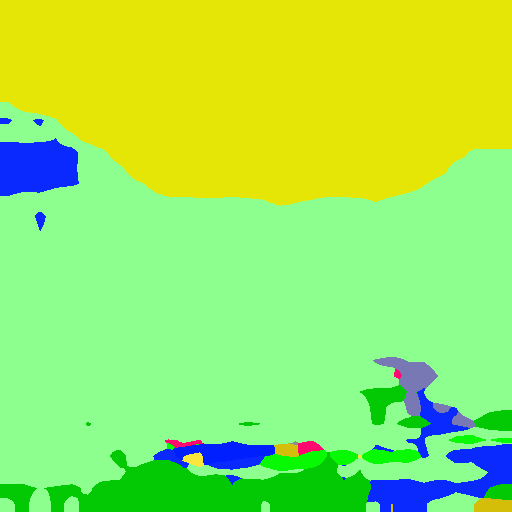}\\ \vspace{0.05cm}
        \includegraphics[width=0.993\textwidth,height=0.7in]{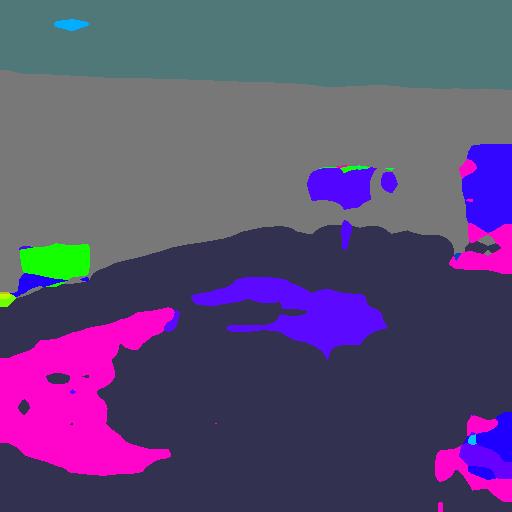}\\     \vspace{0.05cm}    
        \includegraphics[width=0.993\textwidth,height=0.7in]{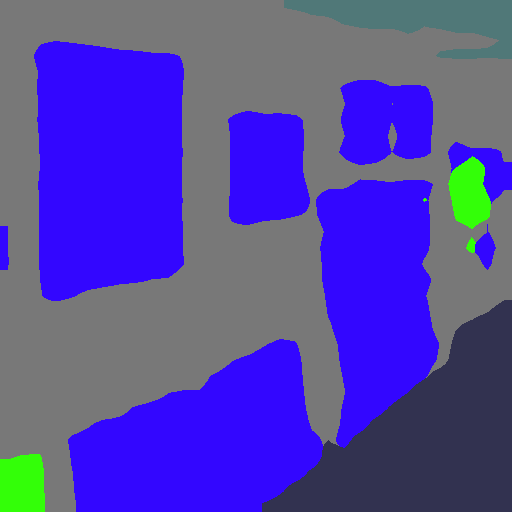}\\ \vspace{0.05cm}
        \includegraphics[width=0.993\textwidth,height=0.7in]{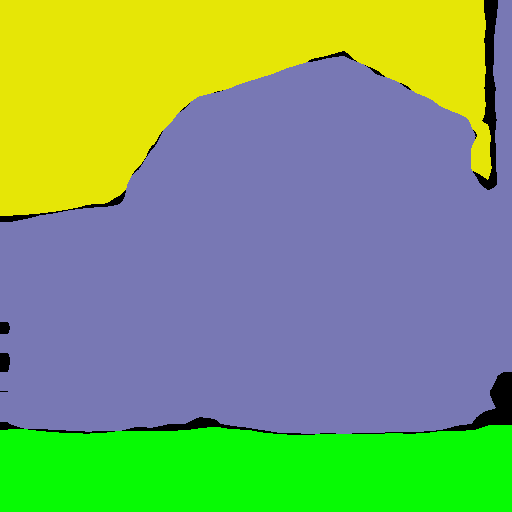}\\  \vspace{0.05cm}
        \includegraphics[width=0.993\textwidth,height=0.7in]{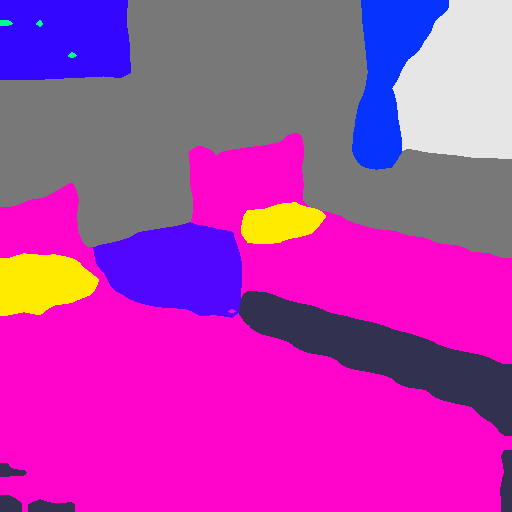}\\ \vspace{0.05cm}
        \includegraphics[width=0.993\textwidth,height=0.7in]{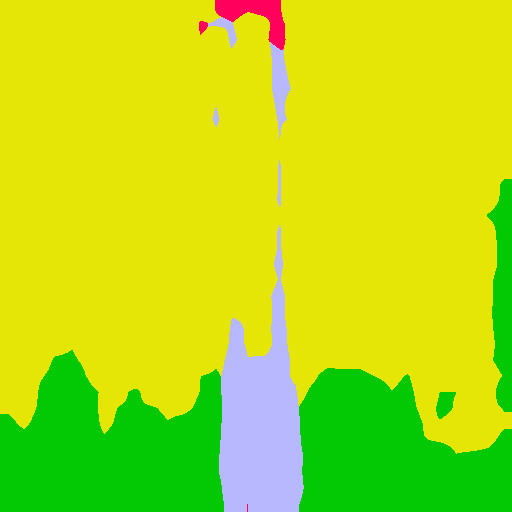}\\ 
        (b) ILT~\cite{michieli2019incremental}
    \end{minipage}%
}%
\subfigure{
    \begin{minipage}{0.191\linewidth}
        \centering
        \includegraphics[width=0.993\textwidth,height=0.7in]{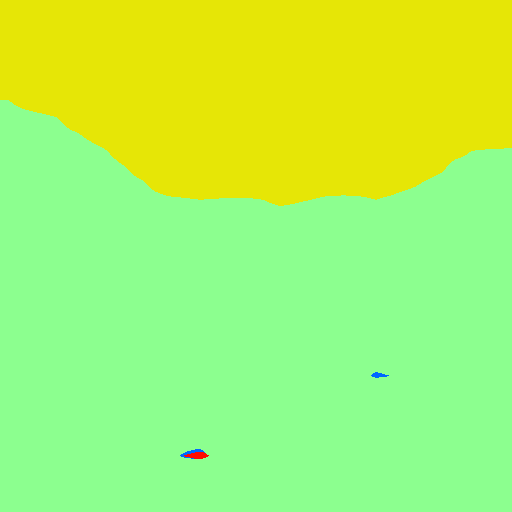}\\ \vspace{0.05cm}
        \includegraphics[width=0.993\textwidth,height=0.7in]{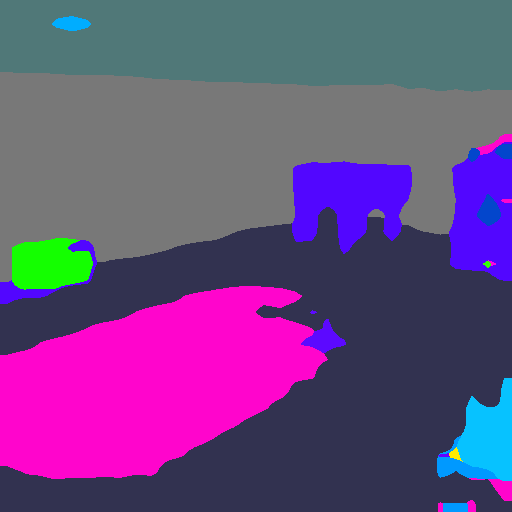}\\     \vspace{0.05cm}    
        \includegraphics[width=0.993\textwidth,height=0.7in]{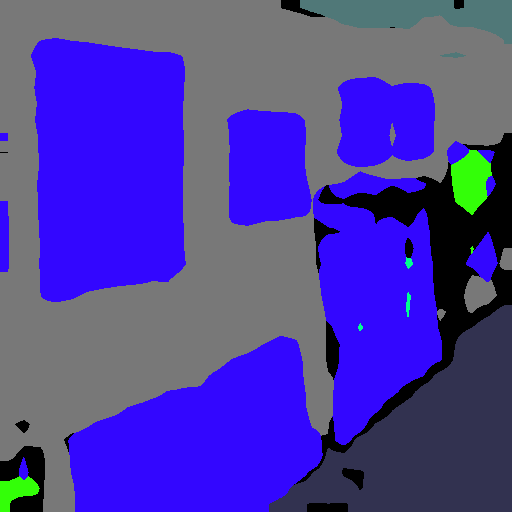}\\ \vspace{0.05cm}
        \includegraphics[width=0.993\textwidth,height=0.7in]{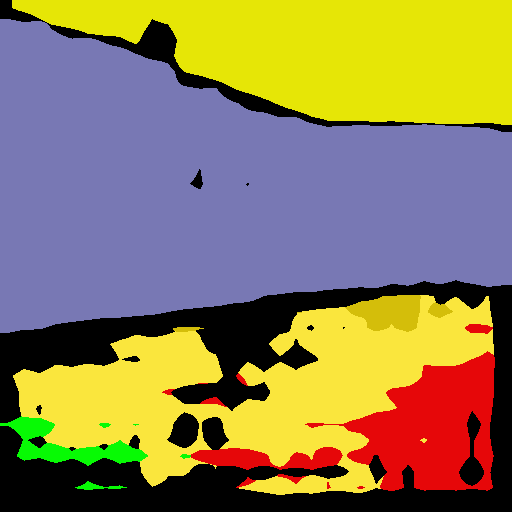}\\ \vspace{0.05cm}
        \includegraphics[width=0.993\textwidth,height=0.7in]{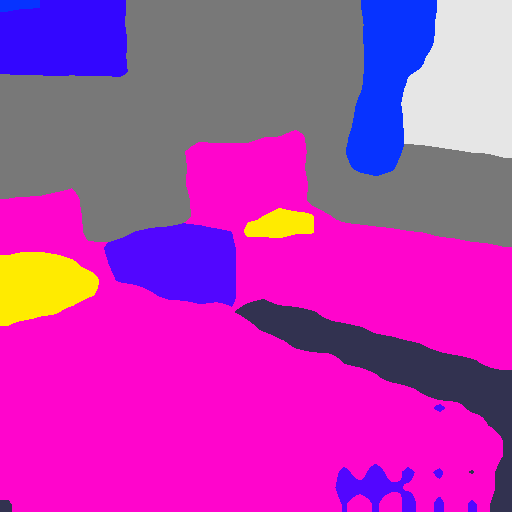}\\ \vspace{0.05cm}
        \includegraphics[width=0.993\textwidth,height=0.7in]{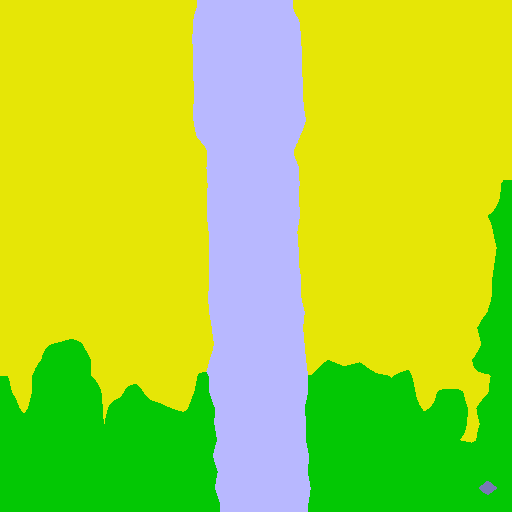}\\ 
        (c) MiB~\cite{cermelli2020modeling}
    \end{minipage}%
}%
\subfigure{
    \begin{minipage}{0.191\linewidth}
        \centering
        \includegraphics[width=0.993\textwidth,height=0.7in]{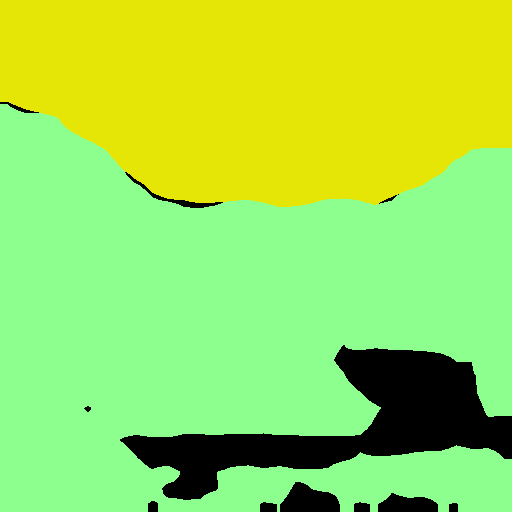}\\ \vspace{0.05cm}
        \includegraphics[width=0.993\textwidth,height=0.7in]{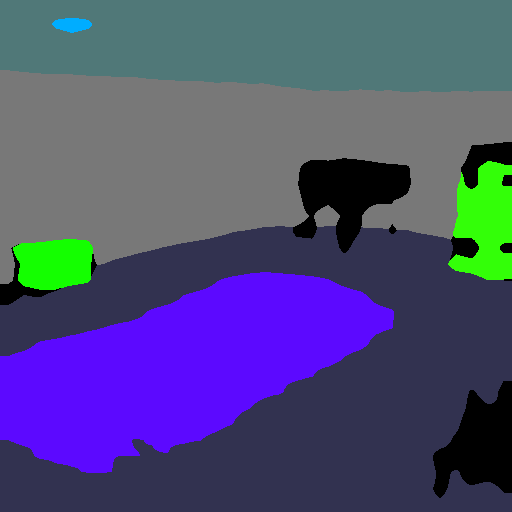}\\     \vspace{0.05cm}    
        \includegraphics[width=0.993\textwidth,height=0.7in]{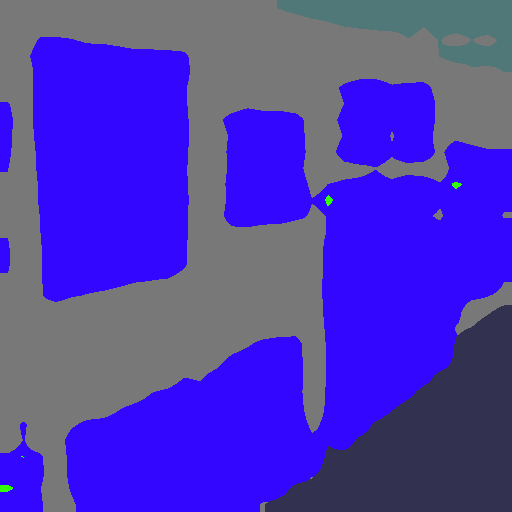}\\ \vspace{0.05cm}
        \includegraphics[width=0.993\textwidth,height=0.7in]{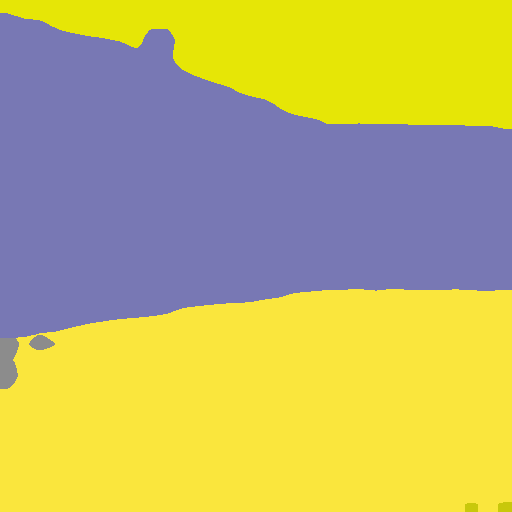}\\ \vspace{0.05cm}
        \includegraphics[width=0.993\textwidth,height=0.7in]{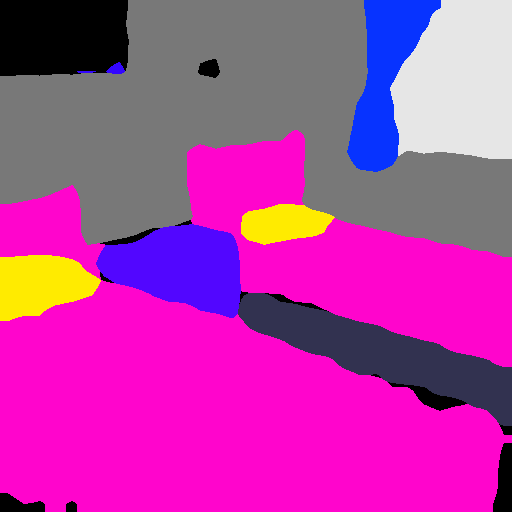}\\ \vspace{0.05cm}
        \includegraphics[width=0.993\textwidth,height=0.7in]{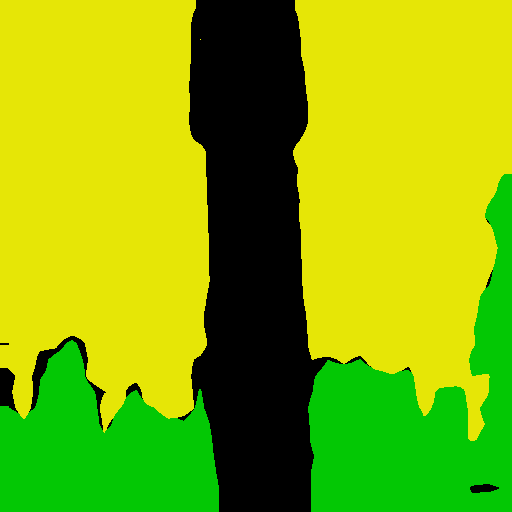}\\ 
        (d) Ours \\
    \end{minipage}%
}%
\subfigure{
    \begin{minipage}{0.191\linewidth}
        \centering
        \includegraphics[width=0.993\textwidth,height=0.7in]{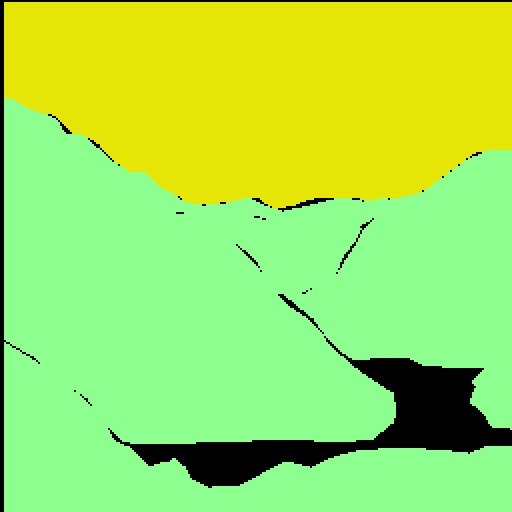}\\ \vspace{0.05cm}
        \includegraphics[width=0.993\textwidth,height=0.7in]{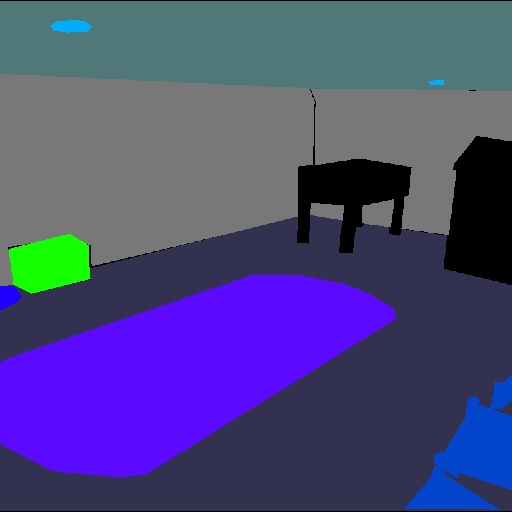}\\    \vspace{0.05cm}     
        \includegraphics[width=0.993\textwidth,height=0.7in]{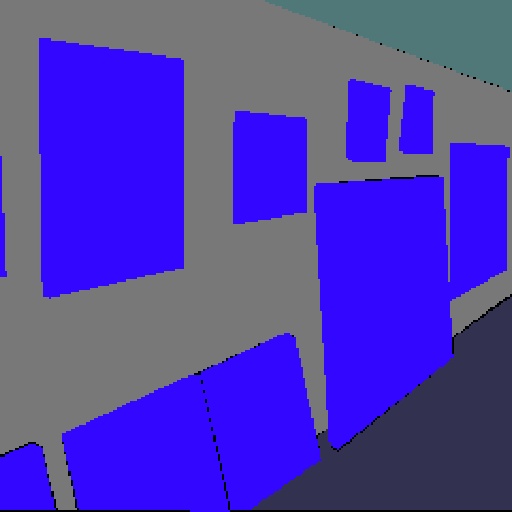}\\ \vspace{0.05cm}
        \includegraphics[width=0.993\textwidth,height=0.7in]{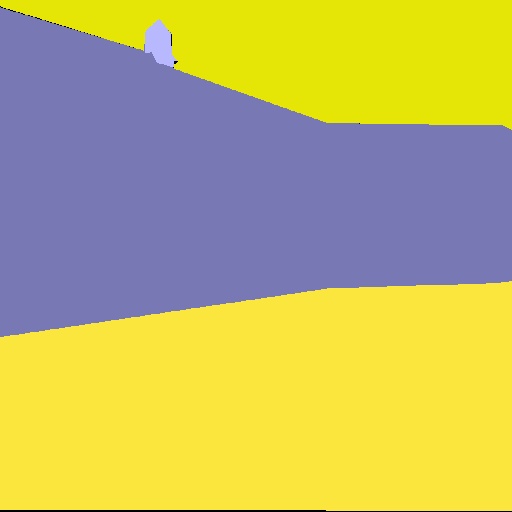}\\ \vspace{0.05cm}
        \includegraphics[width=0.993\textwidth,height=0.7in]{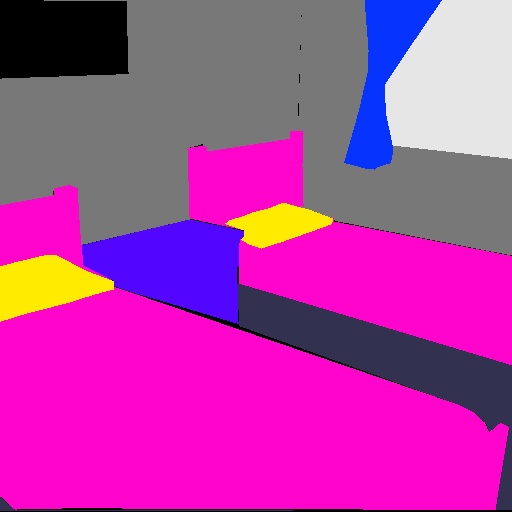}\\ \vspace{0.05cm}
        \includegraphics[width=0.993\textwidth,height=0.7in]{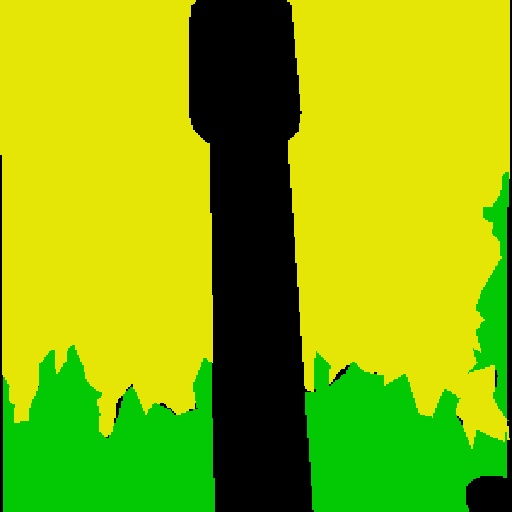}\\  
        (e) Ground Truth
    \end{minipage}%
}%
\centering
\caption{Qualitative results on the ADE20K dataset using 100-50 setup. The image demonstrates the superiority of our approach on both new (\eg\ sky, grass, wall) and old (\eg\ paint, pool, building) classes.}
\label{fig:vis_qua_ade}
\end{figure*}

Finally, we also analyze strategies for minimizing the computational complexity and the memory usage at test time. In particular, we believe that a truly continual learner should retain only one model, \ie{} the old model used for distillation and continual attentive fusion should be discarded. Hence, in Figure~\ref{fig:ablation cm test} we show different solutions for using the fusion module at test time. 
Possible strategies include: (i) \textbf{skip}: completely skipping the fusion module at test time, \textit{i.e.} passing only $\nonlocalfeatures^\ell$ to subsequent layers; (ii) \textbf{padding zero}: concatenating $\nonlocalfeatures^\ell$ with a zero-padding vector of the same size and evaluating the fusion module; (iii) \textbf{concat}: concatenating $\nonlocalfeatures^\ell$ and $\nonlocalfeatures^{\ell-1}$ at training time. To compare the performance of these strategies, we evaluate the accuracy of the network on both new and old classes every three epochs during the training trajectory of the second task (15-5 setting). Interestingly, we obtain comparable overall performance when skipping the fusion module (strategy (i)), while zero-padding seems slightly suboptimal. Also, it is important to notice that at the beginning of the training, concatenating old and new features brings great performance improvements on the old classes, while the difference is negligible at convergence. It validates the idea that the old model is helping the new model through continual attentive fusion, transferring valuable information that the new model then uses to counteract catastrophic forgetting. On the other hand, as expected, the accuracy is unchanged when using concatenation on the new classes. It is reasonable since the old model does not possess any knowledge of the new classes, and therefore cannot help the new model.

\begin{table}[t]
\centering
\caption{ Ablation study about applied position on Pascal- VOC 2012 dataset.}
\label{tab:ab_position}
\resizebox{0.85\linewidth}{!}{%
\begin{tabular}{ccccccc}
\toprule
\multicolumn{1}{l}{\multirow{2}{*}{Combination}} & \multicolumn{3}{c}{15-1 disjoint} & \multicolumn{3}{c}{19-1 disjoint} \\ \cmidrule(lr){2-4} \cmidrule(lr){5-7} 
\multicolumn{1}{l}{} & 1-15 & 16-20 & all & 1-19 & 20 & all \\
\midrule
$z_{\ell}+h_{\ell}$ & \textbf{57.2} & \textbf{15.5} & \textbf{46.7} &\textbf{75.5}  & \textbf{30.8} & \textbf{73.3} \\
$h_{\ell}$ & 56.2 & 15.2 & 45.9 & 74.3 & 29.6 & 72.1 \\
$z_{\ell}$ & 56.4 & 15.3 & 46.1 & 74.8 & 30.1 & 72.5\\
\bottomrule
\end{tabular}%
}
\end{table}

\noindent \textbf{Sensitivity Analysis of $\lambda_{\textsc{AD}}$ and $\lambda_{\textsc{D}}$.}
We provide a more thorough analysis of the loss weighting parameters $\lambda_{\textsc{AD}}$ and $\lambda_{\textsc{D}}$, see~Figure~\ref{fig:vis_ab_para}. We evaluate the average IoU of our method for different values of the loss weights around the working point. We can see that $\lambda_{\textsc{AD}}$ is more of a critical choice than $\lambda_{\textsc{D}}$. Indeed, the operating region of $\lambda_{\textsc{AD}}$ is around $10^3$, while the operating region of $\lambda_{\textsc{D}}$ is much larger, and its choice does not have much impact as long as it is kept below $10^2$.







\noindent \textbf{Attentive Feature Distillation Impact.}
In this section, we deeply analyze the effect of our attentive feature distillation. The results are shown in Table~\ref{tab:ab_attention}.  In the table, FD, $AD_{SP}$ and $AD_{CH}$ denote normal feature distillation without attention, spatial-wise attentive feature distillation, and channel-wise attentive feature distillation, respectively. According to Table~\ref{tab:ab_attention}, only adding channel-wise attentive feature distillation leads to training failure, while spatial-wise attentive feature distillation significantly improves the performance. Also, a combination of both attentive feature distillation further boosts the result. It proves that the performance improvement is not due to complex attention modules like the squeeze-and-attention module, but the choice of an appropriate attention module and a suitable combination. Moreover, we also run a position sensitivity analysis of attentive feature distillation in Table~\ref{tab:ab_position}. According to the results, we choose $z_{\ell}+h_{\ell}$ as the input of attentive feature distillation for all experiments.
The results also confirm that using the old model $\phi^{\ell-1}$ during training can significantly improve the performance compared to only using $\phi^{\ell}$. Furthermore, using the CAF module without the old model does not improve performance, suggesting that the improvement does not come from the additional parameters introduced in the CAF module.

\noindent \textbf{Qualitative Results.}
Qualitative results associated with our method on ADE20K and VOC 2012 dataset are shown in Figures~\ref{fig:vis_qua_voc} and~\ref{fig:vis_qua_ade}. We found that CAF not only preserves more knowledge on the old classes with respect to MiB and ILT, but also produces accurate segmentations for the objects of the new categories.

\begin{figure*}[htp]
    \centering
    \includegraphics[width=1\textwidth]{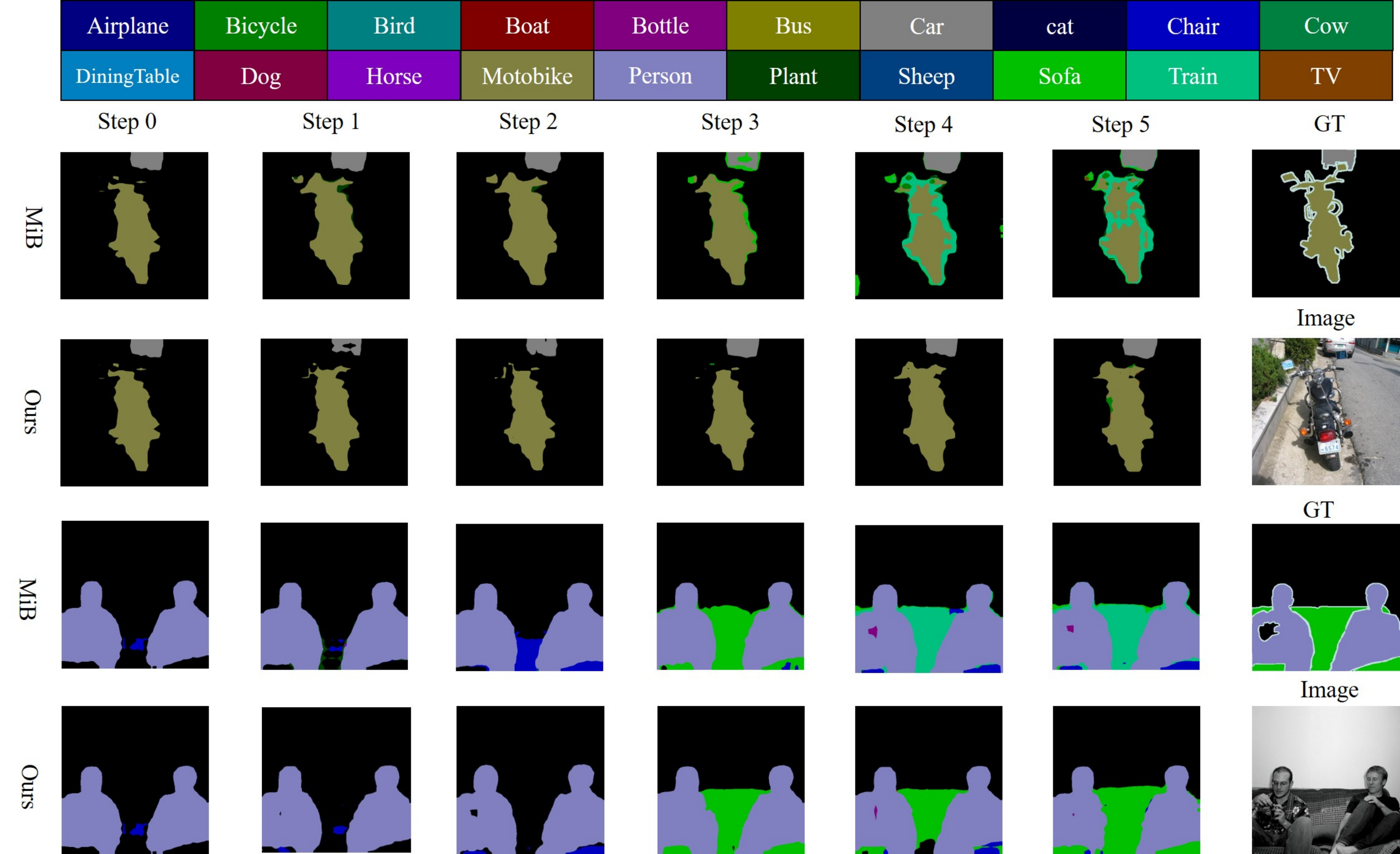}
    \caption{Visualization of MiB and our method predictions across time in VOC 15-1 for two test images.}
    \label{fig:voc-15-5-step}
\end{figure*}

\begin{figure*}
    \centering
    \includegraphics[width=0.993\textwidth]{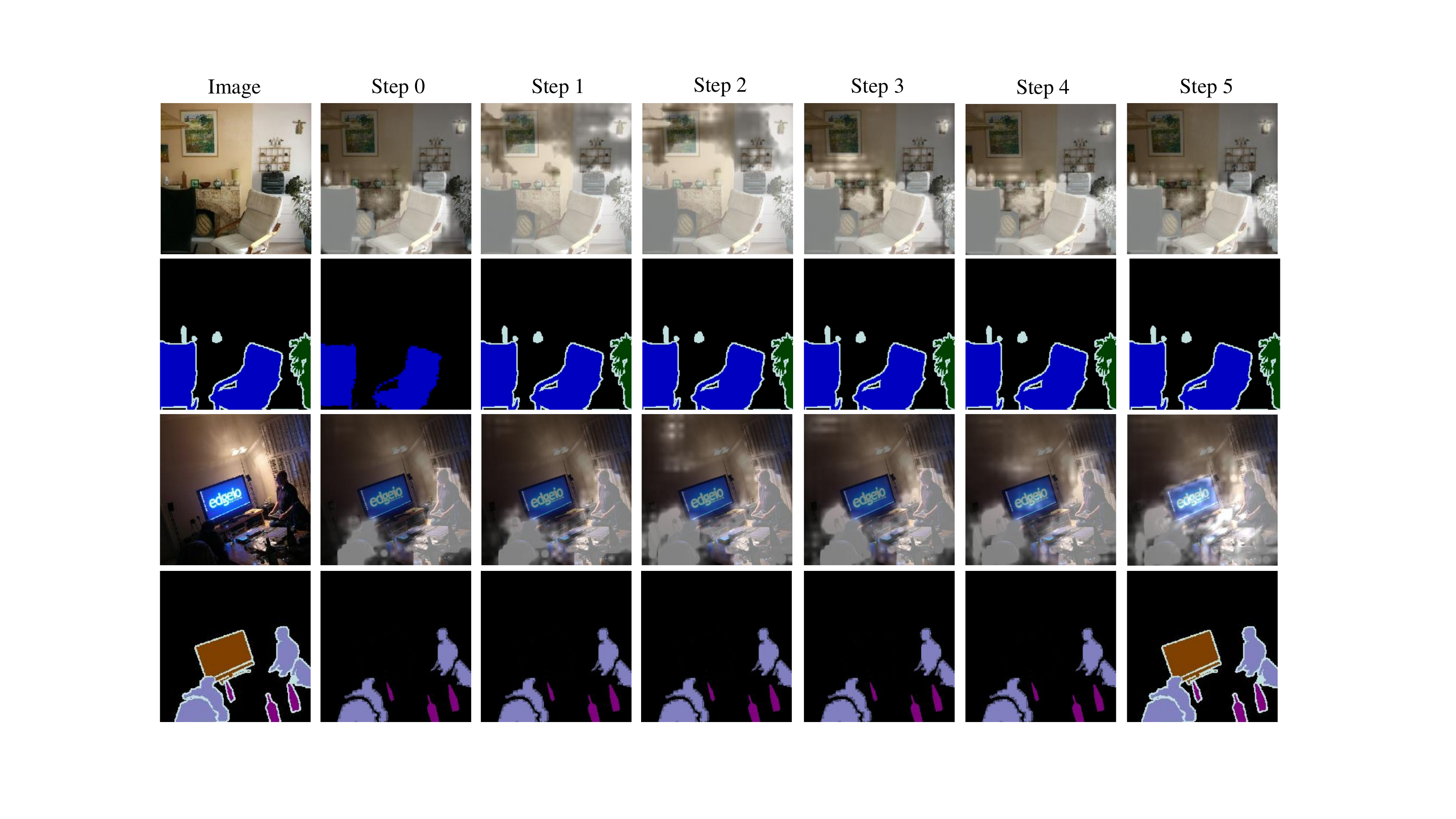}
    \caption{Visualization of our method's attention across time in VOC 15-1 for two test images.}
    \label{fig:att-step}
\end{figure*}

Figure~\ref{fig:voc-15-5-step} shows the predictions for both MiB and our method on VOC 15-1 across time. It seems clear that, MiB quickly forgets the previous classes and becomes biased towards new classes. On the other hand, our method's predictions are much more stable,
owing to the CAF module for alleviating catastrophic forgetting by spatially constraining representations, and to attentive feature distillation for dealing with the background shift.
To analyze the relationship between continual attentive fusion and incremental learning, we show the qualitative results of our attention on VOC 15-1 disjoint task in Figure~\ref{fig:att-step}. In detail, the first and third rows show the networks' attention on each step while the second and fourth rows show the known (seen) classes on each step.
According to Figure~\ref{fig:att-step}, thanks to the continual attentive fusion, our model can keep focusing on the old classes. When a new class is introduced (\eg~plant, television) the model is able to focus on the new object without losing attention for the old ones.  

\begin{figure}[t]
\centering
\subfigure{
    \begin{minipage}{0.24\linewidth}
        \centering
        \includegraphics[width=0.993\textwidth,height=0.7in]{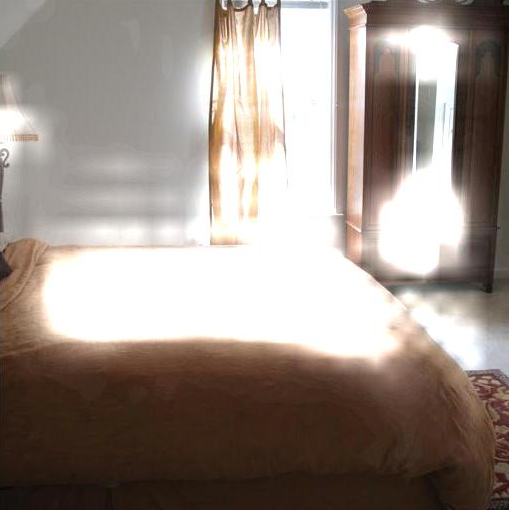}\\ \vspace{0.05cm}
        \includegraphics[width=0.993\textwidth,height=0.7in]{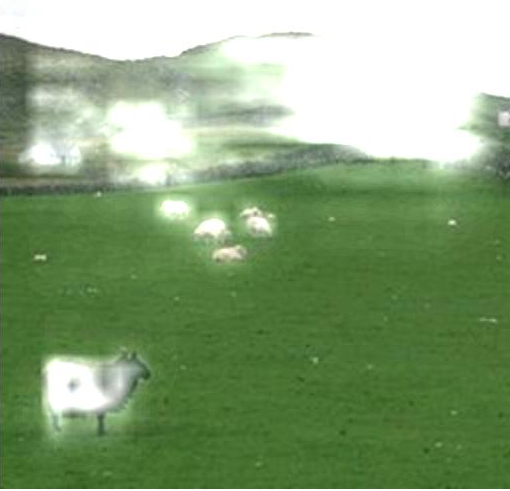}\\ \vspace{0.05cm}
        \includegraphics[width=0.993\textwidth,height=0.7in]{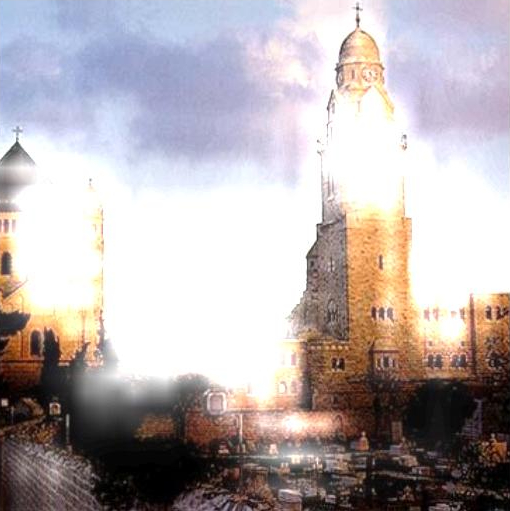}\\ \vspace{0.05cm}
        \includegraphics[width=0.993\textwidth,height=0.7in]{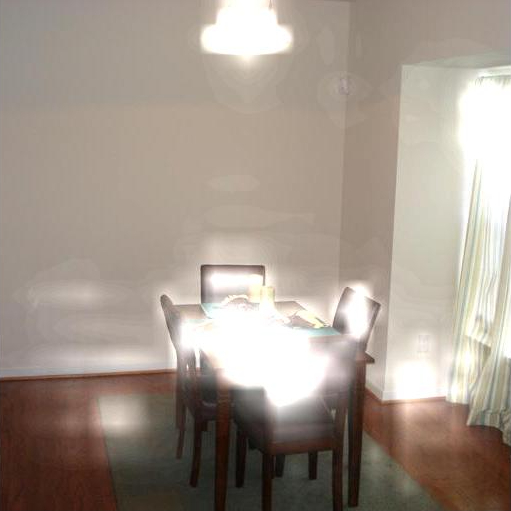}\\
    \end{minipage}%
}%
\subfigure{
    \begin{minipage}{0.24\linewidth}
        \centering
        \includegraphics[width=0.993\textwidth,height=0.7in]{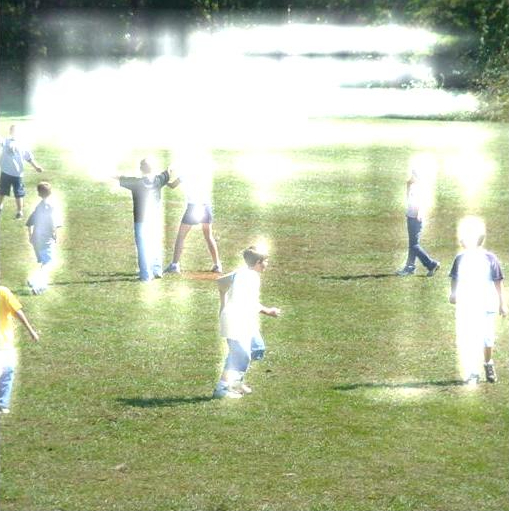}\\ \vspace{0.05cm}
        \includegraphics[width=0.993\textwidth,height=0.7in]{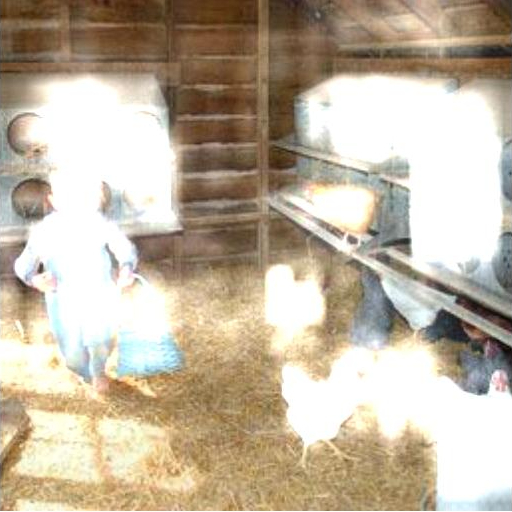}\\ \vspace{0.05cm}
        \includegraphics[width=0.993\textwidth,height=0.7in]{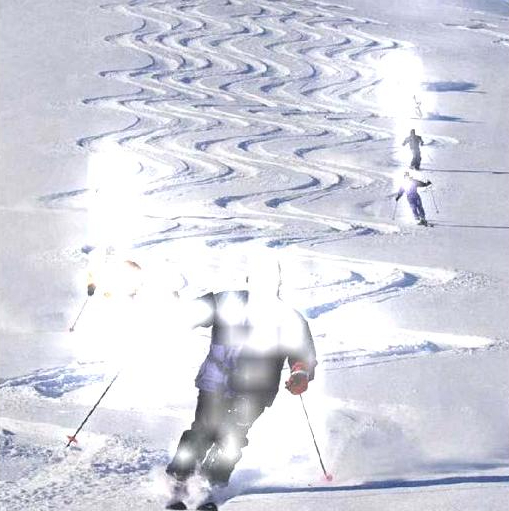}\\ \vspace{0.05cm}
        \includegraphics[width=0.993\textwidth,height=0.7in]{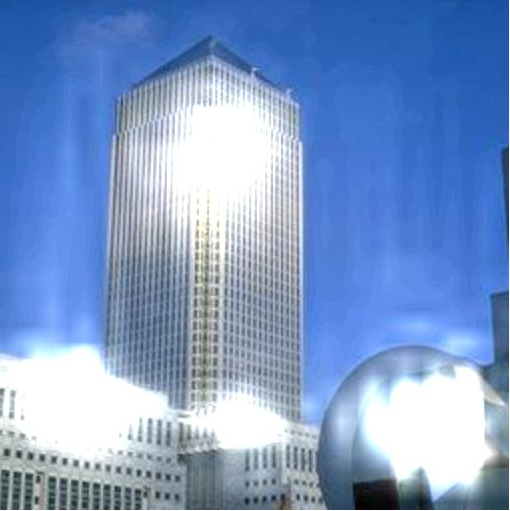}\\
    \end{minipage}%
}%
\subfigure{
    \begin{minipage}{0.24\linewidth}
        \centering
        \includegraphics[width=0.993\textwidth,height=0.7in]{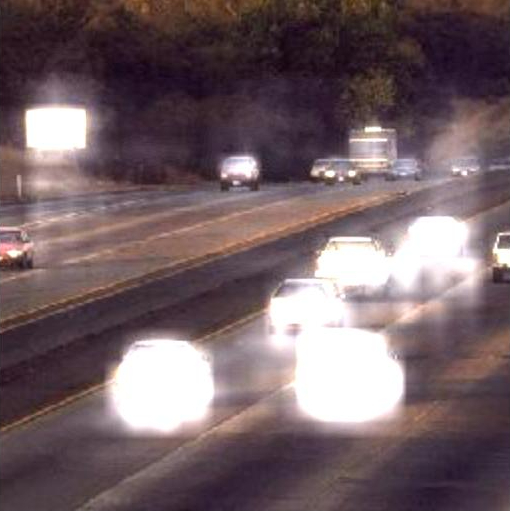}\\ \vspace{0.05cm}
        \includegraphics[width=0.993\textwidth,height=0.7in]{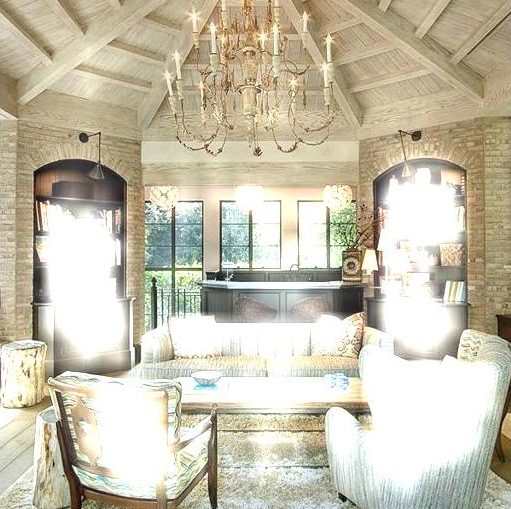}\\ \vspace{0.05cm}
        \includegraphics[width=0.993\textwidth,height=0.7in]{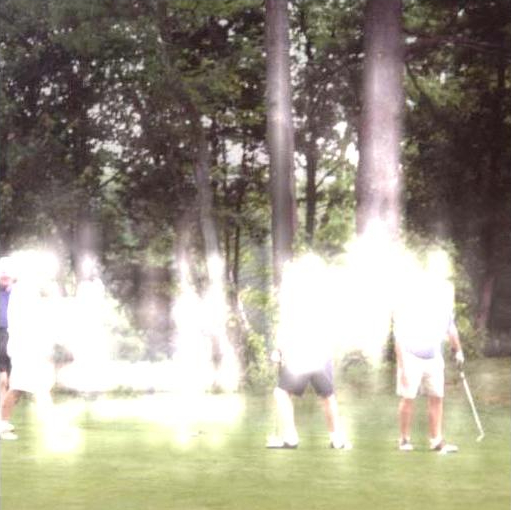}\\ \vspace{0.05cm}
        \includegraphics[width=0.993\textwidth,height=0.7in]{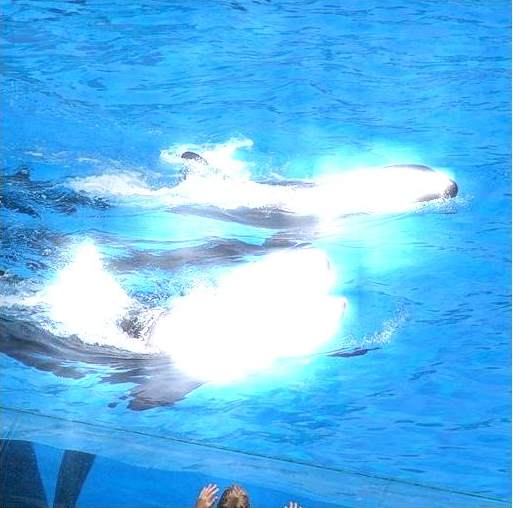}\\
    \end{minipage}%
}%
\subfigure{
    \begin{minipage}{0.24\linewidth}
        \centering
        \includegraphics[width=0.993\textwidth,height=0.7in]{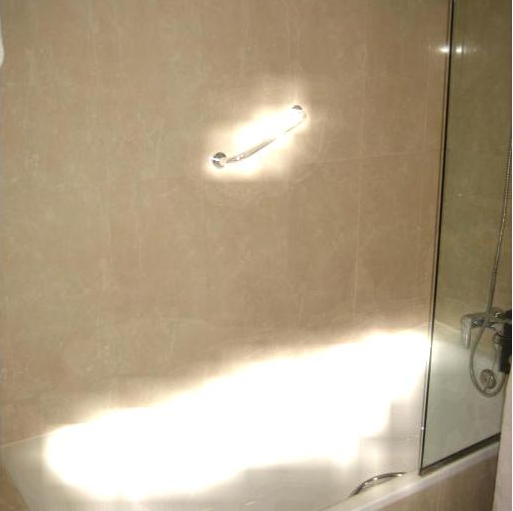}\\ \vspace{0.05cm}
        \includegraphics[width=0.993\textwidth,height=0.7in]{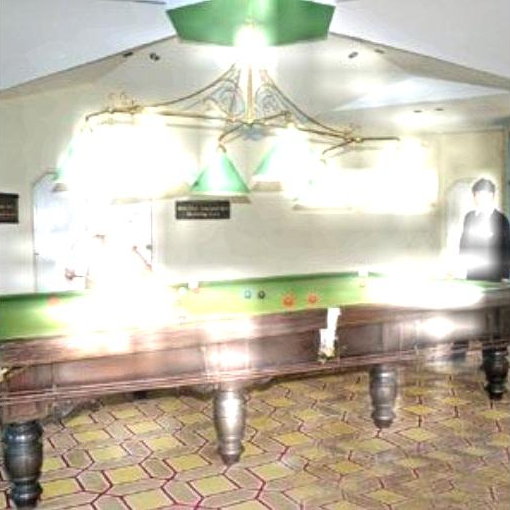}\\ \vspace{0.05cm}
        \includegraphics[width=0.993\textwidth,height=0.7in]{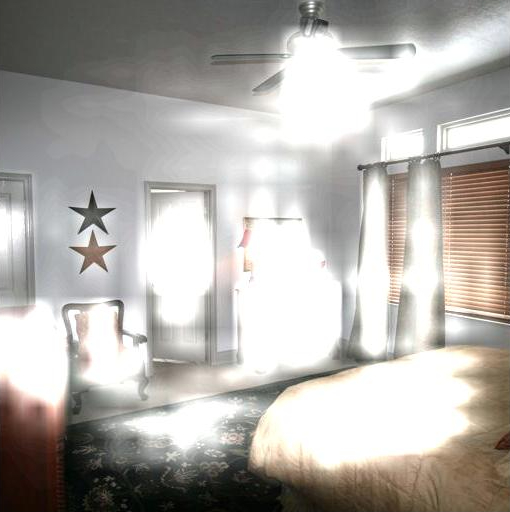}\\ \vspace{0.05cm}
        \includegraphics[width=0.993\textwidth,height=0.7in]{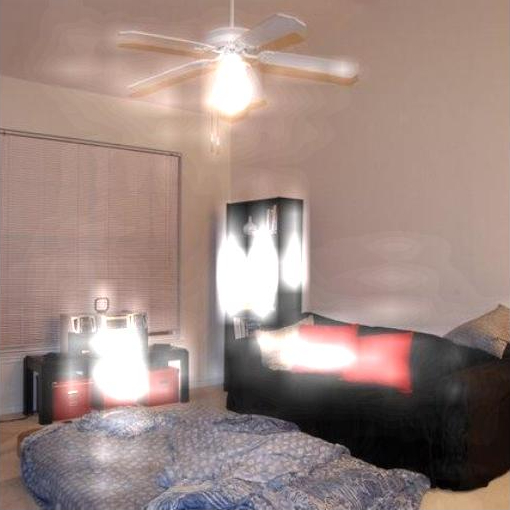}\\
    \end{minipage}%
}%
\caption{Attention map on the ADE20K dataset (50-50).}
\label{fig:vis_ade_att}
\end{figure}

Figure~\ref{fig:vis_ade_att} shows examples of the attention maps that our model learns on ADE20K (50-50 setting). It is quite clear that the network is able to focus on regions containing objects of the old classes (person, chair, building) as well as new classes (sideboard, animal, shower).

\section{Conclusions}

We propose the first attention-based ICL method for semantic segmentation. Our methodological contribution is three-fold. First, a new continual attentive fusion module that updates the current features by using the information of the previous model was proposed. While the information from the previous model is used at training time through a fusion module, it is discarded at test time to save resources. We also propose a new attentive distillation loss that leverages both channel-wise and spatial attention to transfer compelling information. Finally, we introduce a new method for balancing old and new background probabilities in the distillation loss.
Our extensive experimental evaluation demonstrates outstanding performance of our method in several datasets (VOC 2012 and ADE20K) and settings (14 in total).

\ifCLASSOPTIONcompsoc
  \section*{Acknowledgments}
\else
  \section*{Acknowledgment}
\fi

The authors would like to thank Nvidia Corporation for GPU donations to support our research.




%
%
%
\bibliographystyle{IEEEtran}
\bibliography{egbib}

%




\begin{IEEEbiography}[{\includegraphics[width=1in,height=1.25in,clip,keepaspectratio]{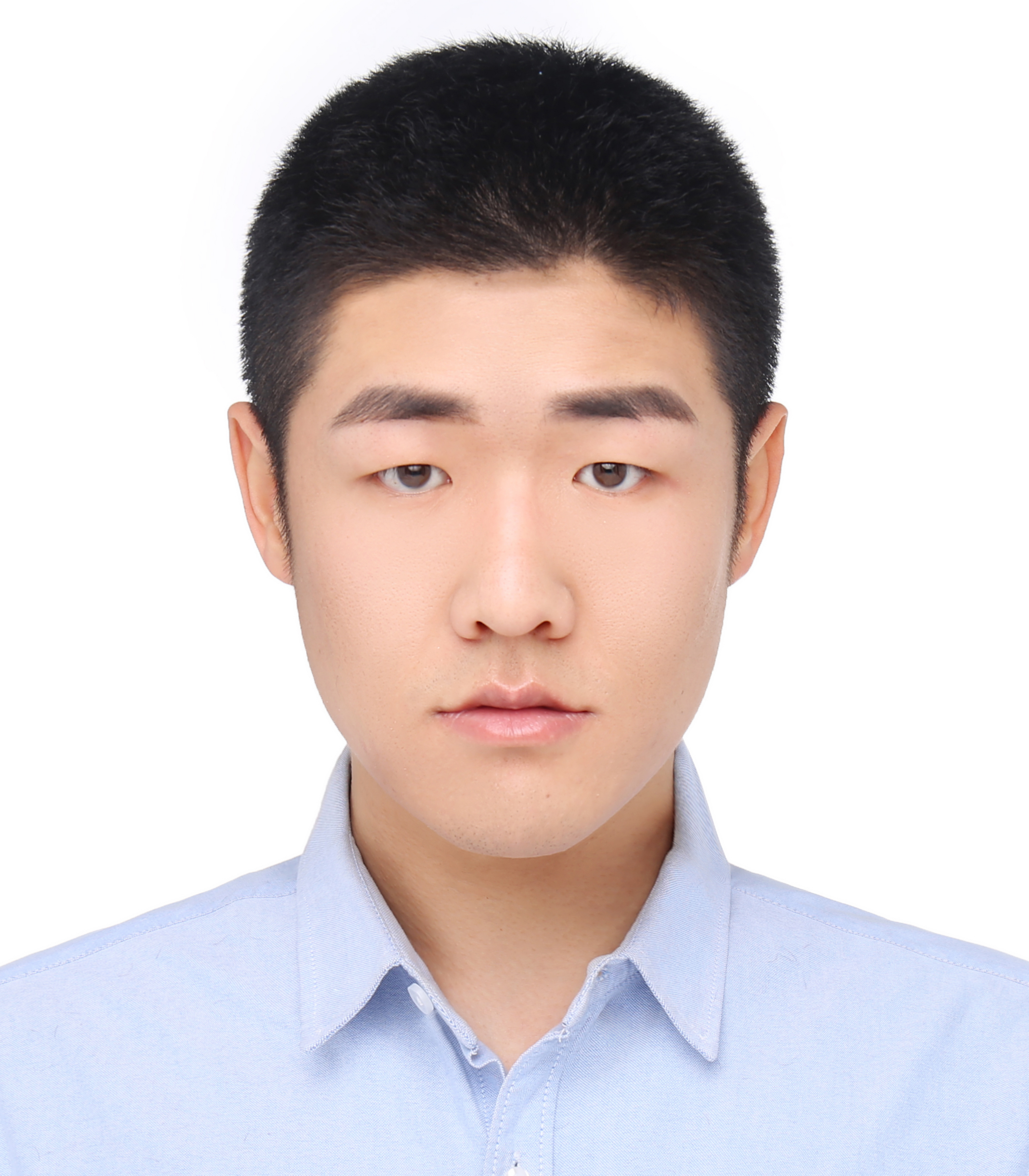}}]{Guanglei Yang}
received the B.S. degree in instrument science and technology from Harbin Institute of Technology (HIT), Harbin, China, in 2016. He is currently pursuing the Ph.D degree in the School of Instrumentation Science and Engineering, Harbin Institute of Technology (HIT), Harbin, China. He is working at University of Trento as a visiting student from 2020 to now. His research interests mainly include domain adaption, pixel-level prediction and attention gate.
\end{IEEEbiography}

\begin{IEEEbiography}[{\includegraphics[width=1in,height=1.25in,clip,keepaspectratio]{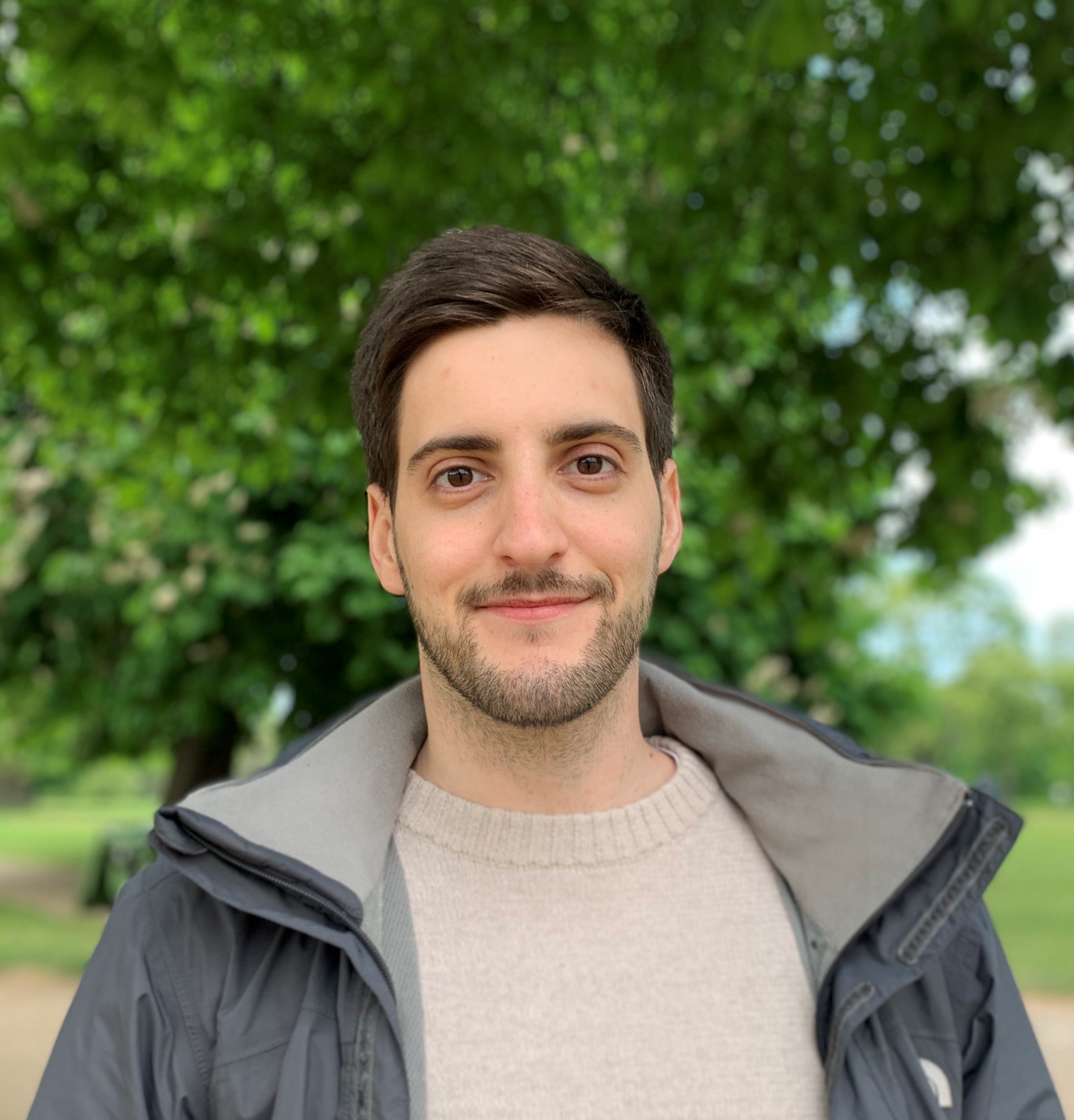}}]{Enrico Fini} is a Ph.D. student at the University of Trento. His research focuses on continual learning and self-supervised learning.
He received the B.S. degree in computer engineering from the University of Parma, Italy, in 2015 and the M.S. degree in computer science and engineering from Politecnico di Milano, Italy, in 2019. In 2018, He spent one year at the European Space Astronomy Centre of the European Space Agency in Madrid, Spain, working on machine learning for automatic sunspot detection.
\end{IEEEbiography}

\begin{IEEEbiography}[{\includegraphics[width=1in,height=1.25in,clip,keepaspectratio]{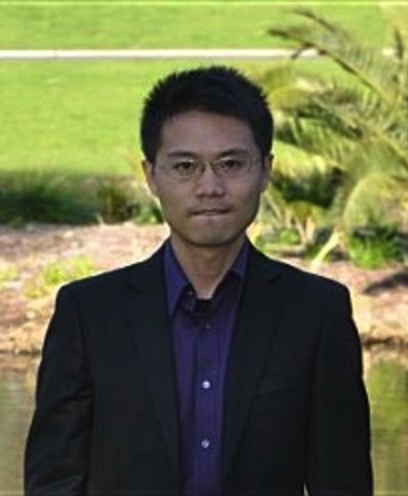}}]
{Dan Xu}
is an Assistant Professor in the Department of Computer Science and Engineering at HKUST. He was a Postdoctoral Research Fellow in VGG at the University of Oxford. He was a Ph.D. in the Department of Computer Science at the University of Trento. He was also a research assistant of MM Lab at the Chinese University of Hong Kong. He received the best scientific paper award at ICPR 2016, and a Best Paper Nominee at ACM MM 2018. He served as Area Chairs of ACM MM 2020, WACV 2021 and ICPR 2020.
\end{IEEEbiography}

\begin{IEEEbiography}[{\includegraphics[width=1in,height=1.25in,clip,keepaspectratio]{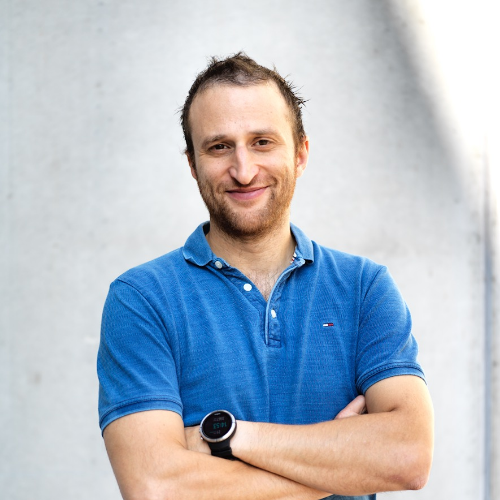}}]
{Paolo Rota}
is an assistant professor (RTDa) at University of Trento (in the MHUG group), working on computer vision and machine learning. He received his PhD in Information and Communication Technologies from the University of Trento in 2015. Prior joining UniTN he worked as Post-doc at the TU Wien and at the Italian Institute of Technology (IIT) of Genova. He is also collaborating with the ProM facility of Rovereto on assisting companies in inserting machine learning in their production chain.
\end{IEEEbiography}

\begin{IEEEbiography}[{\includegraphics[width=1in,height=1.25in,clip,keepaspectratio]{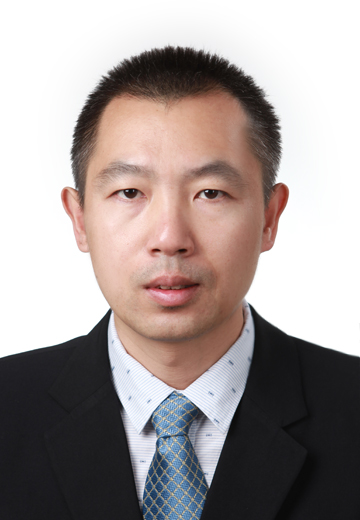}}]{Mingli Ding}
received the B.S., M.S. and Ph.D. degrees in instrument science and technology from Harbin Institute of Technology (HIT), Harbin, China, in 1996, 1997 and 2001, respectively. He worked as a visiting scholar in France from 2009 to 2010. Currently, he is a professor in the  School  of  Instrumentation Science and Engineering at Harbin Institute of Technology. Prof. Ding’s research interests are intelligence tests and information processing, automation test technology, computer vision, and machine learning. He has published over 40 papers in peer-reviewed journals and conferences.
\end{IEEEbiography}

\begin{IEEEbiography}[{\includegraphics[width=1in,height=1.25in,clip,keepaspectratio]{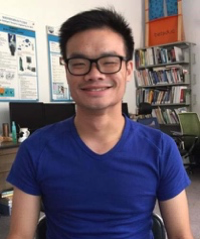}}]{Hao Tang}
is currently a Postdoctoral with Computer Vision Lab, ETH Zurich, Switzerland. 
He received the master’s degree from the School of Electronics and Computer Engineering, Peking University, China and the Ph.D. degree from Multimedia and Human Understanding Group, University of Trento, Italy.
He was a visiting scholar in the Department of Engineering Science at the University of Oxford. His research interests are deep learning, machine learning, and their applications to computer vision.
\end{IEEEbiography}

\begin{IEEEbiography}[{\includegraphics[width=1in,height=1.25in,clip,keepaspectratio]{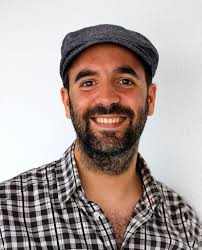}}]
{Xavier Alameda-Pineda} received M.Sc. degrees in mathematics (2008), in
telecommunications (2009) and in computer science
(2010) and a Ph.D. in mathematics and computer
science (2013) from Université Joseph Fourier. Since
2016, he is a Research Scientist at Inria Grenoble
Rhône-Alpes, with the Perception team. He served as
Area Chair at ICCV’17, of ICIAP’19 and of ACM
MM’19. He is the recipient of several paper awards
and of the ACM SIGMM Rising Star Award in 2018.
\end{IEEEbiography}

\begin{IEEEbiography}[{\includegraphics[width=1in,height=1.25in,clip,keepaspectratio]{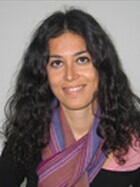}}]
{Elisa Ricci}
received the PhD degree from the
University of Perugia in 2008. She is an associate
professor at the University of Trento and a
researcher at Fondazione Bruno Kessler. She
has since been a post-doctoral researcher at
Idiap, Martigny, and Fondazione Bruno Kessler,
Trento. She was also a visiting researcher at the
University of Bristol. Her research interests are
mainly in the areas of computer vision and
machine learning. She is a member of the IEEE.
\end{IEEEbiography}

\end{document}